\documentclass[10pt,twocolumn,letterpaper]{article}
\pdfoutput=1 
\usepackage{iccv}
\usepackage{times}
\usepackage{epsfig}
\usepackage{graphicx}
\usepackage{amsmath}
\usepackage{amssymb}
\usepackage{multirow}
\usepackage{subfigure}
\usepackage{xfrac}
\usepackage{enumitem}
\usepackage{mathrsfs}
\usepackage{colortbl}
\usepackage{booktabs}
\usepackage[usenames,dvipsnames]{xcolor}
\usepackage{xcolor,colortbl}

\usepackage[pdftex,colorlinks=true,pdfstartview=FitV,linkcolor=black,citecolor=black,urlcolor=black,bookmarks=false]{hyperref}

\newcommand{\norm}[1]{\left\lVert#1\right\rVert}
\newcommand{\abs}[1]{\left\lvert#1\right\rvert}
\newcommand{\deltafun}[1]{\Delta\left(#1\right)}
\newcommand{\AIWE}{\operatorname{AIWE}}

\DeclareMathOperator*{\argmin}{arg\,min}

\newcommand{\image}[1]{I_{#1}}

\newcommand{\asubi}{a^{(i)}}
\newcommand{\latentasubi}{a_{\ell}^{(i)}} %
\newcommand{\bsubi}{b^{(i)}}

\newcommand{\lossfun}{\mathcal{L}}
\newcommand{\aperturesize}{L}
\newcommand{\lfold}{\lossfun_{\mathit{f}}}

\newcommand{\focusdistance}{g}
\newcommand{\worlddistance}{Z}
\newcommand{\depth}{D^*}
\newcommand{\preddepth}{\hat{D}}

\newcommand{\depthconf}{C}

\iccvfinalcopy %

\def\iccvPaperID{1334} %

\ificcvfinal\fi
\begin{document}

\title{Learning Single Camera Depth Estimation using Dual-Pixels}

\author{Rahul Garg \qquad Neal Wadhwa \qquad Sameer Ansari \qquad Jonathan T. Barron\\
Google Research
}

\maketitle
\ificcvfinal\thispagestyle{empty}\fi

\begin{abstract}

Deep learning techniques have enabled rapid progress in monocular depth estimation, but their quality is limited by the ill-posed nature of the problem and the scarcity of high quality datasets.
We estimate depth from a single camera by leveraging the dual-pixel auto-focus hardware that is increasingly common on modern camera sensors.
Classic stereo algorithms and prior learning-based depth estimation techniques underperform when applied on this dual-pixel data, the former due to too-strong assumptions about RGB image matching, and the latter due to not leveraging the understanding of optics of dual-pixel image formation.
To allow learning based methods to work well on dual-pixel imagery, we identify an inherent ambiguity in the depth estimated from dual-pixel cues, and develop an approach to estimate depth up to this ambiguity.
Using our approach, existing monocular depth estimation techniques can be effectively applied to dual-pixel data, and much smaller models can be constructed that still infer high quality depth.
To demonstrate this, we capture a large dataset of in-the-wild 5-viewpoint RGB images paired with corresponding dual-pixel data, and show how view supervision with this data can be used to learn depth up to the unknown ambiguity.
On our new task, our model is $30\%$ more accurate than any prior work on learning-based monocular or stereoscopic depth estimation.

\end{abstract}

\vspace{-0.13in}

\section{Introduction}

Depth estimation has long been a central problem in computer vision, both as a basic component of visual perception, and in service to various graphics, recognition, and robotics tasks.
Depth can be acquired via dedicated hardware that directly senses depth (time-of-flight, structured light, etc) but these sensors are often expensive, power-hungry, or limited to certain environments (such as indoors).
Depth can be inferred from multiple cameras through the use of multi-view geometry, but building a stereo camera requires significant complexity in the form of calibration, rectification, and synchronization.
Machine learning techniques can be used to estimate depth from a single image, but the under-constrained nature of image formation often results in inaccurate estimation.

\begin{figure}[b]
    \centering
    \subfigure[RGB image]{\includegraphics[width=0.15\textwidth]{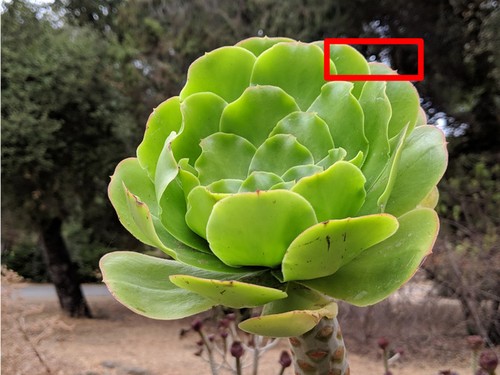}\label{subfig:dual_pixel_blur_rgb}}
    \subfigure[Depth from \cite{wadhwa2018}]{\includegraphics[width=0.15\textwidth]{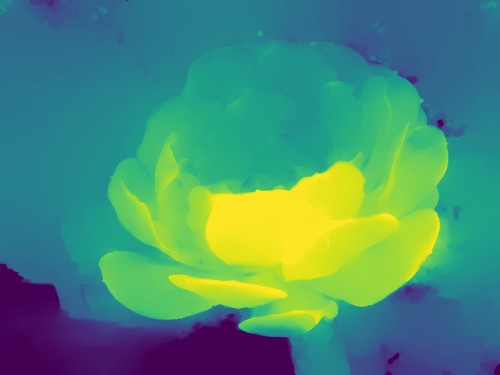}\label{subfig:dual_pixel_blur_pixel2}}
    \subfigure[Our depth]{\includegraphics[width=0.15\textwidth]{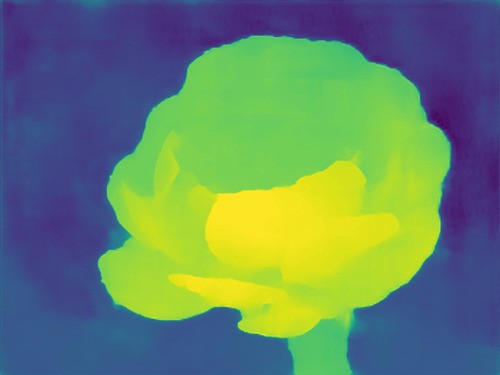}\label{subfig:dual_pixel_blur_our_depth}}
    \subfigure[Left DP view]{\includegraphics[width=0.14\textwidth]{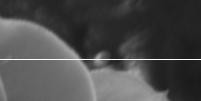}\label{subfig:dual_pixel_blur_left_dp}}
    \subfigure[Right DP view]{\includegraphics[width=0.14\textwidth]{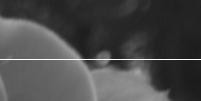}\label{subfig:dual_pixel_blur_right_dp}}
    \subfigure[Left vs Right intensity]{\includegraphics[width=0.18\textwidth]{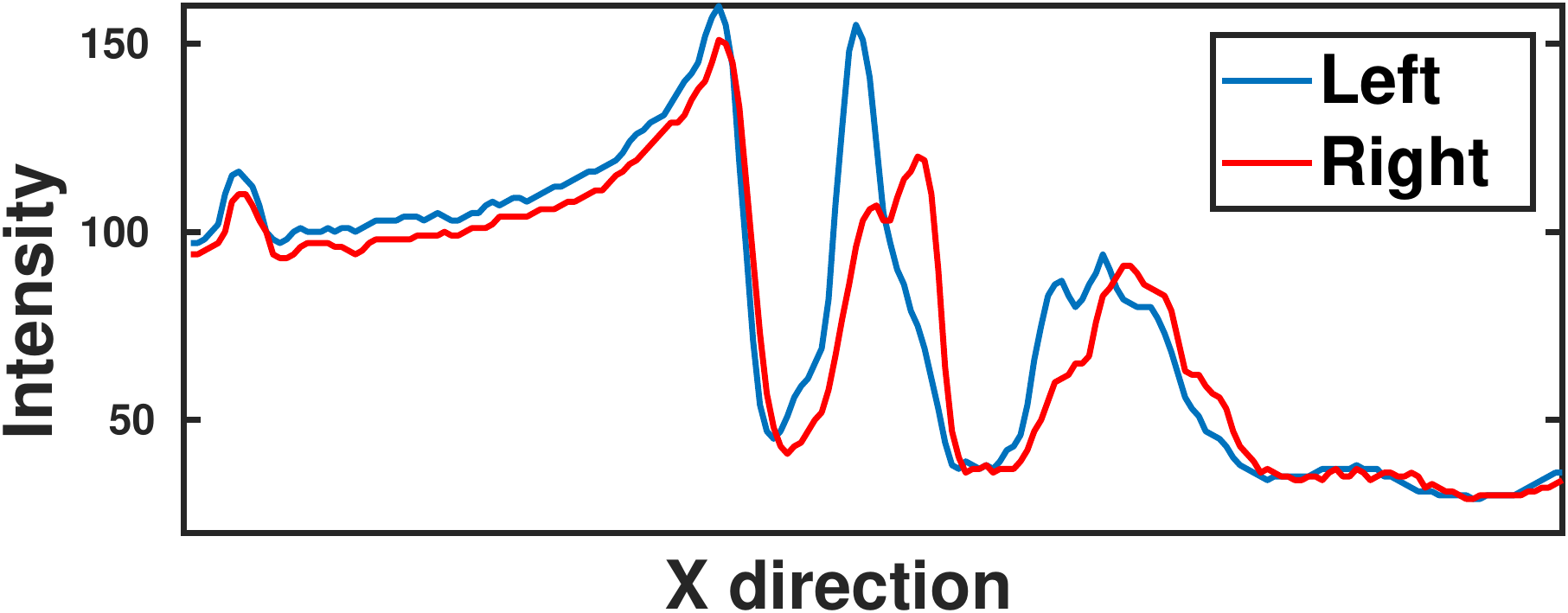}\label{subfig:dual_pixel_blur_plot}}
    \caption{Here we have an RGB image \subref{subfig:dual_pixel_blur_rgb} containing dual-pixel data.
    Crops of the left and right dual-pixel images corresponding to the marked rectangle in \subref{subfig:dual_pixel_blur_rgb} are shown in \subref{subfig:dual_pixel_blur_left_dp}, \subref{subfig:dual_pixel_blur_right_dp}, and their intensity profiles along the marked scanline are shown in \subref{subfig:dual_pixel_blur_plot}.
    While the profiles matches for the in-focus flower, they are considerably different for the out of focus background.
    Because \cite{wadhwa2018} uses traditional stereo matching that assumes that intensity values differ by only a scale factor and a local displacement, it fails to match the background accurately, and produces the depth shown in \subref{subfig:dual_pixel_blur_pixel2}.
    Our technique learns the correlation between depth and differences in dual-pixel data thereby estimates an accurate depth map \subref{subfig:dual_pixel_blur_our_depth}.
    }
    \label{fig:dual_pixel_blur}
\end{figure}

\begin{figure}[t]
    \centering
    \subfigure[Traditional Bayer Sensor]{
    \includegraphics[width=1.5in]{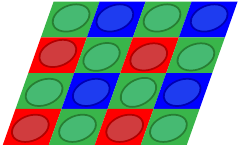}\label{subfig:sensor_bayer}}
    \subfigure[Dual-Pixel Sensor]{
    \includegraphics[width=1.5in]{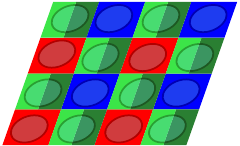}\label{subfig:sensor_dp}}
    \caption{A modern Bayer sensor consists of interleaved red, green, and blue pixels underneath a microlens array. \subref{subfig:sensor_bayer}.
    In dual-pixel sensors, the green pixel under each microlens is split in half \subref{subfig:sensor_dp},  resulting in two green images that act as a narrow-baseline stereo camera, much like a reduced light field camera.
    }
    \label{fig:sensor}
\end{figure}

Recent developments in consumer hardware may provide an opportunity for a new approach in depth estimation.
Cameras have recently become available that allow a single camera to simultaneously capture two images that resemble a stereo pair with a tiny baseline (Fig.~\ref{fig:dual_pixel_blur}), through the use of dense dual-pixel (DP) sensors (Fig.~\ref{fig:sensor}).
Though this technology was originally developed in service of camera auto-focus, dual-pixel images can also be exploited to recover dense depth maps from a single camera, thereby obviating any need for additional hardware, calibration, or synchronization.
For example, Wadhwa \etal \cite{wadhwa2018} used classical stereo techniques (block matching and edge aware smoothing) to recover depth from DP data.
But as shown in Fig.~\ref{fig:dual_pixel_blur}, the quality of depth maps that can be produced by conventional stereo techniques is limited, because the interplay between disparity and focus in DP imagery can cause classic stereo-matching techniques to fail.
Existing monocular learning-based techniques also perform poorly on this task. 
In this paper, we analyze the optics of image formation for dual-pixel imagery and demonstrate that DP images have a fundamentally ambiguous relationship with respect to scene depth --- depth can only be recovered up to some unknown affine transformation.
With this observation, we analytically derive training procedures and loss functions that incorporate prior knowledge of this ambiguity, and are therefore capable of learning effective models for affine-invariant depth estimation.
We then use these tools to train deep neural networks that estimate high-quality depth maps from DP imagery, thereby producing detailed and accurate depth maps using just a single camera.
Though the output of our learned model suffers from the same affine ambiguity that our training data does, the affine-transformed depths estimated by our model can be of great value in certain contexts, such as depth ordering or defocus rendering.

Training and evaluating our model requires large amounts of dual-pixel imagery that has been paired with ground-truth depth maps.
Because no such dataset exists, in this work we also design a capture procedure for collecting ``in the wild'' dual-pixel imagery where each image is paired with multiple alternate views of the scene.
These additional views allow us to train our model using view supervision, and allow us to use multi-view geometry to recover ground-truth estimates of the depth of the scene for use in evaluation.
When comparing against the state-of-the-art in depth estimation, our proposed model produces error rates that are $30\%$ lower than previous dual-pixel and monocular depth estimation approaches.

\section{Related Work}

Historically, depth estimation has seen the most attention and progress in the context of stereo \cite{scharstein2002taxonomy} or multi-view geometry \cite{Hartley2003}, in which multiple views of a scene are used to partially constrain its depth, thereby reducing the inherent ambiguity of the problem.
Estimating the depth of a scene from a single image is significantly more underconstrained, and though it has also been an active research area, progress has happened more slowly.
Classic monocular depth approaches relied on singular cues, such as shading \cite{horn1970shape}, texture \cite{bajcsy1976texture}, and contours \cite{Brady84} to inform depth, with some success in constrained scenarios.
Later work attempted to use learning to explicitly consolidate these bottom-up cues into more robust monocular depth estimation techniques \cite{Barron2012A, Hoiem2005, Saxena2005}, but progress on this problem accelerated rapidly with the rise of deep learning models trained end-to-end for monocular depth estimation \cite{eigen2014, DORN2018}, themselves enabled by the rise of affordable consumer depth sensors which allowed collection of large RGBD datasets \cite{Janoch2011, Menze2015KITTI, SilbermanECCV12NYUdv2}.
The rise of deep learning also yielded progress in stereoscopic depth estimation \cite{zbontar2015computing} and in the related problem of motion estimation \cite{FlowNet}.
The need for RGBD data in training monocular depth estimation models was lessened by the discovery that the overconstraining nature of multi-view geometry could be used as a supervisory cue for training such systems \cite{DeepStereo2016, garg2016unsupervised,godard2017,Jiang2018, MahjourianWA18, zhou2017unsupervised}, thereby allowing ``self-supervised'' training using only video sequences or stereo pairs as input.
Our work builds on these monocular and stereo depth prediction algorithms, as we
construct a learning-based ``stereo'' technique, but using the impoverished dual-pixel data present within a single image.

An alternative strategy to constraining the geometry of the scene is to vary the camera's focus. 
Using this ``depth from (de)focus'' \cite{Grossmann87} approach, depth can be estimated from focal stacks using classic vision techniques~\cite{Suwajanakorn15} or deep learning approaches\cite{hazirbas18ddff}.
Focus can be made more informative in depth estimation by manually ``coding'' the aperture of a camera~\cite{levin2007image}, thereby causing the camera's circle of confusion to more explicitly encode scene depth.
Focus cues can also be used as supervision in training a monocular depth estimation model \cite{Srinivasan2018}.
Reasoning about the relationship between depth and the apparent focus of an image is critical when considering dual-pixel cameras, as the effective point spread functions of the ``left'' and ``right'' views are different.
By using a flexible learning framework, our model is able to leverage the focus cues present in dual-pixel imagery in addition to the complementary stereo cues.

Stereo cameras and focal stacks are ways of sampling what Adelson and Bergen called ``the plenoptic function'': a complete record of the angle and position of all light passing through space~\cite{adelson1991plenoptic}. An alternative way of sampling the plenoptic function is a light field \cite{Levoy1996}, a 4D function that contains conventional images as 2D slices. Light fields can be use to directly synthesize images from different positions or with different aperture settings \cite{Ng2005fourier}, and light field cameras can be made by placing a microlens array on the sensor of a conventional camera~\cite{adelson1992single, ng2005lightcamera}. Light fields provide a convenient framework for analyzing the equivalence of correspondence and focus cues \cite{tao2013depth}. While light fields have been used to recover depth \cite{jeon2015depth,jeon2019depth}, constructing a light field camera requires sacrificing spatial resolution in favor of angular resolution, and as such light field cameras have not seen rapid consumer adoption. Dual-pixel cameras appear to represent a promising compromise between more ambitious light field cameras and conventional cameras --- DP cameras sacrifice a negligible amount of spatial resolution to sample two angles in a light field, while true monocular cameras sample only a single angle, and light field cameras such as the Lytro Illum sample 196 angles at the cost of significant spatial resolution. As a result, they have seen wider adoption in consumer cameras and in space-constrained applications like endoscopy \cite{Sam2016}.

\section{Dual-Pixel Geometry and Ambiguity}

\begin{figure}
    \centering
    \includegraphics[width=\columnwidth]{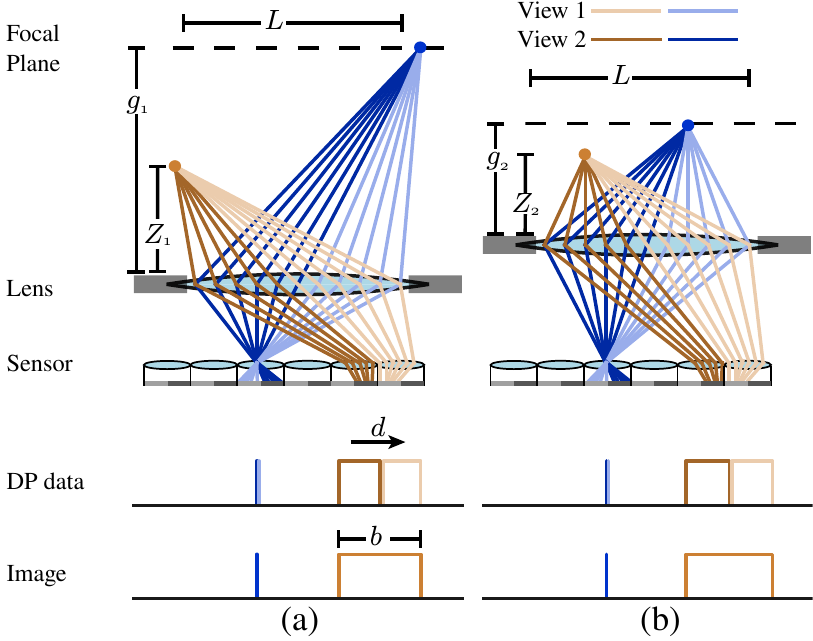}
    \caption{Dual-pixel views see different halves of the aperture, which provides a depth cue.
    However, due to a fundamental ambiguity, different scenes can yield the same dual-pixel images if the focus distance (or the aperture size, or the focal length) of the camera changes.
    In (a), a camera with focus distance $\focusdistance_1$ images an in-focus blue point and an out-of-focus orange point a distance $\worlddistance_1$ away.
    Light refracting through the left half of the aperture (dark blue and orange rays) arrives at the right half of each dual-pixel, and vice versa.
    This results in a dual-pixel image in which the out-of-focus orange point is displaced by $d$ pixels (a, ``DP Data'') and blurred by $b$ pixels (a, ``Image'').
    In (b), a different focus distance and set of scene depths yields the same dual-pixel and RGB images. However, as shown in the text, this scene is related to the one in (a) by an affine transformation on inverse depth.
    }
    \label{fig:ambiguity_diagram}
\end{figure}

Dual-pixel (DP) sensors work by splitting each pixel in half, such that the left half integrates light over the right half aperture and vice versa (Fig.~\ref{fig:ambiguity_diagram}).
Because each half of a dual-pixel integrates light over one half of the aperture, the two halves of a pixel together form a kind of stereo pair, in which nearby objects exhibit some horizontal disparity between the two views in accordance with their distance.
This effect interacts with the optical blur induced by the lens of the camera, such that when image content is far from the focal plane, the effects of optical blur are spread across the two ``views'' of each dual-pixel (Fig.~\ref{fig:ambiguity_diagram}(a, DP data)).
The sum of the two views accounts for all the light going through the aperture and is equal to the ordinary full-pixel image that would be captured by a non dual-pixel sensor.
As a result, the disparity $d$ between the two views in a dual-pixel image is proportional to what the defocus blur size $b$ would be in an equivalent full-pixel image.
Dual-pixel sensors are commonly used within consumer cameras to aid in auto-focus: the camera iteratively estimates disparity from the dual-pixels in some focus region and moves the lens until that disparity is zero, resulting in an image in which the focus region is in focus.

While dual-pixel imagery can be thought of as a stereo pair with a tiny baseline, it differs from stereo in several ways. The views are perfectly synchronized (both spatially and temporally) and have the same exposure and white balance.
In addition, the two views in DP images have different point-spread functions that can encode additional depth information.
Traditional stereo matching techniques applied to dual-pixel data will not only ignore the additional depth information provided by focus cues, but may even fail in out-of-focus regions due to the effective PSFs of the two views being so different that conventional image matching fails (Figs.~\ref{subfig:dual_pixel_blur_left_dp}-\ref{subfig:dual_pixel_blur_plot}).
As an additional complication, the relationship between depth and disparity in dual-pixel views depends not only on the baseline between the two views, but also on the focus distance.
Thus, unlike depth from stereo, which has only a scale ambiguity if the extrinsics are unknown, depth from dual-pixel data has \emph{both} scale and offset ambiguities if the camera's focus distance is unknown (as is the case for most current consumer cameras, such as those we use).
Addressing the ambiguity caused by this unknown scale and offset is critical when learning to estimate depth from dual-pixel imagery, and is a core contribution of this work.
As we will demonstrate, for a network to successfully learn from dual-pixel imagery, it will need to be made aware of this affine ambiguity.

We will now derive the relationship between depth, disparity, and blur size according to the paraxial and thin-lens approximations.
Consider a scene consisting of point light sources located at coordinates $(x, y, \worlddistance(x, y))$ in camera space.
As stated previously, the disparity of one such point on the image plane $d(x, y)$ is proportional to the (signed) blur size $\bar b(x, y)$, where the sign is determined by whether the light source is in front or behind the focal plane.
Therefore, from the paraxial and thin-lens approximations:
\begin{align}
    d(x, y) &= \alpha \bar b(x, y) \\
      & \approx \alpha \frac{\aperturesize f}{1-f/\focusdistance}\left(\frac{1}{\focusdistance} - \frac{1}{\worlddistance(x, y)}\right) \\
      & \triangleq A(\aperturesize, f, \focusdistance) + \frac{B(\aperturesize, f, \focusdistance)}{\worlddistance(x, y)}, \label{eq:lens_blur}
\end{align}
where $\alpha$ is a constant of proportionality, $\aperturesize$ is the diameter of the aperture, $f$ is the focal length of the lens and $\focusdistance$ is the focus distance of the camera.
We make the affine relationship between inverse depth and disparity explicit in Eqn.~\ref{eq:lens_blur} by defining image-wide constants $A(\aperturesize, f, \focusdistance)$ and $B(\aperturesize, f, \focusdistance)$.
This equation reflects our previous assertion that perfect knowledge of disparity $d$ and blur size $b$ only gives enough information to recover depth $\worlddistance$ if the parameters $\aperturesize$, $f$ and $\focusdistance$ are known. 
Please see the supplement for a derivation.

Eqn.~\ref{eq:lens_blur} demonstrates the aforementioned affine ambiguity in dual-pixel data. This means that different sets of camera parameters and scene geometries can result in identical dual-pixel images (Fig.~\ref{fig:ambiguity_diagram}(b)).  Specifically, two sets of camera parameters can result in two sets of affine coefficients $(A_1, B_1)$ and $(A_2, B_2)$ such that the same image-plane disparity is produced by two different scene depths
\begin{equation} d(x, y) = A_1 +\frac{B_1}{\worlddistance_1(x, y)}  = A_2 + \frac{B_2}{\worlddistance_2(x, y)}.
\end{equation}
Consumer smartphone cameras are not reliable in recording camera intrinsic metadata~\cite{Diverdi2016}, thereby eliminating the easiest way that this ambiguity could be resolved.
But Eqn.~\ref{eq:lens_blur} does imply that it is possible to use DP data to estimate some (unknown) affine transform of inverse depth.
This motivates our technique of training a CNN to estimate inverse depth only up to an affine transformation.

Though absolute depth would certainly be preferred over an affine-transformed depth, the affine-transformed depth that can be recovered from dual-pixel imagery is of significant practical use. Because affine transformations are monotonic, an affine-transformed depth still allows for reasoning about relative ordering of scene depths. Affine-invariant depth is a natural fit for synthetic defocus (simulating wide aperture images by applying a depth dependent blur to a narrow aperture image~\cite{barron2015fast, wadhwa2018}) as the affine parameters naturally map to the user controls --- the depth to focus at, and the size of the aperture to simulate. Additionally, this affine ambiguity can be resolved using heuristics such as the likely sizes of known objects~\cite{hoiemcvpr06}, thereby enabling the many uses of metric depth maps.

\section{View supervision for Affine Invariant Depth}

A common approach for training monocular depth estimation networks from multi-view data is to use self supervision. This is typically performed by warping an image from one viewpoint to the other according to the estimated depth and then using the difference between the warped image and the actual image as some loss to be minimized. Warping is implemented using a differentiable spatial transformer layer \cite{STN2015} that allows end-to-end training using only RGB views and camera poses.
Such a loss can be expressed as:
\begin{equation}
\lossfun(\image{0}, \Theta) = \sum_{(x, y)} \deltafun{I_0(x, y), \image{1} \left(M \left(x, y; F\left(I_0, \Theta \right)\right)\right)}
\label{eqn:view_sup_stereo}
\end{equation}
Where $\image{0}$ is the RGB image of interest,
$\image{1}$ is a corresponding stereo image, $F(\image{0}, \Theta)$ is the (inverse) depth estimated by a network for $\image{0}$, $M(x, y; \preddepth)$ is the warp induced on pixel coordinates $(x, y)$ by that estimated depth $\preddepth=F(\image{0}, \Theta)$ and by the known camera poses, and $\deltafun{\cdot, \cdot}$ is some arbitrary function that scores the per-pixel difference between two of RGB values.
$\deltafun{\cdot, \cdot}$ will be defined in Sec.~\ref{sec:training}, but for our current purposes it can be any differentiable penalty.
Because we seek to predict inverse depth up to an unknown affine transform, the loss in Eqn.~\ref{eqn:view_sup_stereo} cannot be directly applied to our case. Hence, we introduce two different methods of training with view supervision while predicting inverse depth up to an affine ambiguity.

\subsection{3D Assisted Loss}
\label{sec:3d_loss}

If we assume that we have access to a ground truth inverse depth $\depth$ and corresponding per-pixel confidences $\depthconf$ for that depth, we can find the unknown affine mapping by solving
\begin{equation}
\resizebox{\linewidth}{!}{
    $\displaystyle
    \argmin_{a, b} \sum_{(x,y)} \depthconf(x,y) \left( \depth(x,y) - \left(a F\left(\image{0}, \Theta \right)(x,y) + b \right) \right)^2
    $}
    \label{eqn:lst_sqr}
\end{equation}
While training our model $\Theta$, during each evaluation of our loss we solve Eqn.~\ref{eqn:lst_sqr} using a differentiable least squares solver (such as the one included in TensorFlow) to obtain $a$ and $b$, which can be used to obtain absolute depth that can then be used to compute a standard view supervision loss.
Note that since we only need to solve for two scalars, a sparse ground truth depth map with a few confident depth samples suffices.

\subsection{Folded Loss}
\label{sec:folded_loss}

Our second strategy does not require ground truth depth and folds the optimization required to solve the affine parameters into the overall loss function. We associate variables $a$ and $b$ with each training example $\image{0}$ and define our loss function as:
\begin{equation}
\resizebox{\linewidth}{!}{
    $\displaystyle
    \lfold(\image{0}, \Theta, a, b) = \sum_{(x,y)} \deltafun{\image{0}(x,y), \image{1}\left( M \left(x, y; a F \left(\image{0}, \Theta \right) + b \right) \right)}
    $} \label{eq:folded_indiv_loss}
\end{equation}
and then let the gradient descent optimize for $\Theta$, $\{\asubi\}$ and $\{\bsubi\}$ by solving
\begin{equation}
\argmin_{\Theta, \{\asubi\}, \{\bsubi\} } \sum_i \lfold(\image{0}^{(i)}, \Theta, \asubi, \bsubi).
\label{eqn:folded_loss}
\end{equation}
To avoid degeneracies as $\asubi$ approaches zero, we parameterize $\asubi=\epsilon + \log(\exp(\latentasubi) + 1)$ where $\epsilon=10^{-5}$. We initialize $\{ \latentasubi \}$ and $\{\bsubi\}$ from a uniform distribution in $[-1,1]$.
To train this model, we simply construct one optimizer instance in which $\Theta$, $\{ \latentasubi \}$, and $\{\bsubi\}$ are all treated as free variables and optimized over jointly.

\section{Data Collection}
\begin{figure}
    \centering
    \subfigure[Our capture rig]{\includegraphics[width=0.15\textwidth]{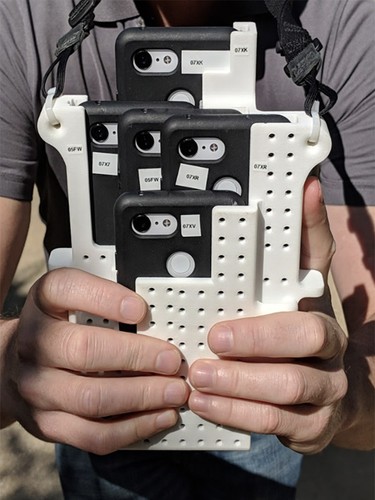}\label{subfig:rig_1}}
    \subfigure[COLMAP depth]{\includegraphics[width=0.15\textwidth]{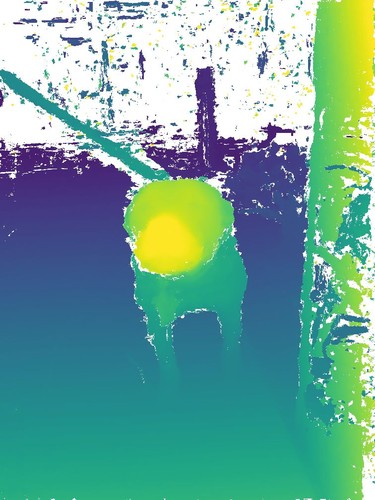}\label{subfig:rig_depth1}}
    \subfigure[Our depth]{\includegraphics[width=0.15\textwidth]{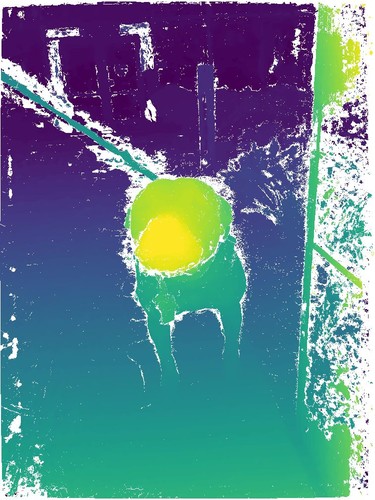}\label{subfig:rig_depth2}}
    \subfigure[An example capture]{\includegraphics[width=0.47\textwidth]{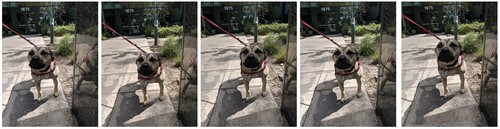}\label{subfig:rig_capture}} 
    \caption{Our portable capture rig with synchronized cameras \subref{subfig:rig_1} can be used to capture natural in-the-wild photos, where each central image is accompanied with 4 additional views \subref{subfig:rig_capture}. These multiple views allow us to use multi-view stereo algorithms to compute ``ground truth'' depths and confidences, as shown in \subref{subfig:rig_depth1} and \subref{subfig:rig_depth2} (low confidence depths are rendered as white).}
    \label{fig:rig}
\end{figure}

To train and evaluate our technique, we need dual-pixel data paired with ground-truth depth information. We therefore collected a large dataset of dual-pixel images captured in a custom-made capture rig in which each dual-pixel capture is accompanied by 4 simultaneous images with a moderate baseline, arranged around the central camera (Figure~\ref{subfig:rig_1}).
We compute ``ground truth'' depths by applying established multi-view geometry techniques to these 5 images.
These depths are often incomplete compared to those produced by direct depth sensors, such as the Kinect or LIDAR. However, such sensors can only image certain kinds of scenes --- the Kinect only works well indoors, and it is difficult to acquire LIDAR scans of scenes that resemble normal consumer photography. Synchronization and registration of these sensors with the dual-pixel images is also cumbersome.
Additionally, the spatial resolutions of direct depth sensors are far lower than the resolutions of RGB cameras.
Our approach allows us to capture a wide variety of high-resolution images, captured both indoors and outdoors, that resemble what people typically capture with their cameras: pets, flowers, etc (we do not include images of faces in our dataset, due to privacy concerns).
The plus-shaped arrangement means that it is unlikely that a pixel in the center camera is not visible in at least one other camera (barring small apertures or very nearby objects) thereby allowing us to recover accurate depths even in partially-occluded regions.
The cameras are synchronized using the system of \cite{SoftwareSync2019}, thereby allowing us to take photos from all phones at the same time (within $\sim \!16$ milliseconds, or half a frame) which allows us to reliably image moving subjects.
Though the inherent difficulty of the aperture problem means that our ground-truth depths are rarely perfect, we are able to reliably recover high-precision \emph{partial} depth maps, in which high-confidence locations have accurate depths and inaccurate depths are flagged as low-confidence (Figures~\ref{subfig:rig_depth1}, \ref{subfig:rig_depth2}).
To ensure that our results are reliable and not a function of some particular stereo algorithm, we compute two separate depth maps (each with an associated confidence) using two different algorithms: the established COLMAP stereo technique~  \cite{SchonbergerMvs16, COLMAP}, and a technique we designed for this task. See the supplement for a detailed description.

Our data is collected using a mix of two widely available consumer phones with dual-pixels: The Google Pixel 2 and the Google Pixel 3. 
For each capture, all 5 images are collected using the same model of phone.
We captured $3{,}573$ scenes resulting in $3{,}573\times 5 = 17,865$ RGB and DP images.
Our photographer captured a wide variety of images that reflect the kinds of photos people take with their camera, with a bias towards scenes that contain interesting nearby depth variation, such as a subject that is 0.5 - 2 meters away.
Though all images contain RGB and DP information, for this work we only use the DP signal of the center camera. All other images are treated as conventional RGB images. We process RGB and DP images at a resolution of $1512 \times 2016$, but compute ``ground truth'' depth maps at half this resolution to reduce noise.
We use inverse perspective sampling in the range 0.2 - 100 meters to convert absolute depth to inverse depth $\depth$.
Please see the supplement for more details.

Though our capture rig means that the relative positions of our 5 cameras are largely fixed, and our synchronization means that our sets of images are well-aligned temporally, we were unable to produce a single fixed intrinsic and extrinsic calibration of our camera rig that worked well across all sets of images.
This is likely due to the lens not being fixed in place in the commodity smartphone cameras we use. As a result, the focus may drift due to mechanical strain or temperature variation, the lens may jitter off-axis while focusing, and optical image stabilization may move the camera's center of projection \cite{Diverdi2016}.
For this reason, we use structure from motion \cite{Hartley2003} with priors provided by the rig design to solve for the extrinsics and intrinsics of the 5 cameras individually for each capture, which results in an accurate calibration for all captures. This approach introduces a variable scale ambiguity in the reconstructed depth for each capture, but this is not problematic for us as our training and evaluation procedures assume an unknown scale ambiguity.

\section{Experiments}

We describe our data, evaluation metrics and method of training our CNN for depth prediction. In addition, we compare using affine-invariant losses to using scale-invariant and ordinary losses and demonstrate that affine-invariant losses improve baseline methods for predicting depth from dual-pixel images.

\subsection{Data Setup}

\begin{table}
    \centering
    \begin{tabular}{l|ccc}
    &  $\AIWE(1)$ & $\AIWE(2)$ & $1 - \abs{\rho_s}$ \\
    \hline
    Folded Loss & .0225 & .0318 & .195 \\
    3D Assisted Loss & \textbf{.0175} & \textbf{.0264} & \textbf{.139} \\
    \end{tabular}
    \caption{
    Accuracy of DPNet trained with two different methods.
    Our ``3D Assisted Loss'', which has access to ground truth depth to fully-constrain the ambiguity, tends to outperform the alternative approach of our ``Folded Loss'', which circumvents the lack of known depth by folding an optimization problem into the computation of the loss function during training.
    }
    \label{table:3D_vs_folded}
\end{table}

\definecolor{Yellow}{rgb}{1,1, 0.6}

\begin{table*}[h]
    \centering
    \resizebox{\linewidth}{!}{
    \begin{tabular}{l|c||ccc|ccc|c}
    \multirow{2}{*}{Method} & \multirow{2}{*}{Invariance} & \multicolumn{3}{c|}{Evaluated on Our Depth} & \multicolumn{3}{c|}{Evaluated on COLMAP Depth} & Geometric\\
    &  & $\AIWE(1)$ & $\AIWE(2)$ & $1 - \abs{\rho_s}$ & $\AIWE(1)$ & $\AIWE(2)$ & $1 - \abs{\rho_s}$ & Mean \\
    \hline
    \rowcolor[gray]{.95} \multicolumn{9}{c}{RGB Input} \\
    \hline
    \multirow{3}{*}{DPNet}
    & None   &     .0602 &     .0754 &     .631 &     .0607 &     .0760 &     .652 & .1432\\
    & Scale  &     .0409 &     .0544 &     .490 &     .0419 &     .0557 &     .514 & .1047\\
    & Affine &  \bf .0398 &  \bf .0530 & \bf .464 & \bf .0410 &  \bf .0546 & \bf  .493 & \bf .1014\\
    \hline 
    \multicolumn{2}{l||}{DORN \cite{DORN2018} (NYUDv2 model)} & .0421 & .0555 &  .407 &  .0426 & .0557 &  .419 &  .0990\\
    \hline
    \multicolumn{2}{l||}{DORN \cite{DORN2018} (KITTI model)}  & .0490 & .0631 & .549 &  .0492 & .0630 &  .558 & .1196\\
    
    \multicolumn{9}{c}{\,} \\
    \hline
    \rowcolor[gray]{.95} \multicolumn{9}{c}{RGB + DP Input} \\
    \hline
    \multirow{3}{*}{DPNet}
    & None  & .0581 & .0735 & .827 & .0587 & .0742 & .834 & .1530\\
    & Scale  & .0202 & .0295 & .162 & .0213 & .0322 & .178 & .0477 \\
    & Affine  & \cellcolor{Yellow} \bf .0175 & \cellcolor{Yellow} \bf .0264 & \cellcolor{Yellow} \bf .139 &  \cellcolor{Yellow} \bf .0190 & \cellcolor{Yellow} \bf .0298 & \cellcolor{Yellow} \bf .156 & \cellcolor{Yellow} \bf .0422\\
    \hline
    \multirow{3}{*}{VGG}
    & None & .0370 & .0492 & .350 & .0383 & .0513 & .360 & .0876 \\
    & Scale  & .0224 & .0321 & .181 & .0242 & .0356 & .208 & .0535 \\
    & Affine & \bf .0186 & \bf .0275 & \bf .149 & \bf  .0202 & \bf .0308 & \bf .166 & \bf .0446\\
    \hline
    \multirow{3}{2.5cm}{Godard \etal \cite{godard2017} \\ (ResNet50)} 
     & None$^\dagger$ & .0562 & .0714 & .738 &  .0568 & .0720 & .745 & .1442\\
     & Scale$^\dagger$ & .0260 & .0367 & .227 & .0270 & .0383 & .239 & .0613\\
     & Affine$^\dagger$ & \bf .0251 & \bf .0356 & \bf .222 & \bf .0257 & \bf .0366 & \bf .232 & \bf .0592\\
    \hline
    \multirow{3}{2.5cm}{Garg \etal \cite{garg2016unsupervised} \\ (ResNet50)}
     & None & .0571 & .0722 & .761 & .0577 & .0728 & .772 & .1472\\
     & Scale$^\dagger$ & .0261 & .0369 & .228 & .0267 & .0382 & .237 & .0613\\
     & Affine$^\dagger$ & \bf .0248 & \bf .0352 & \bf .216 & \bf .0255 & \bf .0365 & \bf .227 & \bf .0584\\
    \hline
    \multicolumn{2}{l||}{Wadhwa \etal \cite{wadhwa2018}} & .0270 & .0375 & .236 &  .0276 & .0388 & .245 & .0630\\
    \end{tabular}
    }
    \caption{
    Accuracy of different models and approaches evaluated on our depth and COLMAP depth with the right-most column containing the geometric mean of all the metrics.
    For models trained with different degrees of invariance, the best-performing invariance's score is bolded. The overall best-performing technique is highlighted in yellow. A $\dagger$ indicates that we use only DP images as input to a model, which we do if it produces better results compared to using RGB+DP input.}
    \label{table:losses_comparison}
\end{table*}

Following the procedure of \cite{wadhwa2018}, we center crop our dual-pixel images to $66.67\%$ of the original resolution to avoid spatially varying effects in dual-pixel data towards the periphery, and to remove the need for radial distortion correction.
We do not downsample the dual-pixel images, as doing so would destroy the subtle disparity and blur cues they contains.
After cropping, the input to our network is of resolution $1008 \times 1344$ while the output is $504 \times 672$, i.e., the same resolution as our ground truth depth.
Our evaluation metrics are computed on a center crop of the output of size $384 \times 512$, as the center of the image is where our additional stereo views are most likely to overlap with the center view.
We randomly split our data into train and test sets, under the restriction that all images from each capture session are contained entirely within one of the two sets.
Our training and test sets contains $2{,}757$  and $718$ images respectively.
During training we use only our own ground-truth depth, though we evaluate on both our depth and COLMAP's depth.
COLMAP's SfM failed to converge on $47$ images in our test set, so we report the mean error of the remaining $671$ images.

\subsection{Training a Neural Net for Depth Prediction}
\label{sec:training}

Now that we have defined our loss function and our dataset, we can construct a neural network architecture for our task for predicting depth from dual-pixel and RGB images.
We use both the VGG model architecture similar to \cite{godard2017} and a lightweight network (DPNet) similar to a U-Net \cite{Ronneberger2015} with residual blocks \cite{resent2016}.
While the VGG model has $\sim19.8$ million parameters and $\sim295$ billion flops per inference, the DPNet has only $\sim0.24$ million parameters and $\sim5.5$ billion flops.
The architectures are detailed in the supplement.

For our difference $\deltafun{\image{0}, \image{j}}$ between the source image $\image{0}$ and the warped image $\image{j}$ from the $j^{th}$ neighbor, we use a weighted combination of a DSSIM loss and a Charbonnier loss with weights set to $0.8$ and $0.2$ respectively. Our DSSIM loss is the same as that of \cite{godard2017}: a window of size $3 \times 3$, with $c_1=0.01^2$ and $c_2=0.03^2$. The Charbonnier loss is computed by setting $\alpha=1$ and $c=0.1$ in the parametrization described in \cite{Barron19RobustLoss}. Images are normalized to $[0,1]$ range and the losses are computed on three channel RGB images with the losses per channel averaged together. Similar to \cite{godard2017,zhou2017unsupervised}, we predict depths at multiple resolutions (5 for DPNet and 3 for VGG), each scaled down a factor of 2, and aggregate losses across them.

To adapt the view supervision loss for stereo images (Eqn. \ref{eqn:view_sup_stereo}) to multi-view data, we use the approach of \cite{Godard2018} and compute $\deltafun{\image{0}, \image{j}}$ for each neighbor  and then take per-pixel minimum using the heuristic that a pixel must be visible in at least one other view, which applies for our case since the neighboring views surround the center view in the capture rig.

Our implementation is in Tensorflow \cite{tensorflow2015-whitepaper} and trained using Adam \cite{kingma:adam} with a learning rate of $0.001$ for $2$ million steps with a batch size of $4$ for the lightweight model and $2$ for the VGG model.
Our model weights are initialized randomly using Tensorflow's default initialization \cite{Glorot2010}.
We perform data augmentation by applying uniformly random translations to our imagery, limiting maximum translation in either direction to $10$ pixels.

\subsection{Evaluation Metrics}

The optics of dual-pixel cameras means that that we should not expect the depth estimated from dual-pixel imagery to be accurate in absolute terms --- at best, it should be accurate up to some unknown affine transformation.
This ambiguity prohibits the use of conventional metrics (such as those used by the Middlebury Stereo benchmark~\cite{scharstein2002taxonomy}) for evaluating the depth maps estimated from dual-pixel imagery, and requires that we construct metrics that are invariant to this ambiguity.

Instead, we use a weighted-variant of Spearman's rank correlation $\rho_s$, which evaluates the ordinal correctness of the estimated depth with ground truth depth confidences as weight. In addition, we use affine invariant weighted versions of MAE and RMSE, denoted AIWE(1) and AIWE(2) respectively. Please see the supplemental for details.

\subsection{Folded Loss vs 3D Assisted Loss}

Our first experiment is to investigate which of our two proposed solutions for handling affine invariance during training performs best.
Training our DPNet with both approaches, as shown in Table \ref{table:3D_vs_folded}, shows that the 3D assisted loss (Sec.~\ref{sec:3d_loss}) converges to a better solution than the folded loss (Sec.~\ref{sec:folded_loss}).
We therefore use our 3D assisted loss in all of the following experiments.

\subsection{Comparison to Other Methods}

\newcommand{\resultswidth}{0.135\textwidth}{
\newcommand{\fakecaption}[1]{\raisebox{0.5 em}{\footnotesize #1}}
\begin{figure*}
    \centering
    \begin{tabular}{@{}c@{\,\,}c@{\,\,}c@{\,\,}c@{\,\,}c@{\,\,}c@{\,\,}c@{}}
    \includegraphics[width=\resultswidth]{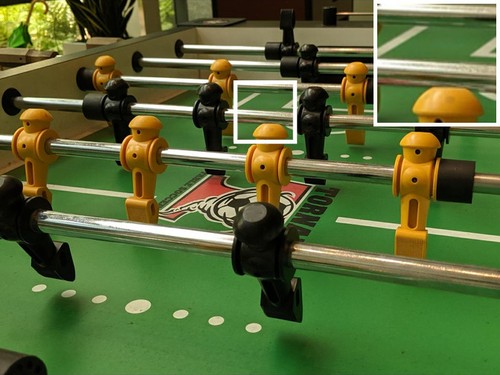} &
    \includegraphics[width=\resultswidth]{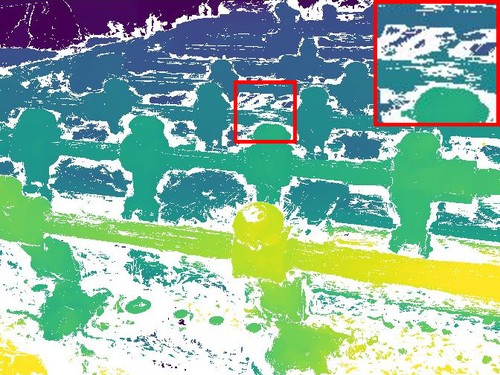} & 
    \includegraphics[width=\resultswidth]{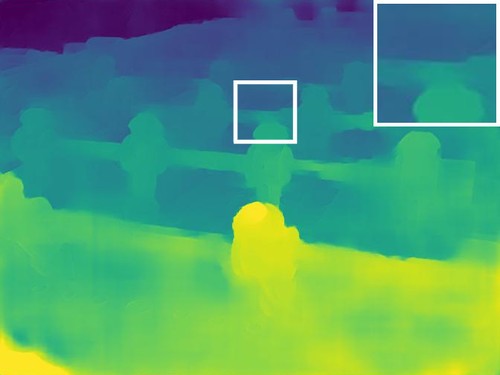} &
    \includegraphics[width=\resultswidth]{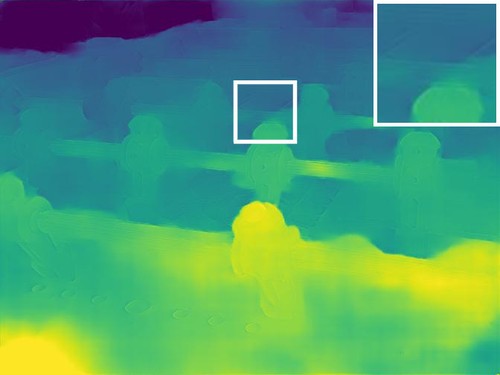} &
    \includegraphics[width=\resultswidth]{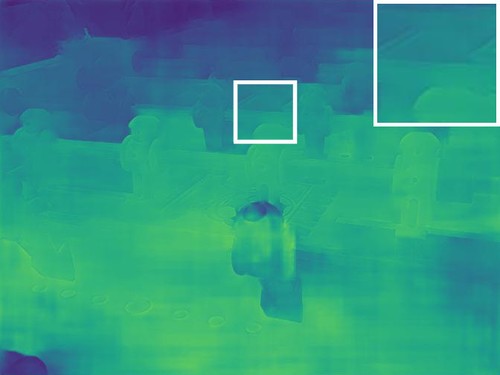} &
    \includegraphics[width=\resultswidth]{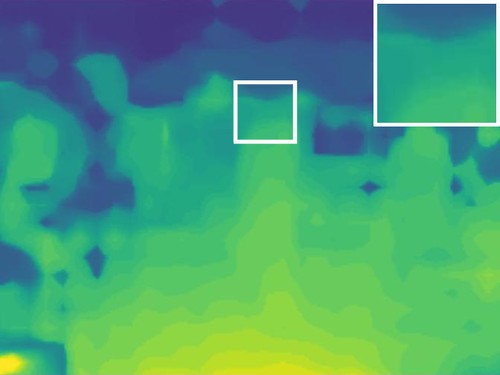} &
    \includegraphics[width=\resultswidth]{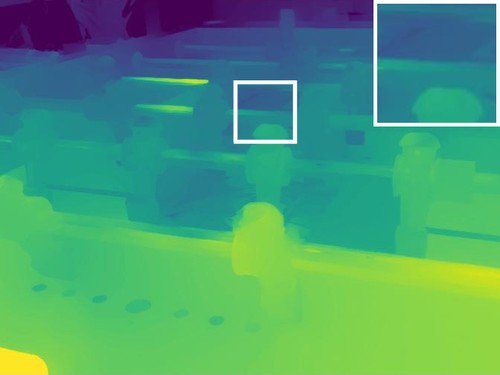} \\
    \includegraphics[width=\resultswidth]{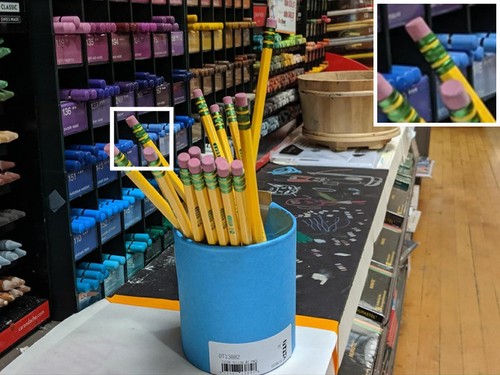} &
    \includegraphics[width=\resultswidth]{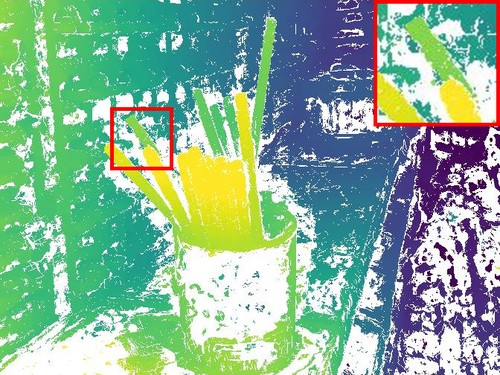} & 
    \includegraphics[width=\resultswidth]{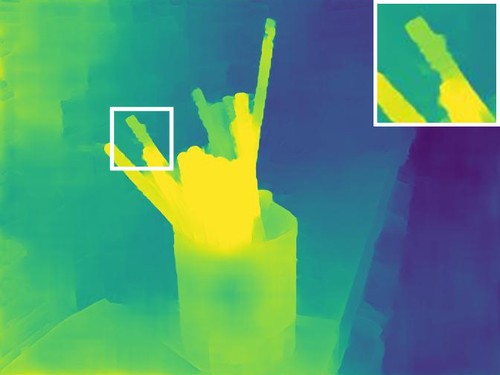} &
    \includegraphics[width=\resultswidth]{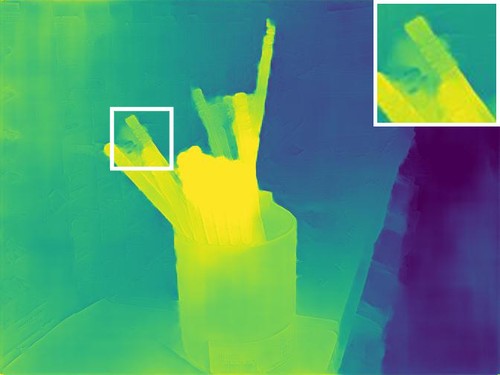} &
    \includegraphics[width=\resultswidth]{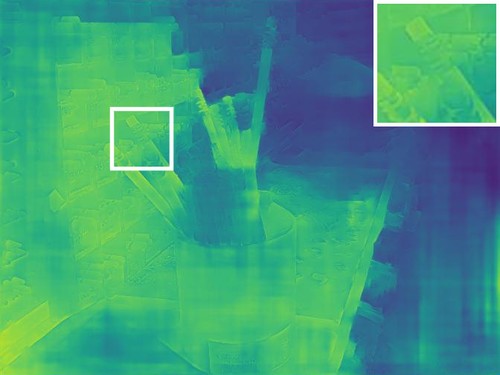} &
    \includegraphics[width=\resultswidth]{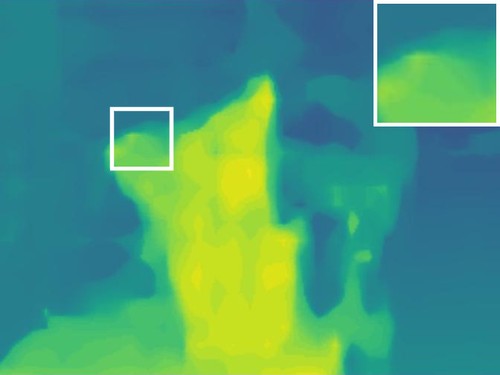} &
    \includegraphics[width=\resultswidth]{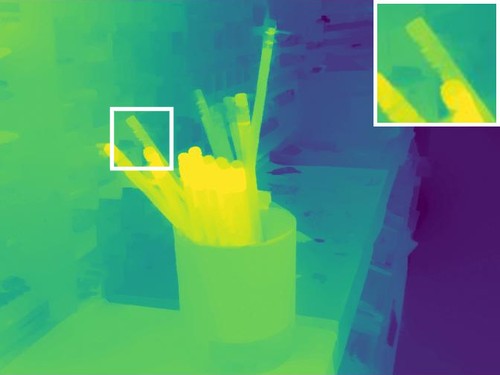} \\
    \includegraphics[width=\resultswidth]{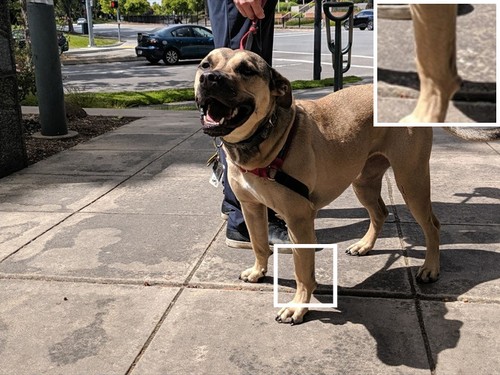} &
    \includegraphics[width=\resultswidth]{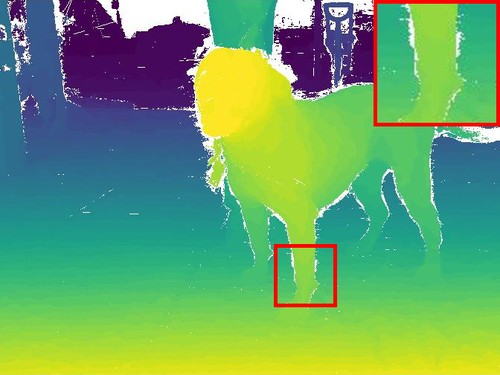} & 
    \includegraphics[width=\resultswidth]{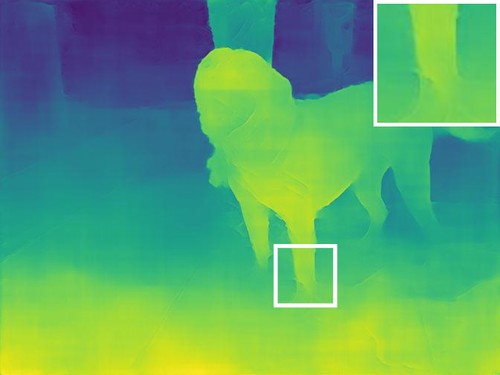} &
    \includegraphics[width=\resultswidth]{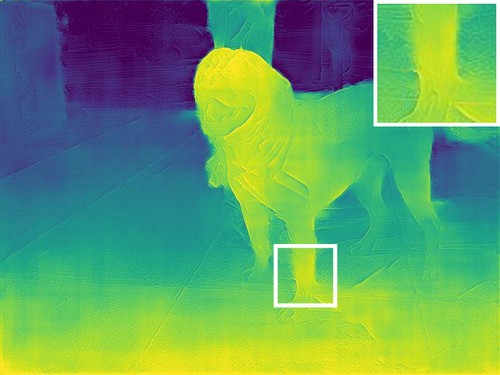} &
    \includegraphics[width=\resultswidth]{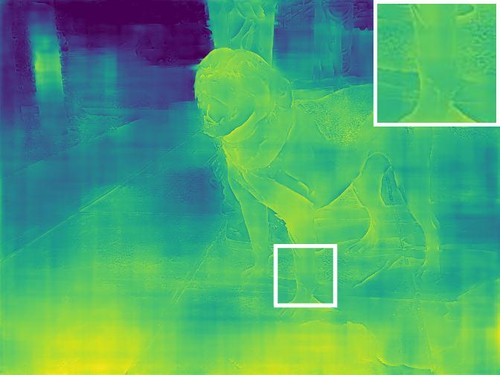} &
    \includegraphics[width=\resultswidth]{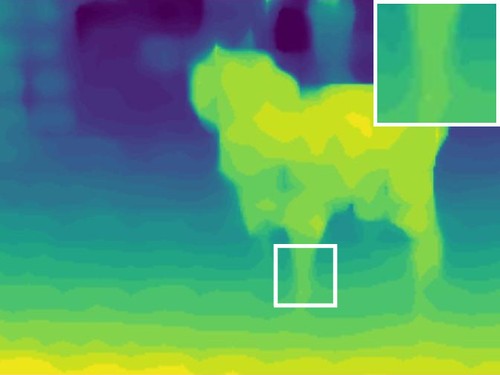} &
    \includegraphics[width=\resultswidth]{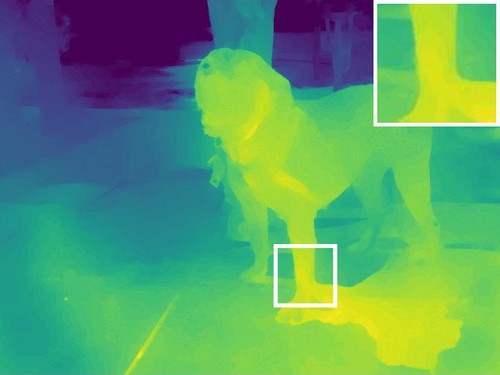} \\
    \includegraphics[width=\resultswidth]{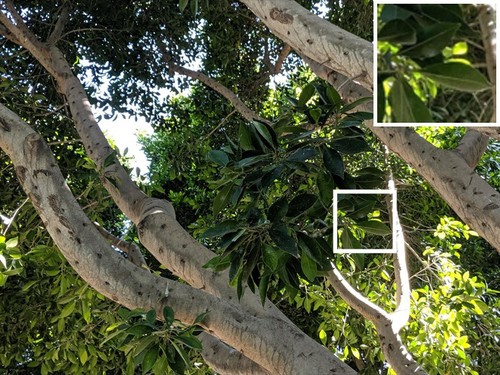} &
    \includegraphics[width=\resultswidth]{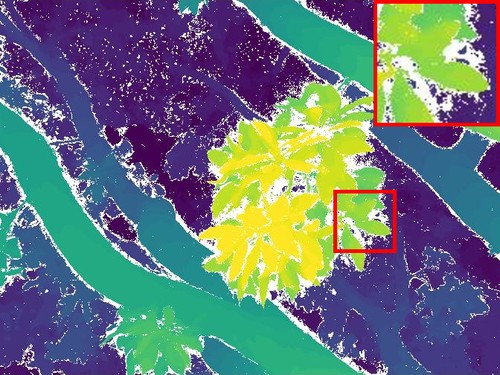} & 
    \includegraphics[width=\resultswidth]{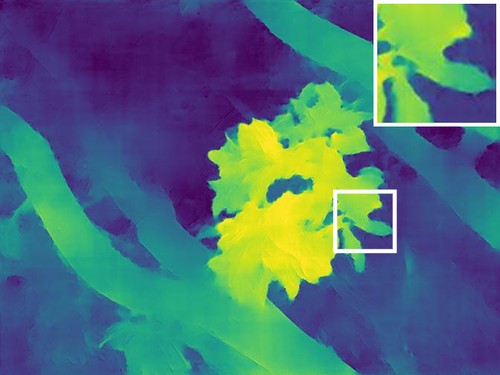} &
    \includegraphics[width=\resultswidth]{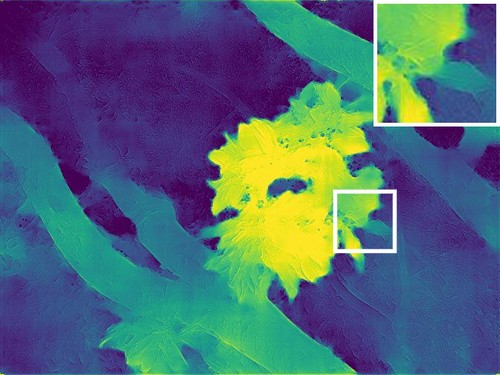} &
    \includegraphics[width=\resultswidth]{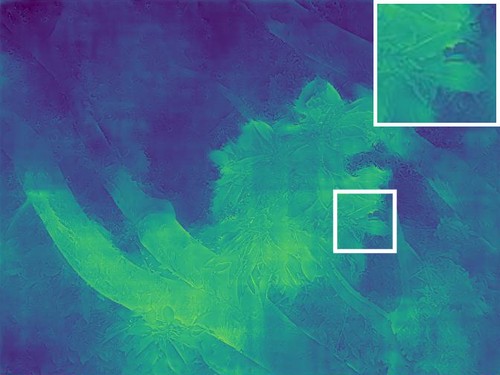} &
    \includegraphics[width=\resultswidth]{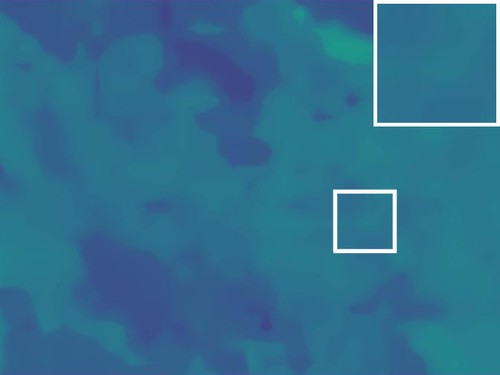} &
    \includegraphics[width=\resultswidth]{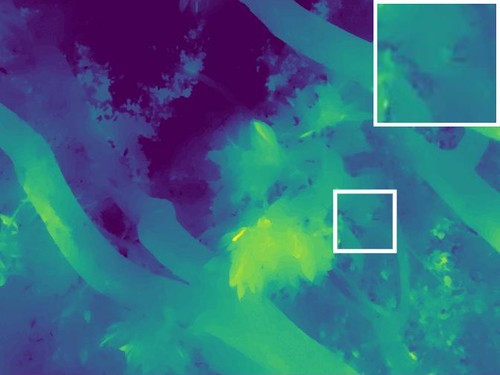} \\
    \includegraphics[width=\resultswidth]{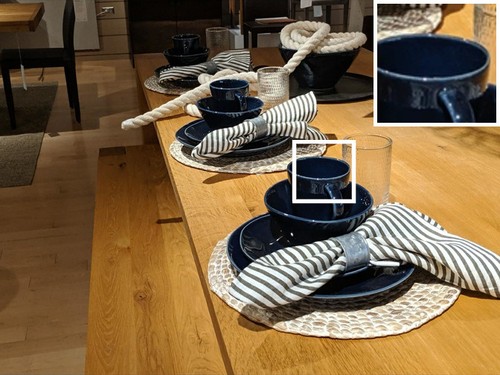} &
    \includegraphics[width=\resultswidth]{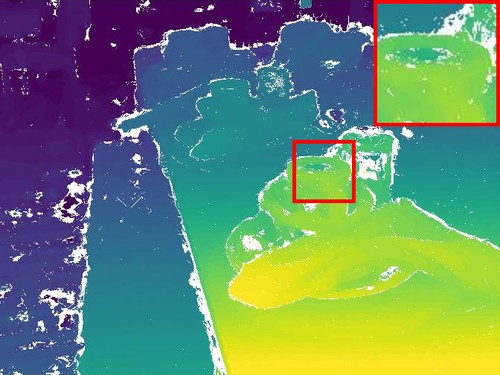} & 
    \includegraphics[width=\resultswidth]{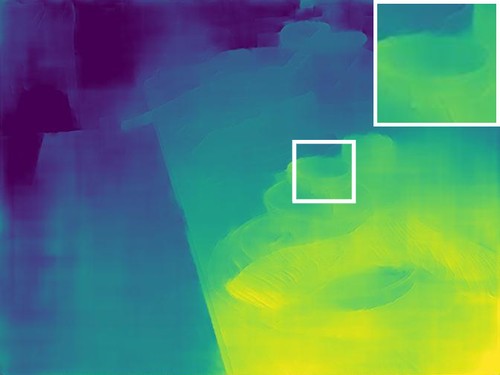} &
    \includegraphics[width=\resultswidth]{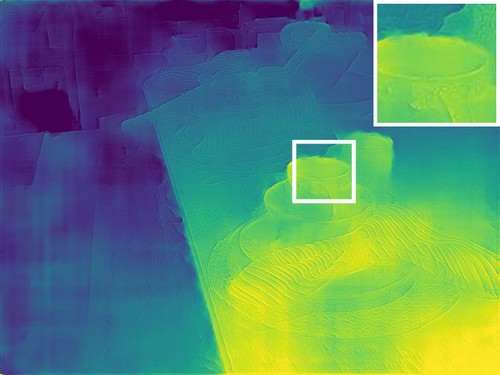} &
    \includegraphics[width=\resultswidth]{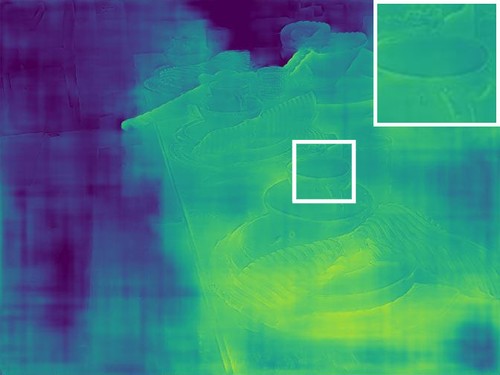} &
    \includegraphics[width=\resultswidth]{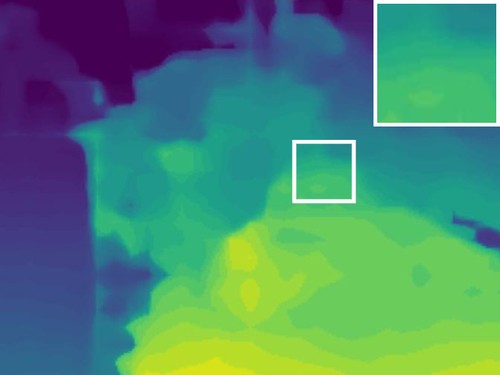} &
    \includegraphics[width=\resultswidth]{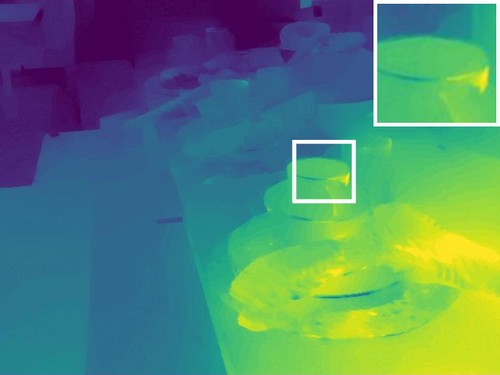} \\
    \fakecaption{(a) \,Input Image} & \fakecaption{(b)\,GT Depth} & \fakecaption{(c)\,DPNet (Affine)} & \fakecaption{(d)\, DPNet (Scale)} & \fakecaption{(e)\, DPNet (Affine)} & \fakecaption{(f)\,DORN \cite{DORN2018}} & \fakecaption{(g)\,Wadhwa \etal} \\
    \addlinespace[-1.2ex]
    \fakecaption{} & \fakecaption{} & \fakecaption{RGB + DP} & \fakecaption{RGB + DP} & \fakecaption{RGB} & \fakecaption{RGB} & \fakecaption{RGB + DP \cite{wadhwa2018}} \\
    \end{tabular}
    \caption{
    Input images (a) from the test set of our dataset, their ground-truth depth (b), the output of DPNet with RGB + DP input trained with affine invariance (c) and scale invariance (d), output of DPNet with RGB input trained with affine invariance (e), and as baselines, the output of \cite{DORN2018} trained on the NYUDv2 dataset ~\cite{SilbermanECCV12NYUdv2} (f) and the output of \cite{wadhwa2018} (g). An affine transform has been applied to all visualizations to best fit the ground truth.
    Results from \cite{wadhwa2018} exhibit fine details due to the use of bilateral smoothing \cite{barron2016fast} as a post process but otherwise show many depth errors, e.g., the shadow of the dog on the ground. DORN \cite{DORN2018} lacks fine details and fails to generalize to the variety of scenes in our dataset. For DPNet, results with RGB + DP input are better than results with RGB input. RGB + DP input with affine invariance yields better results than scale invariance, e.g., the space between the pencils in the second image. Best seen zoomed-in in an electronic version.}
    \label{fig:depth_showcase}
\end{figure*}

We show that our models trained with affine invariant loss have higher accuracy than those trained with conventional losses.
Our loss also improves the accuracy of existing view-supervision based monocular depth estimation methods when applied to dual-pixel data.
As benchmarks, we compare against Fu \etal \cite{DORN2018}, the current top performing monocular depth estimation algorithm on the KITTI \cite{Menze2015KITTI} and ScanNet \cite{Dai2017scannet} benchmarks, which has been trained on large pre-existing external RGBD datasets, and Wadhwa \etal \cite{wadhwa2018}, that applies classical stereo methods to recover depth from dual-pixels.

We evaluate our affine invariant loss against two baseline strategies: a scale invariant loss, and no invariance.
Scale-invariance is motivated by the well-understood inherent scale ambiguity in monocular depth estimation, as used by Eigen \etal~\cite{eigen2014}.
No-invariance is motivated by view-supervised monocular depth estimation techniques that directly predict disparity \cite{garg2016unsupervised, godard2017}.
We implement scale invariance by fixing $b=0$ in Eqn.~\ref{eqn:lst_sqr}.

Our affine-invariant loss can also be used to enable view-supervised monocular depth estimation techniques \cite{garg2016unsupervised, godard2017} to use dual-pixel data.
Since they require stereo data for training, we used images from the center and top cameras of our rig as the left and right images in a stereo pair.
The technique of Godard \etal \cite{godard2017} expects  rectified stereo images as input, which is problematic because our images are not rectified, and rectifying them would require a resampling operation that would act as a lowpass filter and thereby remove much of the depth signal in our dual-pixel data.
We circumvent this issue by replacing the one dimensional bilinear warp used by \cite{godard2017} during view supervision with a two dimensional warp based on each camera's intrinsics and extrinsics.
We also remove the scaled sigmoid activation function used when computing  disparities, which improved performance due to our disparities being significantly larger than those of the datasets used in \cite{godard2017}.
We also decreased the weight of the left-right consistency loss by $0.15$ times to compensate for our larger disparities.
We use the ResNet50 \cite{resent2016} version of \cite{godard2017}'s model as they report it provides the highest quality depth prediction.
We also used the codebase of \cite{godard2017} to implement a version of \cite{garg2016unsupervised} by removing the left-right consistency loss and the requirement that the right disparity map be predicted from the left image.

We show quantitative results in Table~\ref{table:losses_comparison}, and visualizations of depth maps in Fig.~\ref{fig:depth_showcase} (see the supplement for additional images).
Using our affine-invariant loss instead of scale-invariant or not-invariant loss improves performance for all models where different degrees of invariance are investigated.
In particular, while VGG is more accurate than DPNet when using no invariance, affine invariance allows the small DPNet model to achieve the best results.
In comparing the performance of DPNet with and without DP input, we see that taking advantage of dual-pixel data produces significantly improved performance over using just RGB input.
While \cite{DORN2018} trained on external RGBD datasets performs well on this task when compared to our model trained on just RGB data, its accuracy is significantly lower than our model and many baseline models trained on dual-pixel data with our affine-invariant loss, thereby demonstrating the value of DP imagery.

\section{Conclusion}

In summary, we have presented the first learning based approach for estimating depth from dual-pixel cues.
We have identified a fundamental affine ambiguity regarding depth as it relates to dual-pixel cues, and with this observation we have developed a technique that allows neural networks to estimate depth from dual-pixel imagery despite this ambiguity.
To enable learning and experimentation for this dual-pixel depth estimation task, we have constructed large dataset of 5-view in-the-wild RGB images paired with dual-pixel data.
We have demonstrated that our learning technique enables our model (and previously published view-supervision-based depth estimation models) to produce accurate, high-quality depth maps from dual-pixel imagery.

\section*{Acknowledgements}

We thank photographers Michael Milne and Andrew Radin for collecting data. We also thank Yael Pritch and Marc Levoy for technical advice and helpful feedback. 

{\small
\bibliographystyle{ieee_fullname}
\bibliography{references}
}

\clearpage
\appendix
\section*{Supplement}

\section{Derivation of Equation 2} 

Suppose a point light source is located at depth $Z(x, y)$ where the center of the camera lens is at position $(0, 0, 0)$. Light from this point light source is focused by the lens to another point on the opposite side of the lens. Let $Z_i$ be the distance from the lens to this other point. Also, let $g$ be the focus distance and $g_i$ be the position of the sensor. By the paraxial and thin-lens approximations,
\begin{equation}
    \frac{1}{g_i} = \frac{1}{f} - \frac{1}{g} \quad \text{ and } \quad  \frac{1}{Z_i} = \frac{1}{f} - \frac{1}{Z}.
    \label{eq:blur1}
\end{equation}
By similar triangles, the blur size $b$ is 
\begin{equation}
    b = \frac{L(g_i - Z_i)}{Z_i}
    \label{eq:blur2}
\end{equation}

Substituting Eqn.~\ref{eq:blur1} into Eqn.~\ref{eq:blur2}, we get Eqn.~2 in the main paper
\begin{equation}
    b = \frac{Lg}{1-f/g}\left(\frac{1}{g}-\frac{1}{Z}\right)
    \label{eq:blur3}
\end{equation}

\section{Data Processing}

\subsection{Depth from Multi View Stereo}

We use two different stereo algorithms for computing the ``ground truth'' depth maps we use for training and evaluation.
These depth maps are computed at a resolution of $756 \times 1008$, i.e., one quarter the resolution of RGB images.

We use the COLMAP multi-view stereo algorithm \cite{COLMAP, SchonbergerMvs16} that computes per pixel depth and filters based on geometric consistency. We sample depth using inverse perspective sampling in the range [0.2m, 100m] to yield $\depth$. Confidence $\depthconf$ is set to zero wherever COLMAP does not provide depth due to geometric inconsistency, and is set to 1 elsewhere. 

Because the depth maps from COLMAP tend to have edge fattening artifacts on our data, we implemented our own plane-sweep multi-view stereo algorithm. We plane sweep along $256$ planes sampled using inverse perspective sampling in the range [0.2m, 100m] and take the minimum of a filtered cost volume as each pixel's depth. To compute the cost volume, for each pixel, we compute the sum of absolute differences for each of the warped neighbors and then bilaterally filter the cost volume using the grayscale reference image as the guide image. This ensures that we are aggregating costs over similar pixels in a local window thus avoiding edge fattening artifacts \cite{Richardt2010}. We use a spatial sigma of $3$ pixels and a range sigma of $12.5$ for the bilateral filter. Finally, we normalize the plane indices to [$0$, $1$] range so that they are in the same domain as COLMAP depth. To compute confidence, we check for depth coherence across views by checking for left / right consistency \cite{Bleyer2011}. We first compute consistency with each of the $4$ neighboring images:
\begin{equation}
C_j(x,y) = \exp{\left(-\frac{\norm{\depth_0(x,y) -\depth_j\left(M(x,y; \depth_0)\right) }^2}{2\sigma^2}\right)}
\end{equation}
where $\sigma=\sfrac{1}{256}$, and $j$ is the index of the neighboring image. Then, under the assumption that a pixel must be visible in at least two other cameras for its depth to be reliable, we take the product of the largest two $C_j(x,y)$ values for each pixel to compute our final confidence $\depthconf(x,y)$.

A sample of images from our test set with corresponding depth from our method and COLMAP is shown in Fig.~\ref{fig:ours_vs_colmap}.

\newcommand{\resultswidthgt}{0.153\textwidth}{
\begin{figure}
    \centering
    \begin{tabular}{@{}c@{\,\,}c@{\,\,}c@{\,\,}}
    \includegraphics[width=\resultswidthgt]{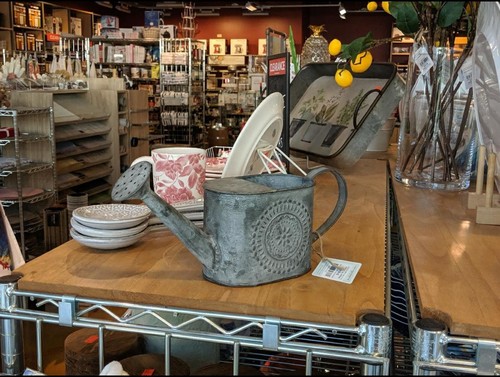} &
    \includegraphics[width=\resultswidthgt]{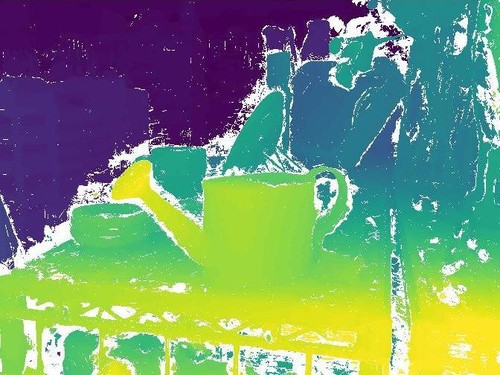} & 
    \includegraphics[width=\resultswidthgt]{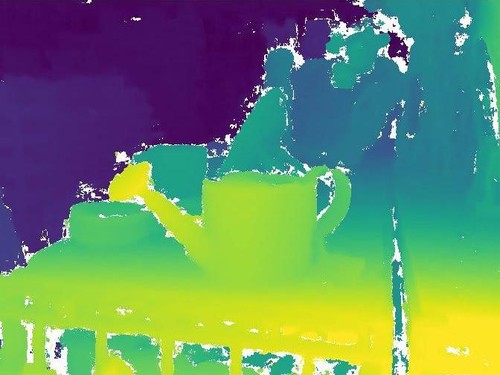} \\
    \includegraphics[width=\resultswidthgt]{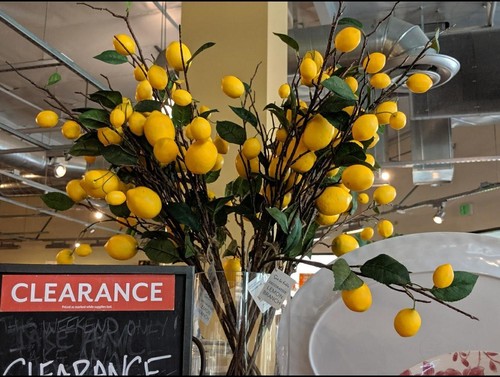} &
    \includegraphics[width=\resultswidthgt]{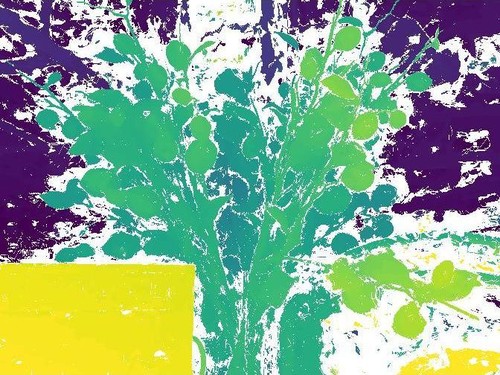} &
    \includegraphics[width=\resultswidthgt]{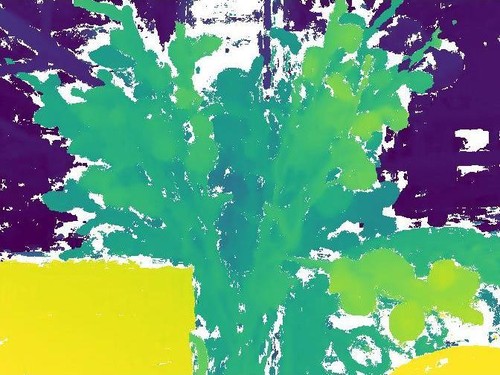} \\
    \includegraphics[width=\resultswidthgt]{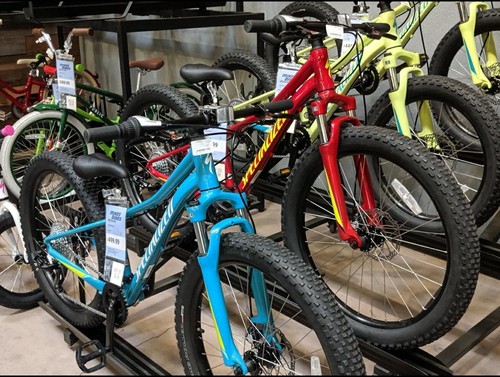} &
    \includegraphics[width=\resultswidthgt]{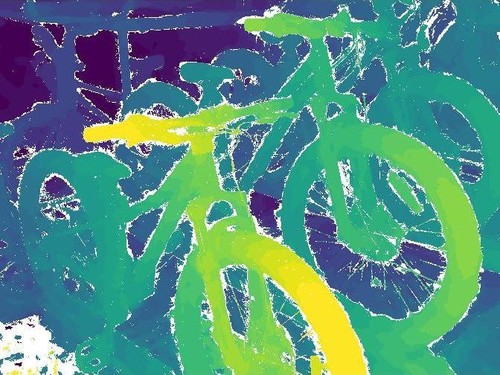} & 
    \includegraphics[width=\resultswidthgt]{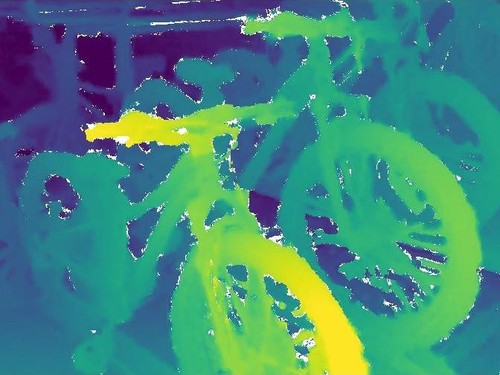} \\
    \fakecaption{(a) \,RGB Image} & \fakecaption{(b)\,Our True Depth} & \fakecaption{(c)\,COLMAP's Depth}\\
    \end{tabular}
    \caption{Images (a) from our test set, ground truth depth (b) computed using our multi-view stereo pipeline, and ground truth depth (c) computed using COLMAP \cite{COLMAP}. Low confidence regions are shown in white. Our depth tends to be conservative in labelling a depth sample confident and avoids edge fattening artifacts.}
    \label{fig:ours_vs_colmap}
\end{figure}

\subsection{Preprocessing of RGB and Dual-Pixel Data}
RGB images are $3024 \times 4032$, but we always downsample them to $1512 \times 2016$. 
The green pixels in each Bayer quad on the camera sensor are split in half (Fig. 2 in main paper).
In the RGB image, the sensor sums adjacent green half-pixels to form the green channel. In the DP data, the sensor bins four green half-pixels in a $4 \times 2$ pattern to yield DP data of size $756 \times 2016 \times 2$. This data is 10-bit raw data and we apply a square root to the raw data and quantize the result to 8-bits. We also upsample the first dimension to $1512$, so that its dimensions match those of the downsampled RGB image. When feeding RGB and DP images as input to the model, they are simply concatenated along the channel dimension to form a 5 channel input.

\section{Evaluation Metrics}

\subsection{Affine Invariant Weighted Error}

One metric we use to evaluate models against our ground-truth depths is the minimum of a $L^p$-norm between an estimated inverse depth $\preddepth$ and the true inverse depth $\depth$ (weighted by the true depth's confidence $\depthconf$ and scaled by the number of pixels) under all possible affine transformations. Let us define $\AIWE(p)$ as:
\begin{equation}
\resizebox{\linewidth}{!}{
$\displaystyle
\min_{a, b} \left( { \sum_{(x,y)} \depthconf(x,y) \abs{\depth(x,y) - \left( a \preddepth(x,y) + b \right)}^{p} \over  \sum_{(x,y)} \depthconf(x,y) }\right)^{1 / p}
$}
\end{equation}
Because root mean squared error and mean absolute error are standard choices when comparing depth maps, in the paper we present $\AIWE(2)$ (affine-invariant weighted RMSE) and $\AIWE(1)$ (affine-invariant weighted MAE).
$\AIWE(2)$ can be evaluated straightforwardly by solving a least-squares problem, and $\AIWE(1)$ can be computed using iteratively reweighted least squares (in our experiments, we use $5$ iterations).

\subsection{Spearman's Rank Correlation Coefficient}

We also use Spearman's rank correlation coefficient $\rho_s$ for evaluation, which evaluates the ordinal correctness of the estimated depth. Because $\rho_s$ is a function of the rank of each pixel's depth, it is invariant to any monotonic transformation of the depth which, naturally, includes affine transformations (with positive scales). Because our ground-truth depths $\depth$ may contain repeated elements, $\rho_s$ is computed by first computing the ranks of all elements in $\depth$ and $\preddepth$ and then computing the Pearson correlation of those ranks. We use the ground-truth depth confidences $\depthconf$ when computing Pearson correlation (using it to weight the expectations used to compute the variances and covariance of the ranks) thereby resulting in a weighted variant of Spearman's $\rho$. To handle cases when the affine scaling is negative, we take the absolute value of $\rho_s$, and we report $1-\abs{\rho_s}$ to maintain consistency with $\AIWE(\cdot)$, in terms of lower values being better.

\section{Model Architecture}

Our DPNet architecture is composed of two key building blocks, an encoder block $E(i, o, s)$ and a decoder block $D(i, o)$ where $i$ denotes number of intermediate features, $o$ denotes number of output features, and $s$ denotes the stride which controls the downsampling done by the encoder block. Each encoder block takes as input the output from the previous encoder block. Each decoder block takes as input the output of the previous decoder block and the output of an encoder block as a skip connection. Unless otherwise mentioned, we use Batch Normalization \cite{Ioffe2015BatchNorm} before each convolution layer and PReLu \cite{He2015PRelu} as an activation function for each output with initial leakiness $a_i$ set to be $0.05$.
\newcommand{\encA}{E_a}
\newcommand{\encB}{E_b}

Each encoder block $\encA(i, o, s)$ consists of a series of 3 convolutional layers, the first of which has $i$ filters with size $3\times3$ and stride $s$, the second of which is a depthwise separable $3\times3$ convolutional layer with $i$ filters, and the third of which is a $1\times1$ convolutional layer with $o$ filters whose output is added to the max-pooled input (with pool size and stride both $s$) before applying a PReLu activation.

We also use a different encoder block $\encB(o, s)$ that is directly applied to the input images, which is a convolutional layer with $o$ filters of size $7\times7$ and stride $s$, whose output is concatenated with max-pooled input images with pool size and stride $s$.

For each decoder block $D(i, o)$, we first apply a $4\times4$ transposed convolutional layer with stride $2$ and $i$ filters to the output of the previous decoder layer, followed by a $3\times3$ depth separable convolutional layer and a $1\times1$ convolutional layer, each with $i$ filters, followed by a $3\times3$ depth separable convolutional layer with $i$ filters whose output is added to the filtered skip connections before which itself has been filtered via a $3\times3$ depth separable convolutional layer. PReLu activation is applied after summing the two. Finally, a $1\times1$ convolutional layer with $o$ filters generates the output for the next decoder block. 

The overall model consists of a series of encoders followed by a series of decoders:

\begin{tabular}{lll}
$\encB(8, 2)$ & $\encA^1(11, 11, 1)$ & \\
$\encA(16, 32, 2)$ & $\encA(16, 32, 1)$ & $\encA^2(16, 32, 1)$ \\
$\encA(16, 64, 2)$ & $\encA(16, 64, 1)$ & $\encA^3(16, 64, 1)$ \\
$\encA(32, 128, 2)$ & $\encA(32, 128, 1)$ & $\encA^4(32, 128, 1)$ \\
$\encA(32, 128, 2)$ & $\encA(32, 128, 1)$ & $\encA(32, 128, 1)$ \\
$D^4(32, 128)$ \\
$D^3(16, 64)$ \\
$D^2(16, 32)$\\
$D^1(8, 8)$
\end{tabular}
where the outputs of each encoder marked with a superscript are connected by skip connections to a corresponding decoder with that superscript. The predictions at 5 different resolutions are obtained by applying a $3\times3$ convolution with a single filter and no activation and no Batch Normalization to the outputs of the decoders and the last encoder.

Our VGG model is same as that of \cite{godard2017} but we remove the last decoder block since our depth maps are at half the input resolution.

\section{Supplementary Results}

\subsection{VGG vs DPNet}

\newcommand{\vggresultswidth}{0.154\textwidth}{
\begin{figure}[b]
    \centering
    \begin{tabular}{@{}c@{\,\,}c@{\,\,}c@{}}
    \includegraphics[width=\vggresultswidth]{images/results/rgb_1.jpg} &
    \includegraphics[width=\vggresultswidth]{images/results/rgb_pd_clean_view_sup_k_1_cropped_1.jpg} &
    \includegraphics[width=\vggresultswidth]{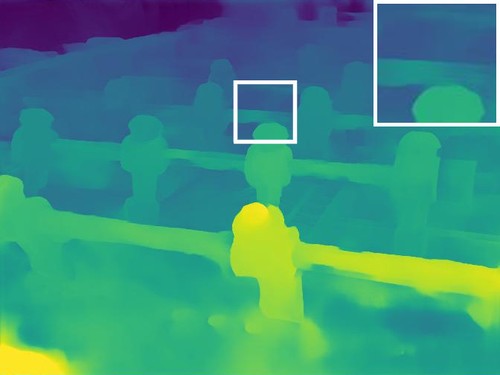}
    \\
    \includegraphics[width=\vggresultswidth]{images/results/rgb_2.jpg} &
    \includegraphics[width=\vggresultswidth]{images/results/rgb_pd_clean_view_sup_k_1_cropped_2.jpg} &
    \includegraphics[width=\vggresultswidth]{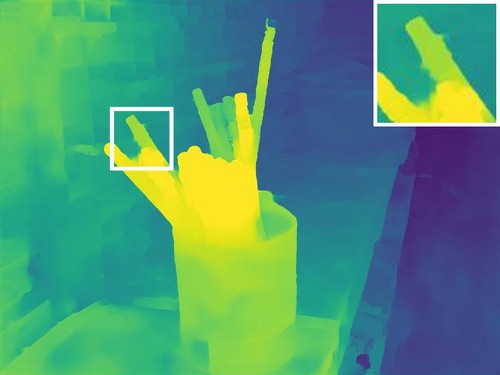}
    \\
    \includegraphics[width=\vggresultswidth]{images/results/rgb_3.jpg} &
    \includegraphics[width=\vggresultswidth]{images/results/rgb_pd_clean_view_sup_k_1_cropped_3.jpg} &
    \includegraphics[width=\vggresultswidth]{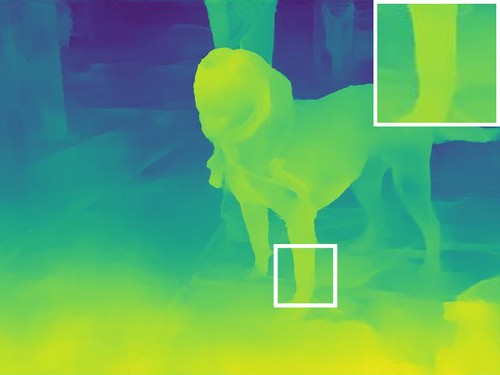}
    \\
    \includegraphics[width=\vggresultswidth]{images/results/rgb_4.jpg} &
    \includegraphics[width=\vggresultswidth]{images/results/rgb_pd_clean_view_sup_k_1_cropped_4.jpg} &
    \includegraphics[width=\vggresultswidth]{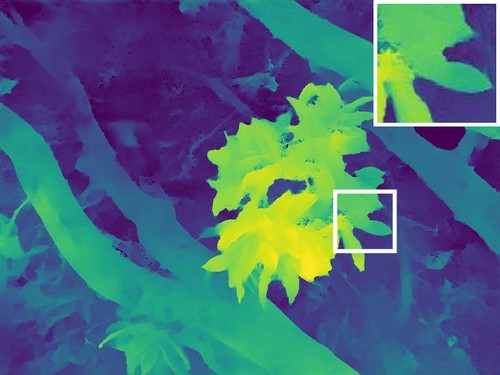}
    \\
    \includegraphics[width=\vggresultswidth]{images/results/rgb_5.jpg} &
    \includegraphics[width=\vggresultswidth]{images/results/rgb_pd_clean_view_sup_k_1_cropped_5.jpg} &
    \includegraphics[width=\vggresultswidth]{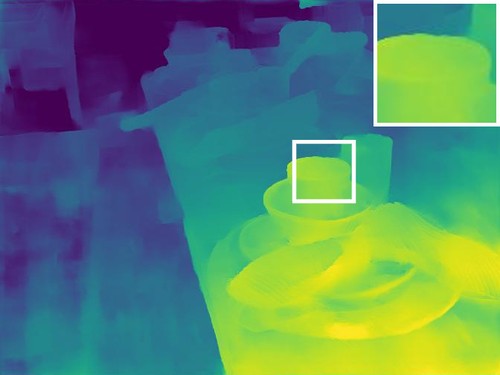}
    \\
    \fakecaption{(a) \,Input Image} & \fakecaption{(c)\,DPNet (Affine)} & \fakecaption{(c)\,VGG (Affine)} \\
    \addlinespace[-1.2ex]
    \fakecaption{} & \fakecaption{RGB + DP} & \fakecaption{RGB + DP}\\
    
    \end{tabular}
    \vspace{-0.05in}
    \caption{Results of DPNet and VGG, both with RGB + DP input and trained with affine invariance.}
    \label{fig:vgg_vs_dpnet}
\end{figure}

As reported in Table 2 in the main paper, the best results with VGG model are slightly inferior than DPNet in spite of having a larger capacity. This is because VGG has a tendency to overfit due to the larger capacity. Qualitatively, we find the results with VGG to be very similar to results with DPNet (Fig~\ref{fig:vgg_vs_dpnet}). VGG also overfits for RGB input, hence those results are omitted from Table 2 in the main paper.

\subsection{Multi-View vs Stereo}

In Table~\ref{table:viewpoint_experiments}, we present additional results in which we demonstrate that simultaneously training our model on all $4$ alternative views provided by our capture setup outperforms an ablation of our technique that has been trained on only stereo views.

\begin{table}
    \centering
    \resizebox{\linewidth}{!}{
    \begin{tabular}{l|ccc}
    &  $\AIWE(1)$ & $\AIWE(2)$ & $1 - \abs{\rho_s}$ \\
    \hline
    DPNet trained on Stereo & .0218 & .0319 & .180 \\
    DPNet trained on Multi-view & \textbf{.0175} & \textbf{.0264} & \textbf{.139}\\
    \end{tabular}
    }
    \caption{Comparison of stereo training data with multi-view training data. Accuracy is higher when using all the views for training vs just using the center-bottom camera pairs from the capture rig.
    }
    \label{table:viewpoint_experiments}
\end{table}

\subsection{Generalization Across Devices}

\begin{table}
    \centering
    \resizebox{\linewidth}{!}{
    \begin{tabular}{l|ccc}
    &  $\AIWE(1)$ & $\AIWE(2)$ & $1 - \abs{\rho_s}$ \\
    \hline
    Extended test set & .0188 & .0276 & .153 \\
    Standard test set & \textbf{.0175} & \textbf{.0264} & \textbf{.139}\\
    \end{tabular}
    }
    \caption{DPNet's accuracy on our standard test set (which only contains devices that are in the training set) vs our extended test set, which contains 7 other devices that were not used to generate the training set. The accuracy is only slightly worse, suggesting that our model has learned to circumvent the need for calibration.}
    \label{table:device_generalization}
    
\end{table}

Our training and test sets consist of images from only three different phones.
As noted by Wadhwa \etal \cite{wadhwa2018}, the relationship between depth and disparity from dual-pixels can vary from device to device because of variations during the manufacturing process, which they compensate for with a calibration procedure.
Though we do not apply any per-device calibration, our performance degrades only slightly for phones that were not used to capture the images in the training set (Table~\ref{table:device_generalization}).
To demonstrate this, we evaluate our model on an extended test set containing devices that were not used to acquire any images in the training set. To construct this set, we use data from all 5 phones on the capture rig, i.e., the center phone and the surrounding phones. This extended set contains data from all 10 devices used for capture while the training set contains only data from just 3 of those devices.

\subsection{Additional Metrics}

\definecolor{Yellow}{rgb}{1,1,0.6}

\begin{table}
    \centering
    \resizebox{\linewidth}{!}{
    \begin{tabular}{l|c|c|c}
    \multirow{2}{*}{Method} & \multirow{2}{*}{Invariance} & \multicolumn{2}{c}{Percentile Based WRMSE}\\
    & & Our Depth & COLMAP Depth\\
    \hline
    \multirow{2}{*}{DPNet (RGB)}
    & Scale  & .0890 & .0908\\
    & Affine  & .1502 & .1484\\
    \hline
    \multirow{2}{*}{DPNet (RGB+DP)}
    & Scale  & .0390 & .0417\\
    & Affine  & \cellcolor{Yellow} \bf .0328 & \cellcolor{Yellow} \bf .0368\\
    \end{tabular}
    }
    \caption{WRMSE metric for a subset of rows from Table 2 in the main paper where affine ambiguity is resolved by considering $\sfrac{1}{3}$ and $\sfrac{2}{3}$ percentile values. This also shows the importance of DP input and affine invariance being useful for training with DP input.}
    \label{table:median_metric}
\end{table}

For scale invariant prediction, \cite{zhou2017unsupervised} introduces a metric where the scale ambiguity is resolved by taking the ratio of the median of the prediction and of the ground truth. This is not directly applicable to our affine invariant prediction, as a single correspondence does not overconstrain an affine transformation. To adapt this technique to the affine invariance case, we compute the $\sfrac{1}{3}$ and $\sfrac{2}{3}$ weighted percentile values (where the median would be the $\sfrac{1}{2}$ percentile) of the prediction and of the ground truth (using the confidences of the ground truth as weights) and then use those two correspondences to recover an affine transformation to resolve the ambiguity. Table~\ref{table:median_metric} shows that DP input is critical and affine invariance helps learning with DP input when measured with this metric.

\subsection{Additional Comparisons}

\newcommand{\resultswidthlandscape}{0.135\textwidth}
\begin{figure*}
    \centering
    \begin{tabular}{@{}c@{\,\,}c@{\,\,}c@{\,\,}c@{\,\,}c@{\,\,}c@{\,\,}c@{}}
    \includegraphics[width=\resultswidthlandscape]{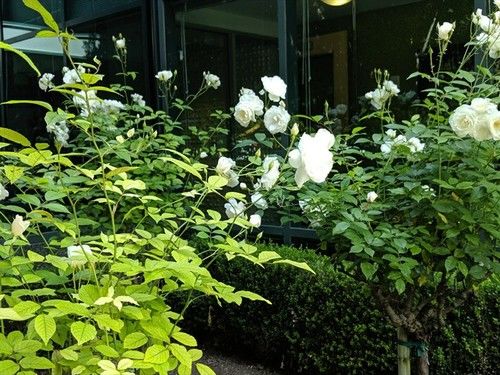} &
    \includegraphics[width=\resultswidthlandscape]{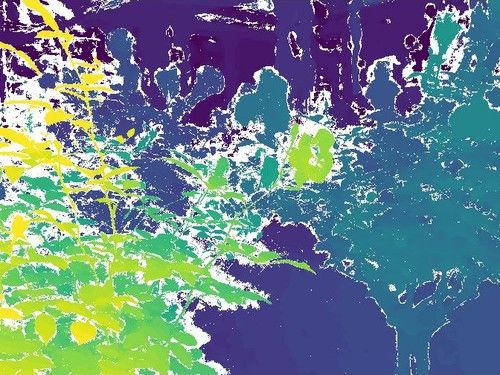} & 
    \includegraphics[width=\resultswidthlandscape]{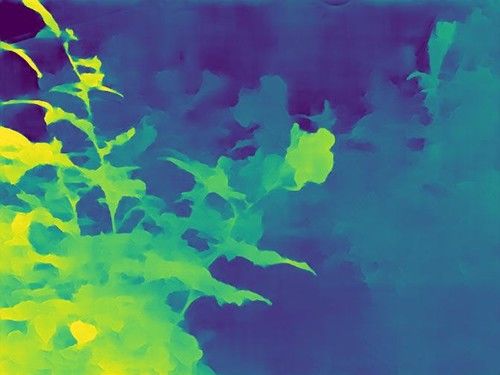} &
    \includegraphics[width=\resultswidthlandscape]{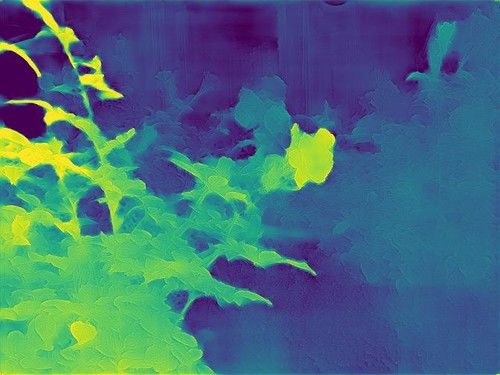} &
    \includegraphics[width=\resultswidthlandscape]{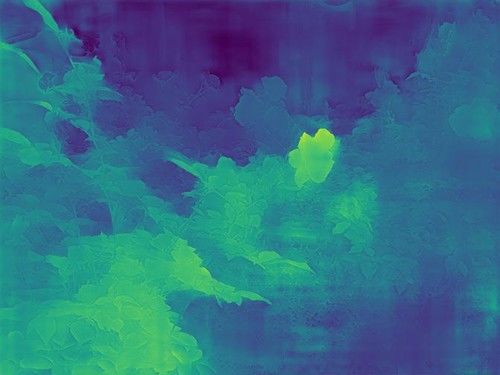} &
    \includegraphics[width=\resultswidthlandscape]{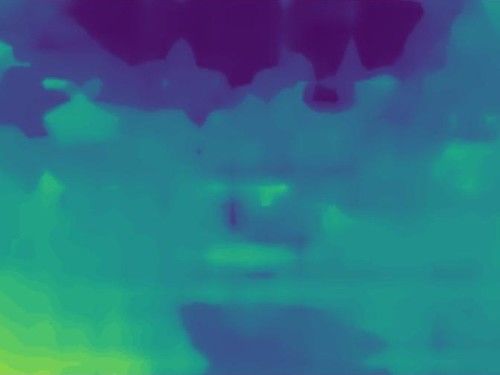} &
    \includegraphics[width=\resultswidthlandscape]{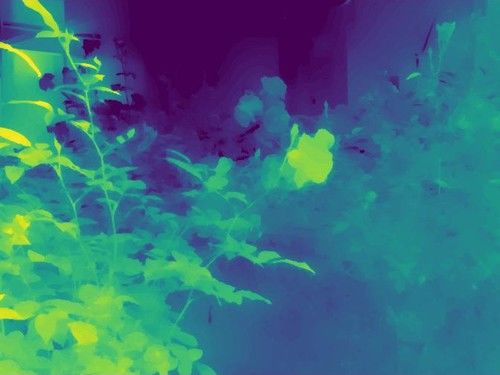} \\
    \includegraphics[width=\resultswidthlandscape]{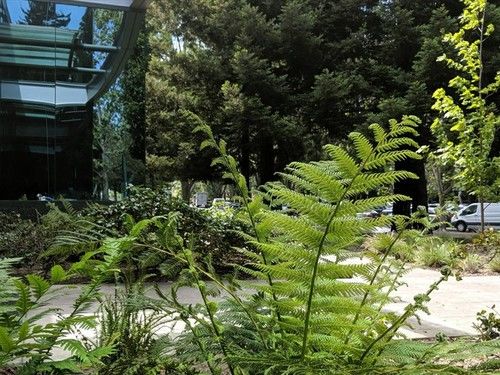} &
    \includegraphics[width=\resultswidthlandscape]{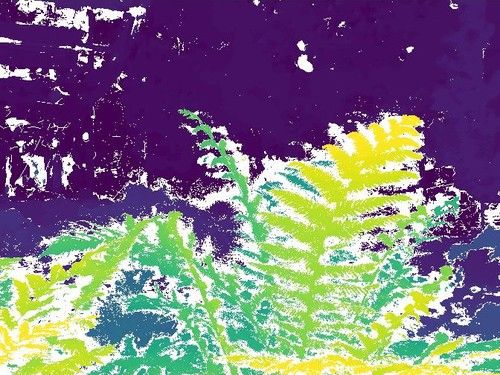} & 
    \includegraphics[width=\resultswidthlandscape]{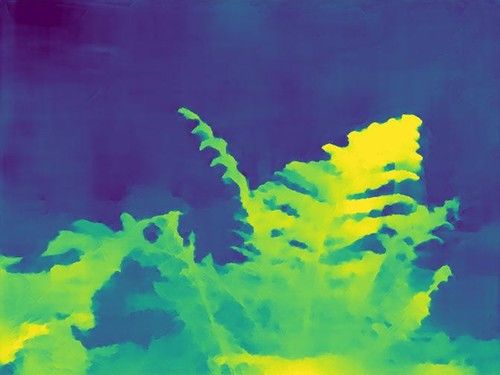} &
    \includegraphics[width=\resultswidthlandscape]{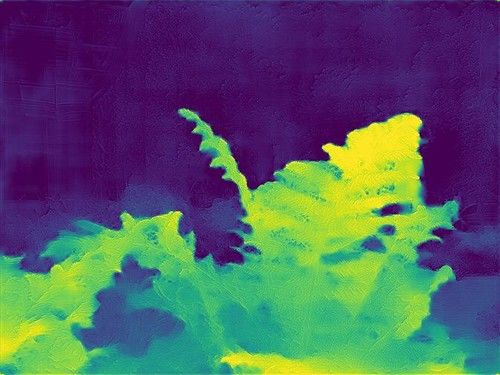} &
    \includegraphics[width=\resultswidthlandscape]{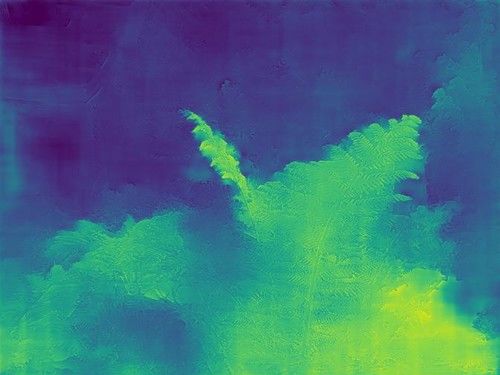} &
    \includegraphics[width=\resultswidthlandscape]{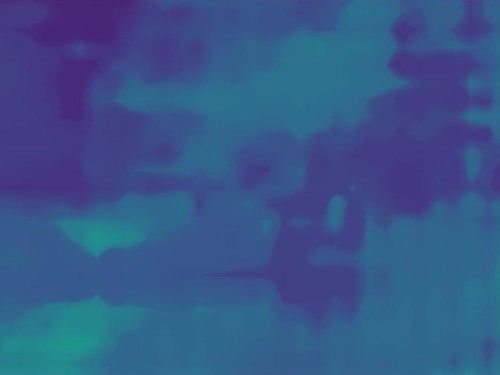} &
    \includegraphics[width=\resultswidthlandscape]{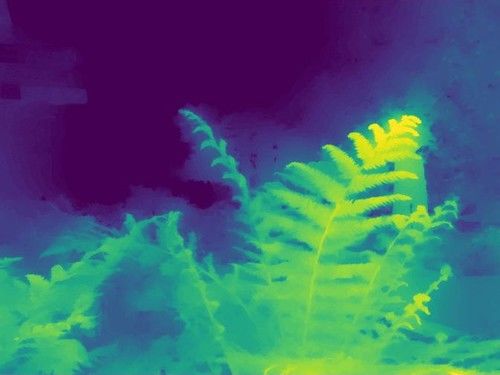} \\
    \includegraphics[width=\resultswidthlandscape]{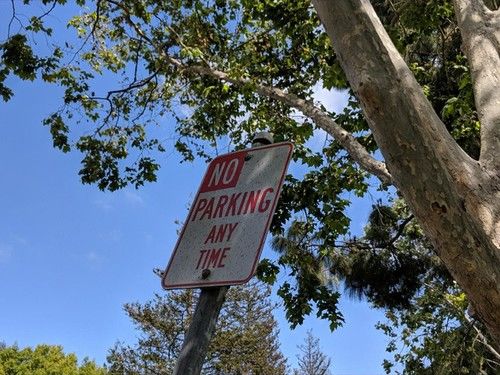} &
    \includegraphics[width=\resultswidthlandscape]{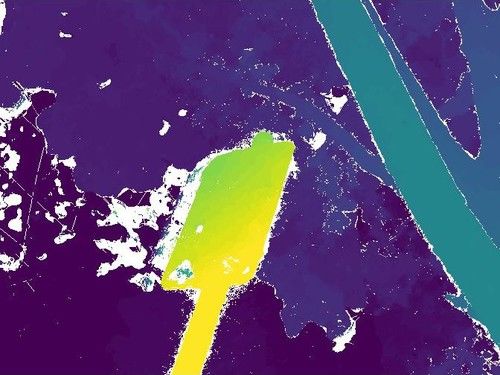} & 
    \includegraphics[width=\resultswidthlandscape]{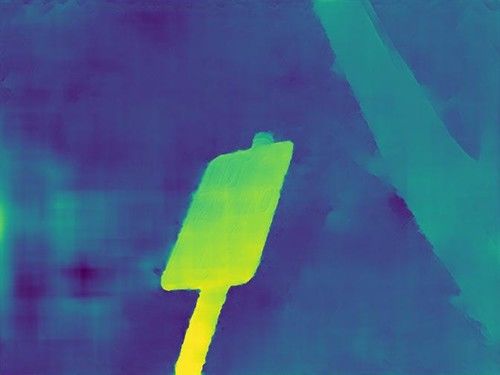} &
    \includegraphics[width=\resultswidthlandscape]{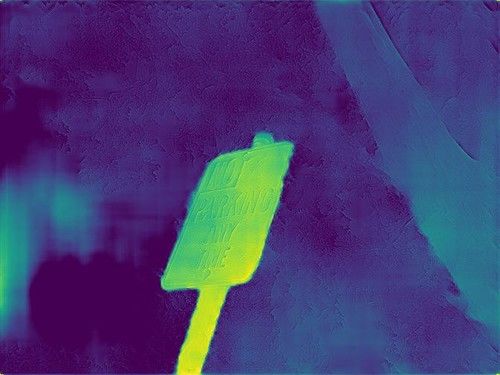} &
    \includegraphics[width=\resultswidthlandscape]{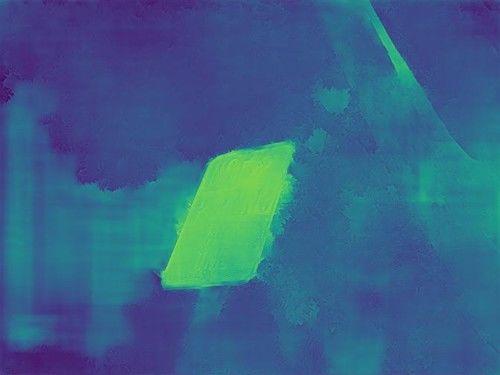} &
    \includegraphics[width=\resultswidthlandscape]{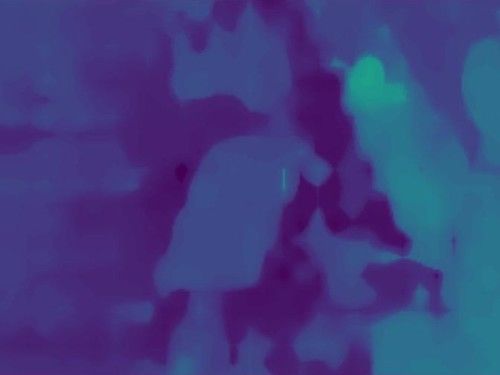} &
    \includegraphics[width=\resultswidthlandscape]{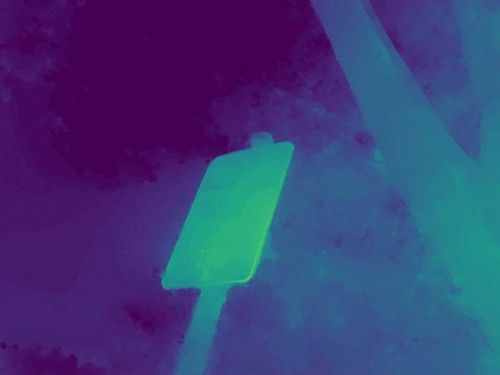} \\
    \includegraphics[width=\resultswidthlandscape]{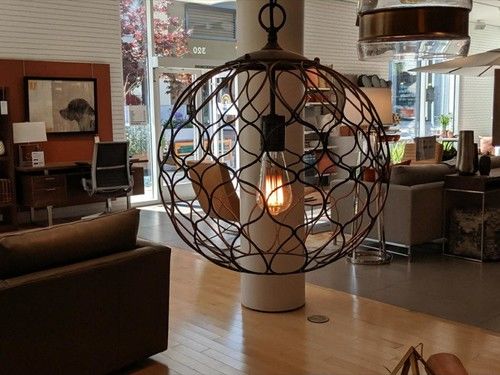} &
    \includegraphics[width=\resultswidthlandscape]{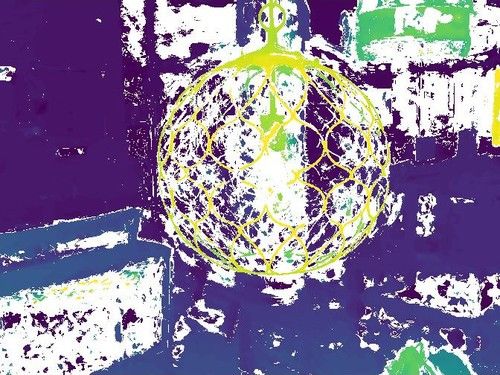} & 
    \includegraphics[width=\resultswidthlandscape]{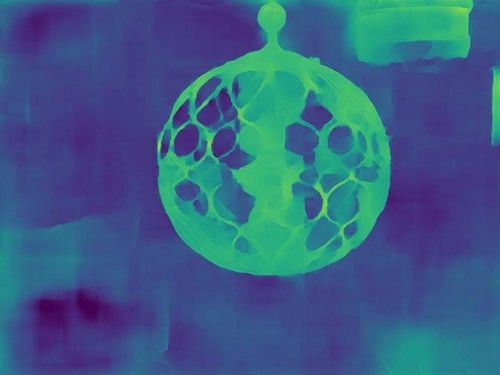} &
    \includegraphics[width=\resultswidthlandscape]{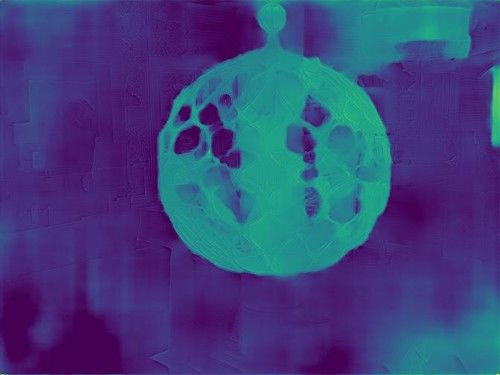} &
    \includegraphics[width=\resultswidthlandscape]{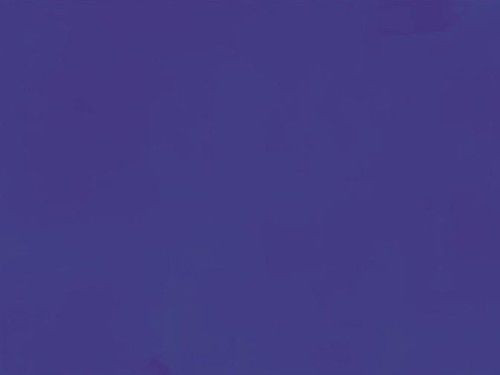} &
    \includegraphics[width=\resultswidthlandscape]{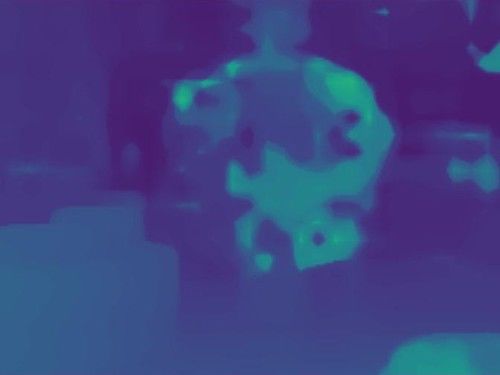} &
    \includegraphics[width=\resultswidthlandscape]{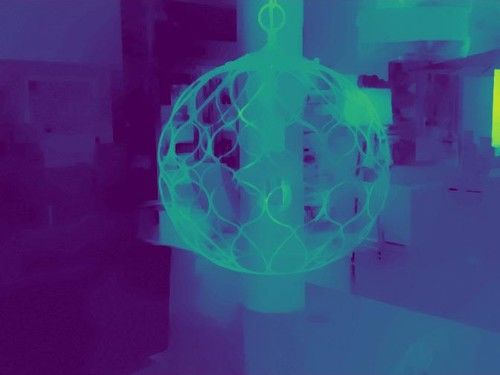} \\
    \includegraphics[width=\resultswidthlandscape]{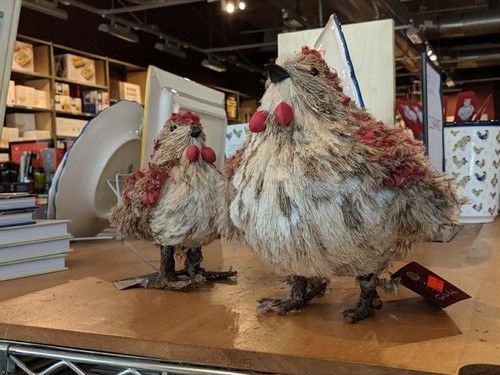} &
    \includegraphics[width=\resultswidthlandscape]{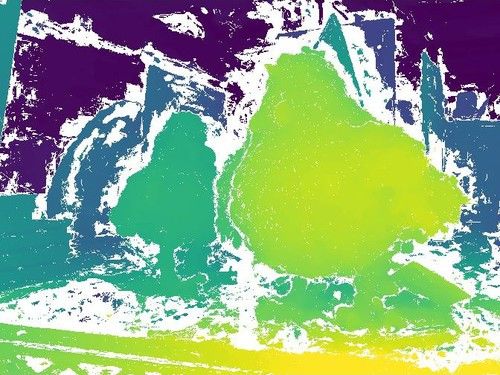} & 
    \includegraphics[width=\resultswidthlandscape]{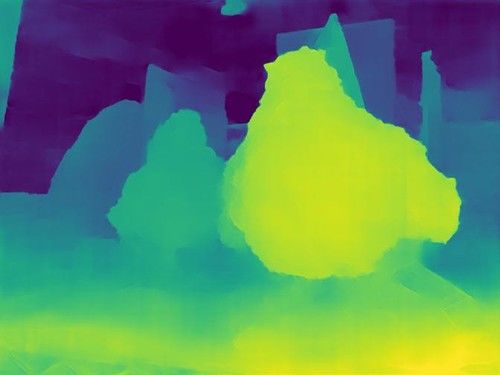} &
    \includegraphics[width=\resultswidthlandscape]{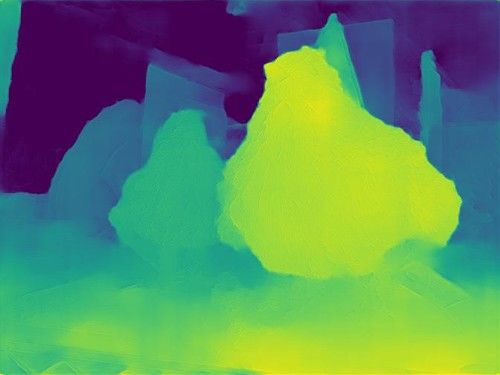} &
    \includegraphics[width=\resultswidthlandscape]{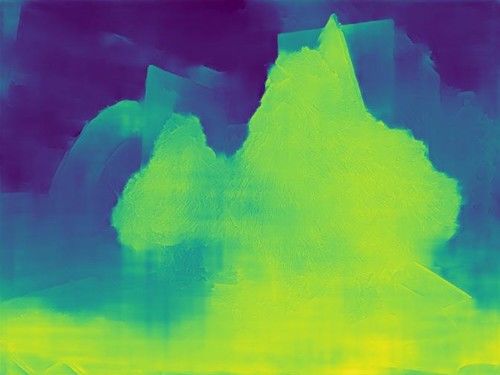} &
    \includegraphics[width=\resultswidthlandscape]{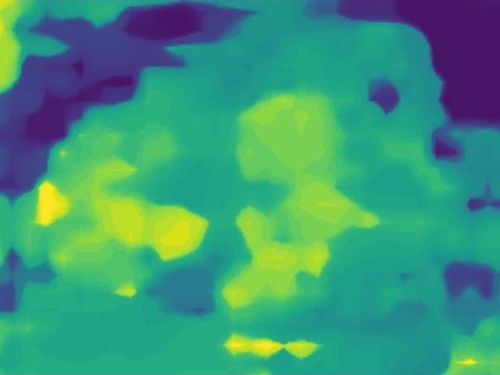} &
    \includegraphics[width=\resultswidthlandscape]{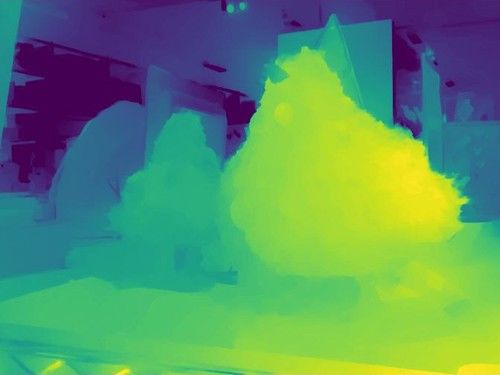} \\
    \includegraphics[width=\resultswidthlandscape]{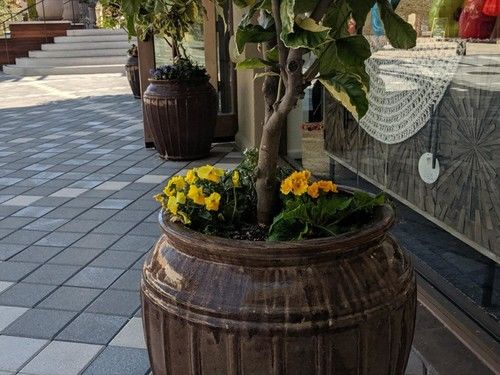} &
    \includegraphics[width=\resultswidthlandscape]{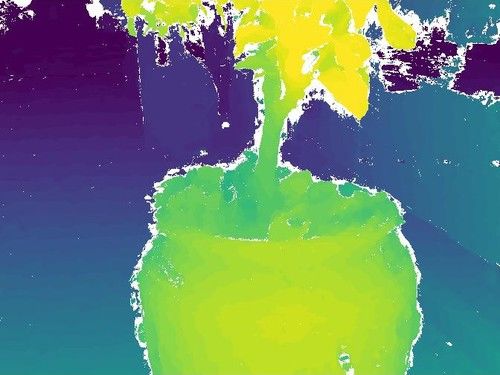} & 
    \includegraphics[width=\resultswidthlandscape]{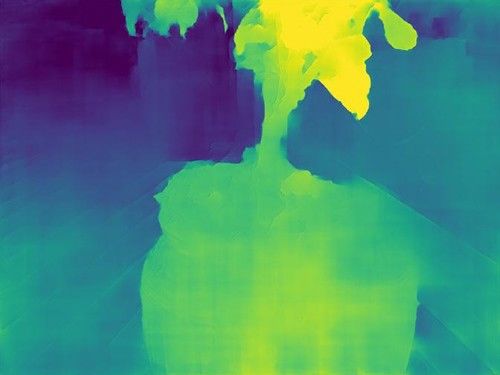} &
    \includegraphics[width=\resultswidthlandscape]{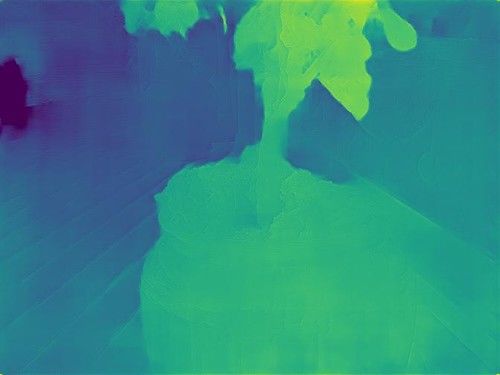} &
    \includegraphics[width=\resultswidthlandscape]{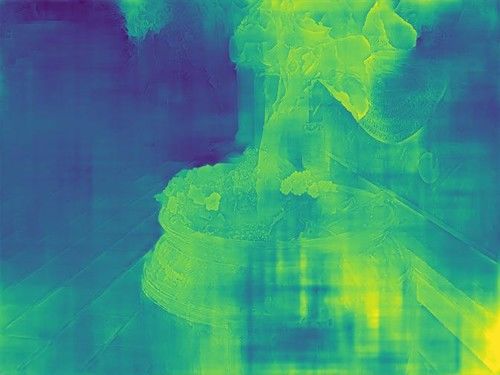} &
    \includegraphics[width=\resultswidthlandscape]{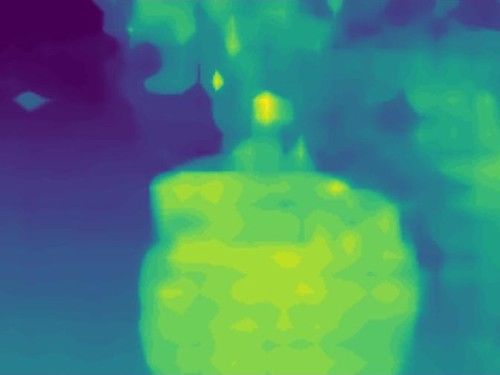} &
    \includegraphics[width=\resultswidthlandscape]{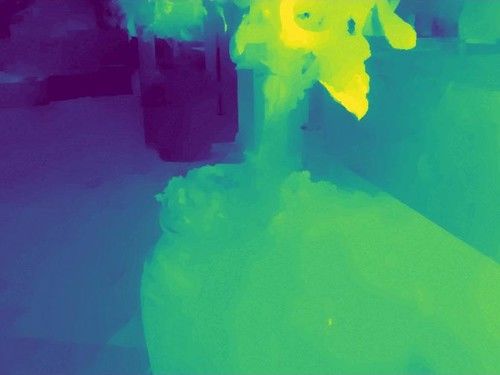} \\
    \includegraphics[width=\resultswidthlandscape]{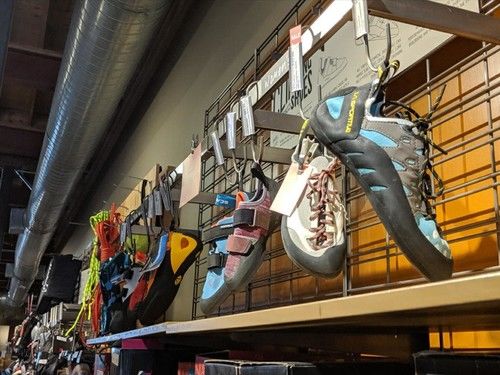} &
    \includegraphics[width=\resultswidthlandscape]{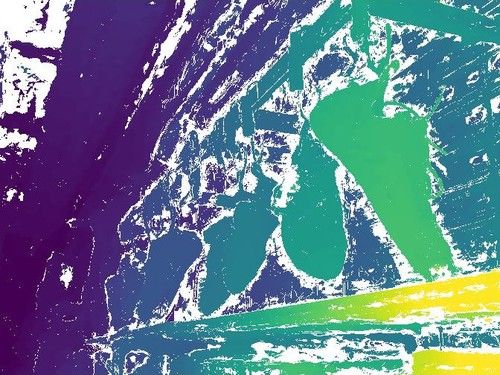} & 
    \includegraphics[width=\resultswidthlandscape]{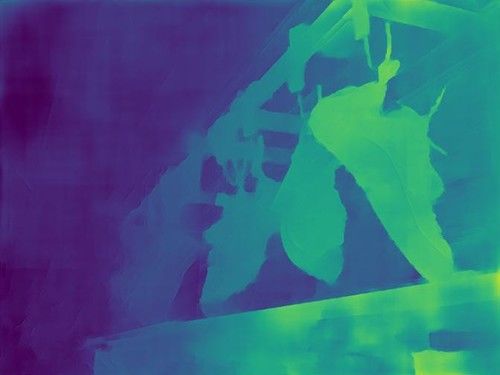} &
    \includegraphics[width=\resultswidthlandscape]{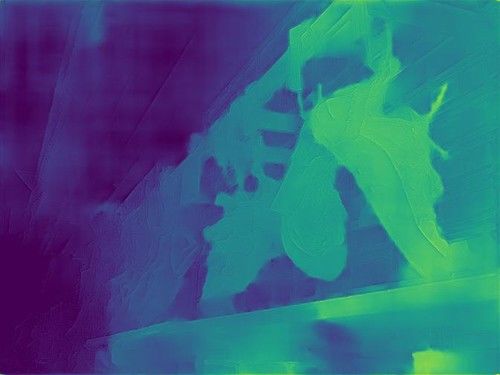} &
    \includegraphics[width=\resultswidthlandscape]{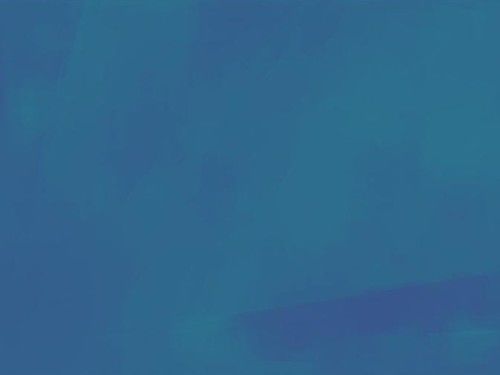} &
    \includegraphics[width=\resultswidthlandscape]{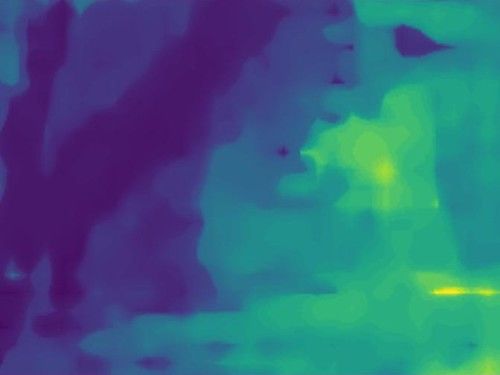} &
    \includegraphics[width=\resultswidthlandscape]{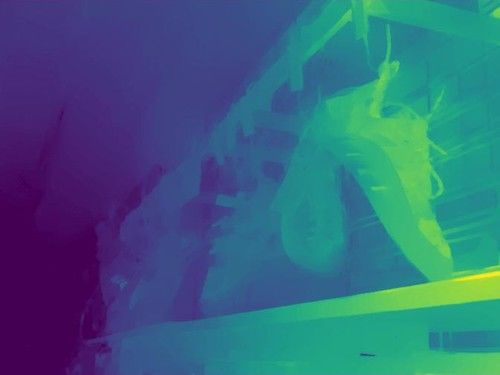} \\
    \includegraphics[width=\resultswidthlandscape]{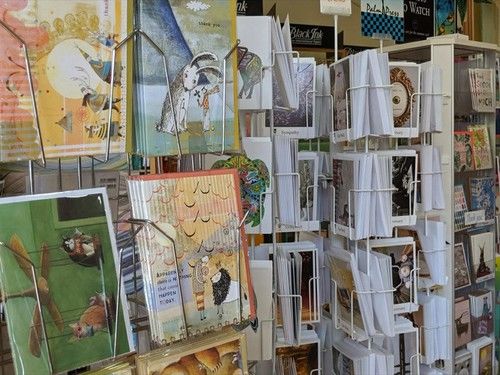} &
    \includegraphics[width=\resultswidthlandscape]{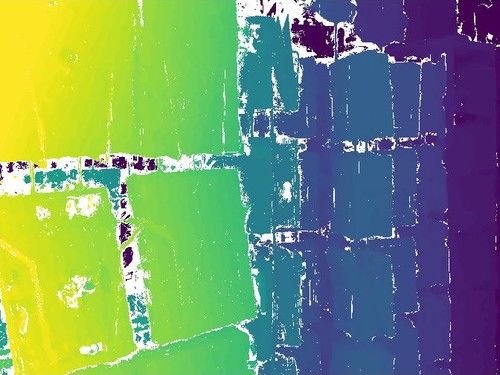} & 
    \includegraphics[width=\resultswidthlandscape]{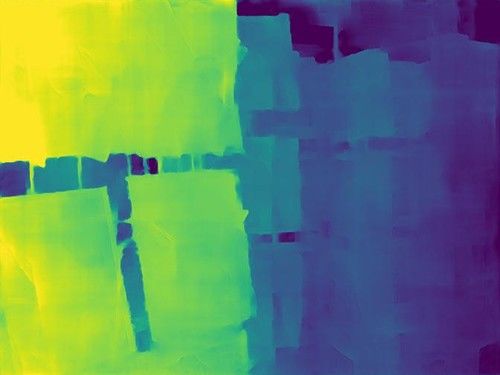} &
    \includegraphics[width=\resultswidthlandscape]{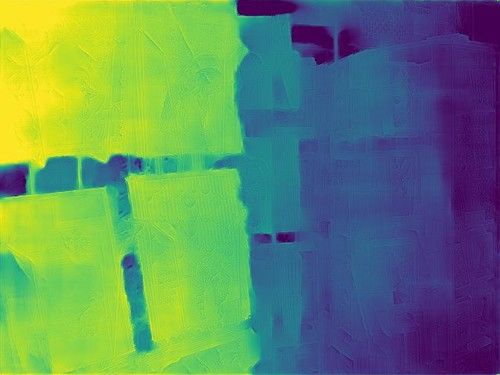} &
    \includegraphics[width=\resultswidthlandscape]{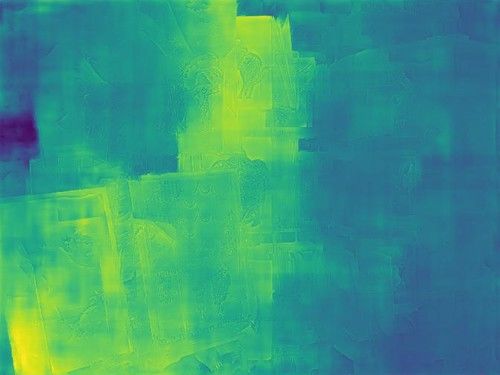} &
    \includegraphics[width=\resultswidthlandscape]{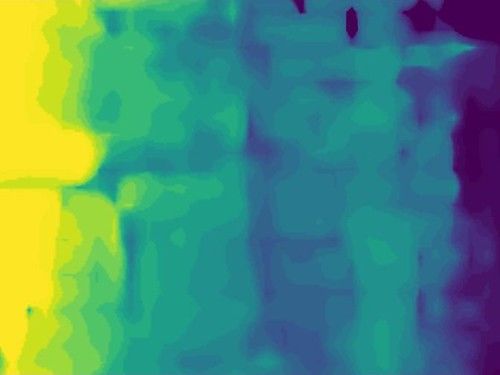} &
    \includegraphics[width=\resultswidthlandscape]{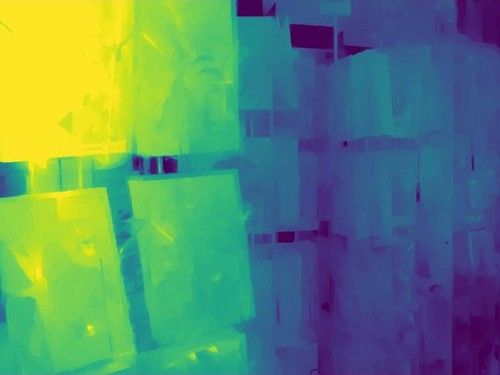} \\
    \includegraphics[width=\resultswidthlandscape]{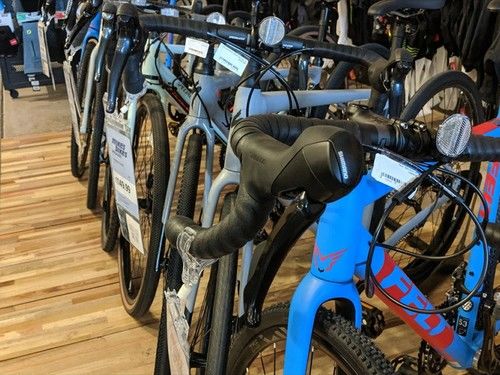} &
    \includegraphics[width=\resultswidthlandscape]{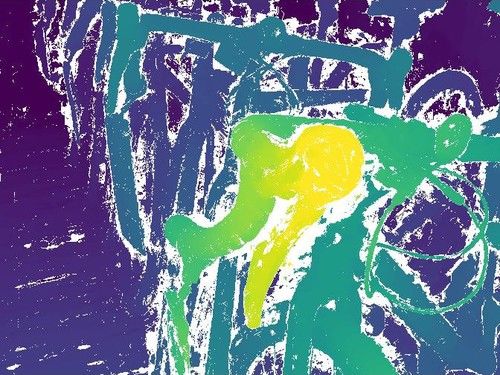} & 
    \includegraphics[width=\resultswidthlandscape]{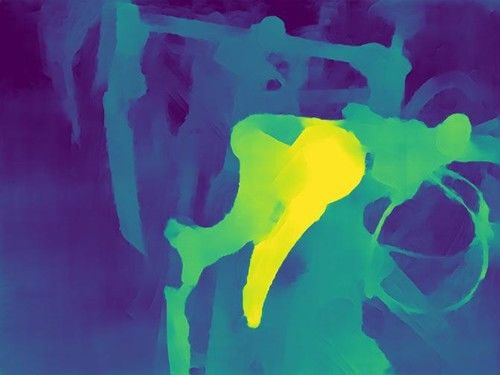} &
    \includegraphics[width=\resultswidthlandscape]{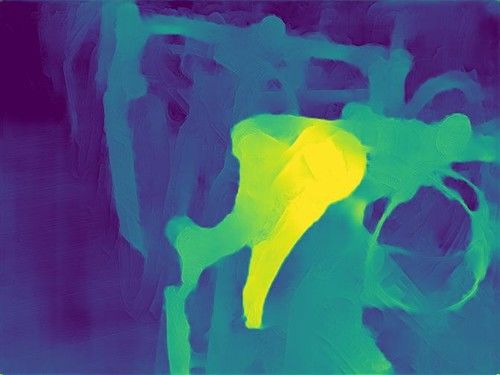} &
    \includegraphics[width=\resultswidthlandscape]{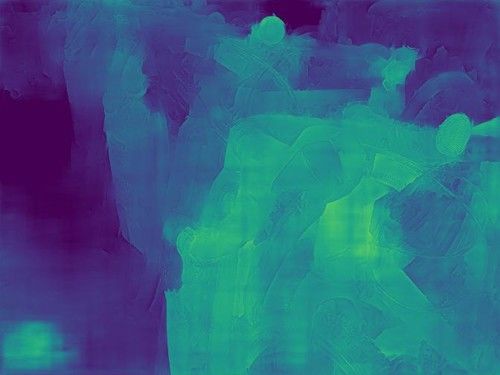} &
    \includegraphics[width=\resultswidthlandscape]{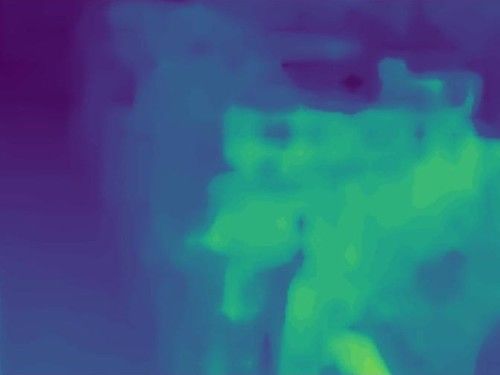} &
    \includegraphics[width=\resultswidthlandscape]{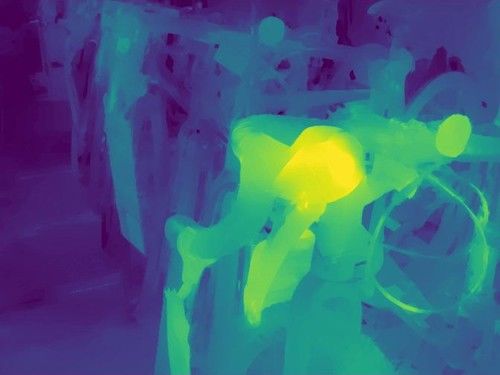} \\
    \fakecaption{(a) \,Input Image} & \fakecaption{(b)\,GT Depth} & \fakecaption{(c)\,DPNet (Affine)} & \fakecaption{(d)\, DPNet (Scale)} & \fakecaption{(e)\, DPNet (Affine)} & \fakecaption{(f)\,DORN \cite{DORN2018}} & \fakecaption{(g)\,Wadhwa \etal} \\
    \addlinespace[-1.2ex]
    \fakecaption{} & \fakecaption{} & \fakecaption{RGB + DP} & \fakecaption{RGB + DP} & \fakecaption{RGB} & \fakecaption{RGB} & \fakecaption{RGB + DP  \cite{wadhwa2018}} \\
    \end{tabular}
    \caption{Additional results similar to Table 5 in the main paper.}
    \label{fig:depth_showcase_landscape}
\end{figure*}

\newcommand{\resultswidthportrait}{0.135\textwidth}
\begin{figure*}
    \centering
    \begin{tabular}{@{}c@{\,\,}c@{\,\,}c@{\,\,}c@{\,\,}c@{\,\,}c@{\,\,}c@{}}
    \includegraphics[width=\resultswidthportrait]{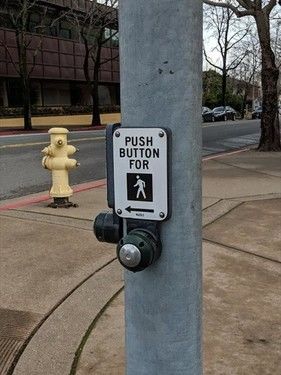} &
    \includegraphics[width=\resultswidthportrait]{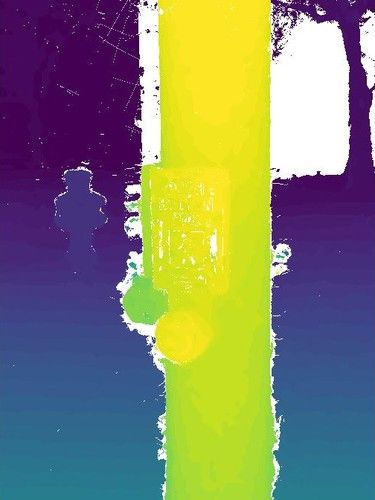} & 
    \includegraphics[width=\resultswidthportrait]{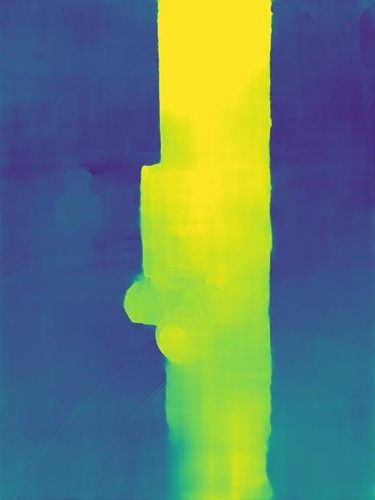} &
    \includegraphics[width=\resultswidthportrait]{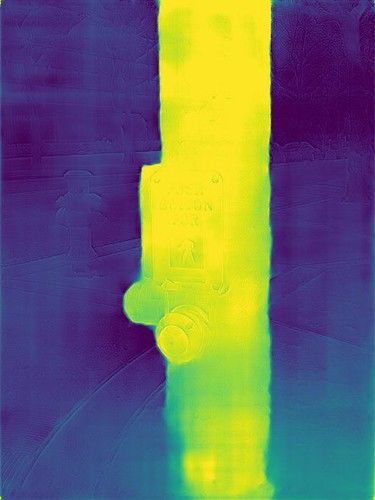} &
    \includegraphics[width=\resultswidthportrait]{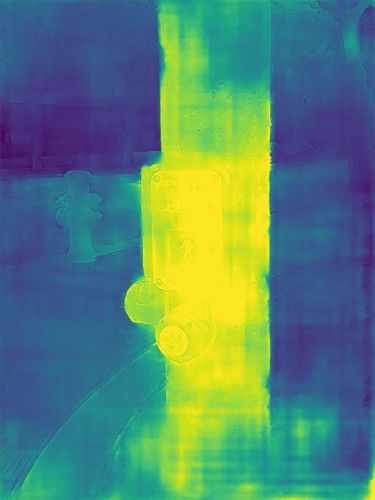} &
    \includegraphics[width=\resultswidthportrait]{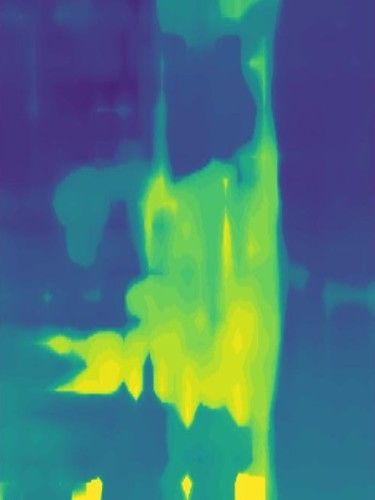} &
    \includegraphics[width=\resultswidthportrait]{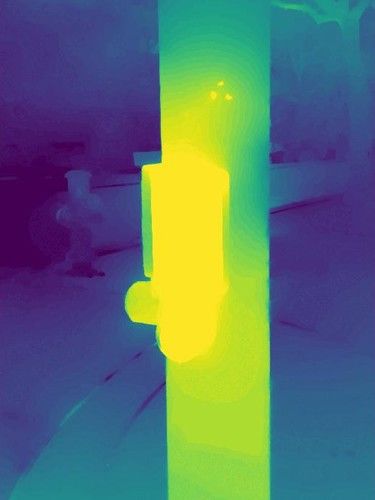} \\
    \includegraphics[width=\resultswidthportrait]{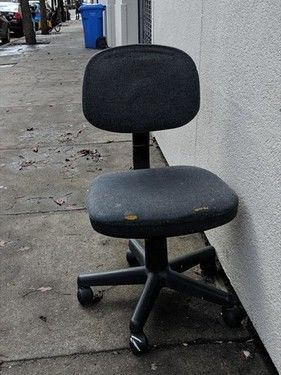} &
    \includegraphics[width=\resultswidthportrait]{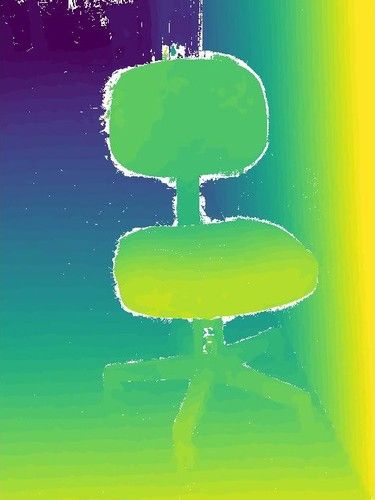} & 
    \includegraphics[width=\resultswidthportrait]{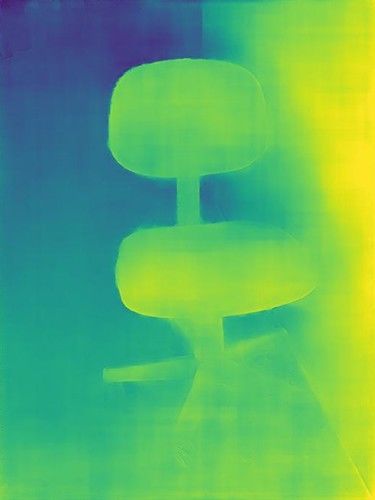} &
    \includegraphics[width=\resultswidthportrait]{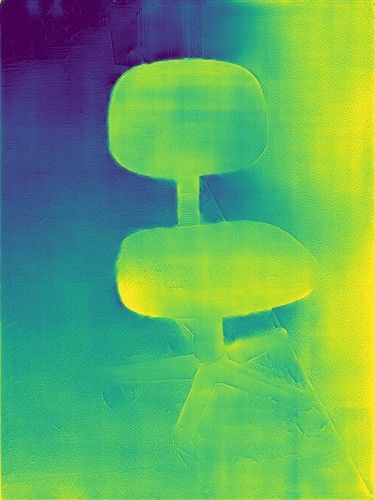} &
    \includegraphics[width=\resultswidthportrait]{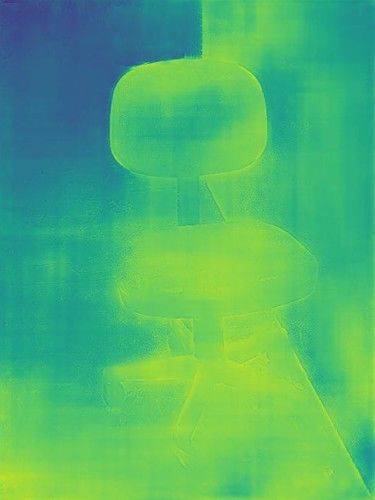} &
    \includegraphics[width=\resultswidthportrait]{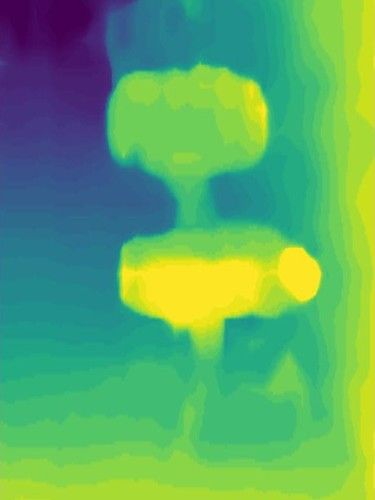} &
    \includegraphics[width=\resultswidthportrait]{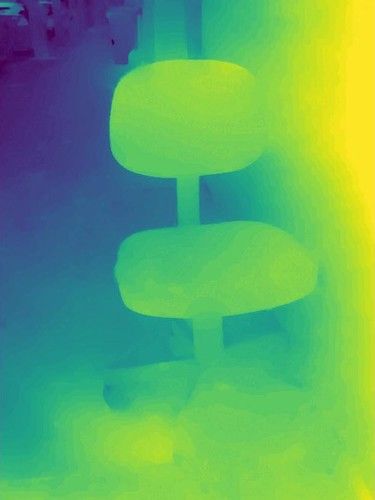} \\
    \includegraphics[width=\resultswidthportrait]{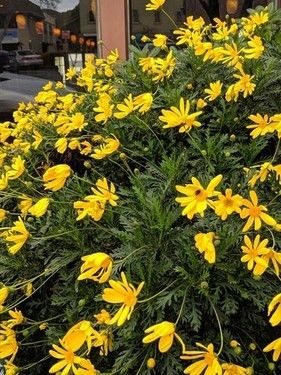} &
    \includegraphics[width=\resultswidthportrait]{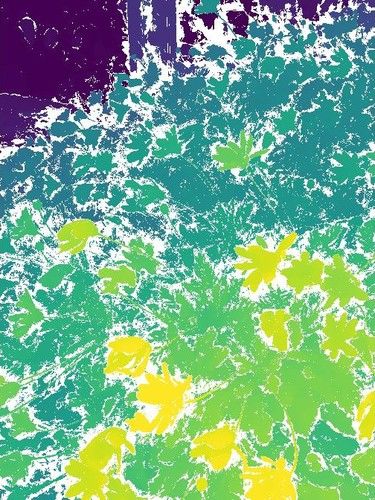} & 
    \includegraphics[width=\resultswidthportrait]{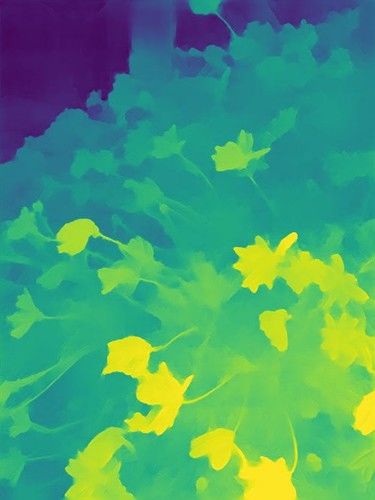} &
    \includegraphics[width=\resultswidthportrait]{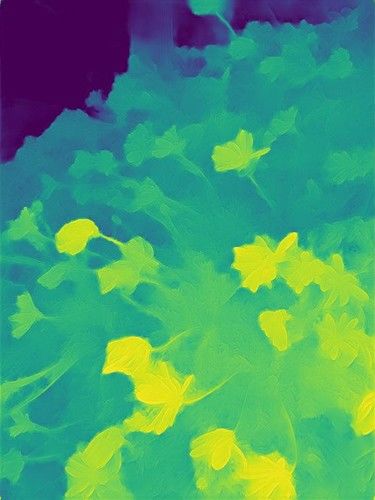} &
    \includegraphics[width=\resultswidthportrait]{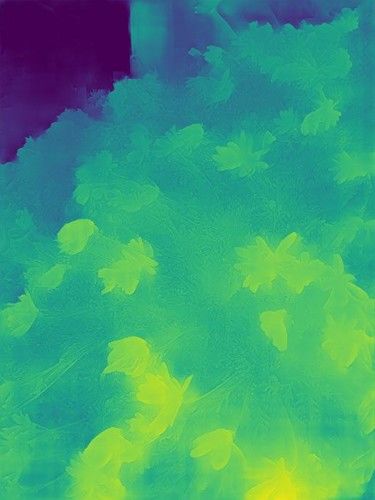} &
    \includegraphics[width=\resultswidthportrait]{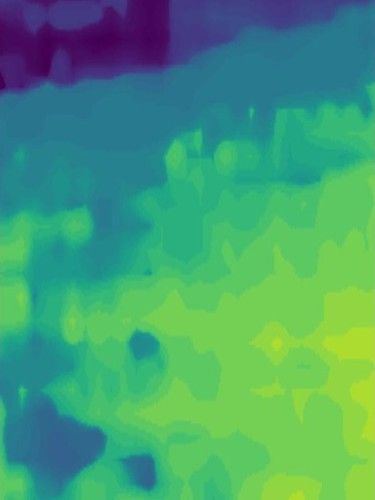} &
    \includegraphics[width=\resultswidthportrait]{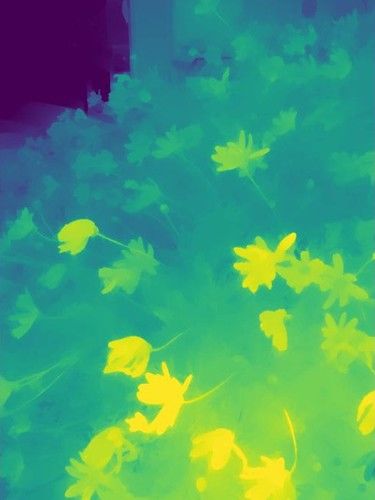} \\
    \includegraphics[width=\resultswidthportrait]{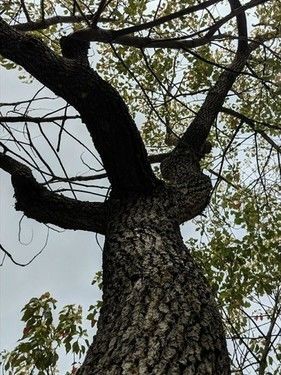} &
    \includegraphics[width=\resultswidthportrait]{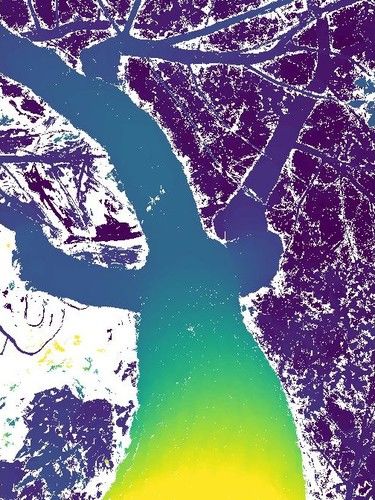} & 
    \includegraphics[width=\resultswidthportrait]{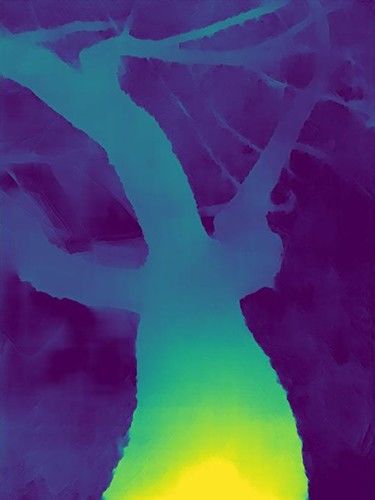} &
    \includegraphics[width=\resultswidthportrait]{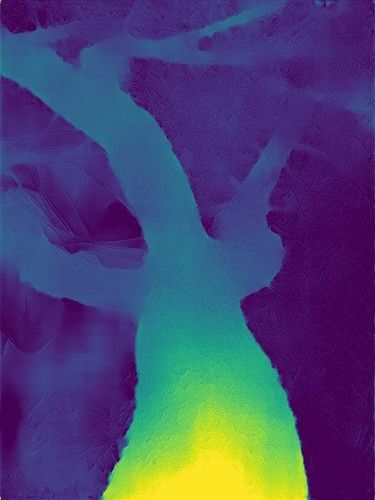} &
    \includegraphics[width=\resultswidthportrait]{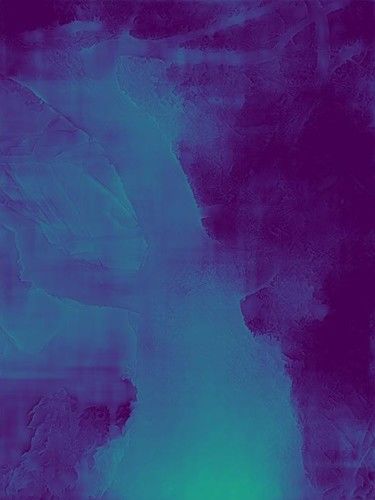} &
    \includegraphics[width=\resultswidthportrait]{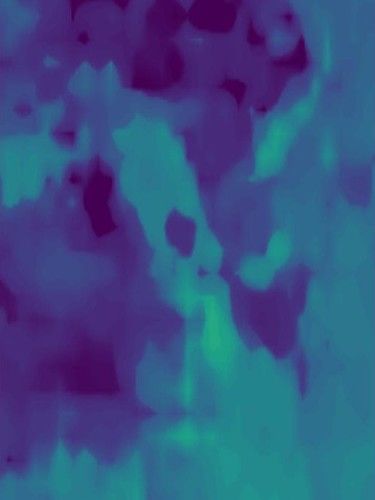} &
    \includegraphics[width=\resultswidthportrait]{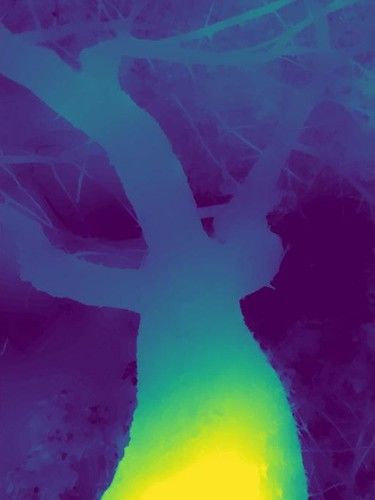} \\
    \includegraphics[width=\resultswidthportrait]{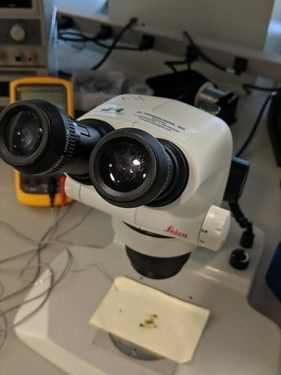} &
    \includegraphics[width=\resultswidthportrait]{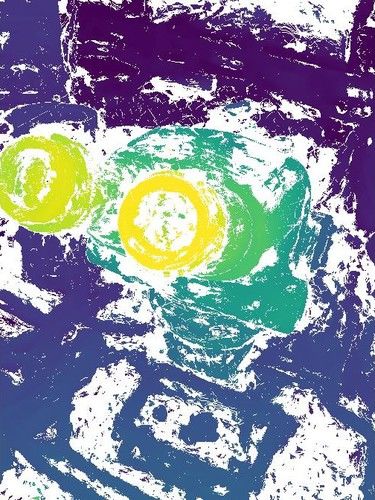} & 
    \includegraphics[width=\resultswidthportrait]{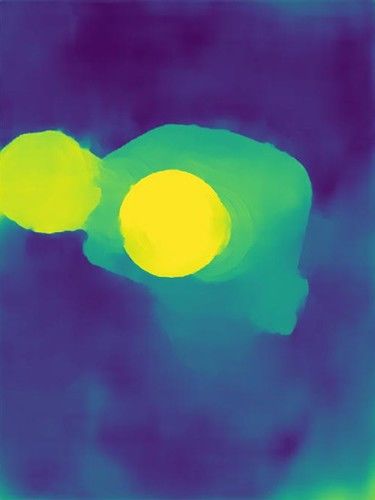} &
    \includegraphics[width=\resultswidthportrait]{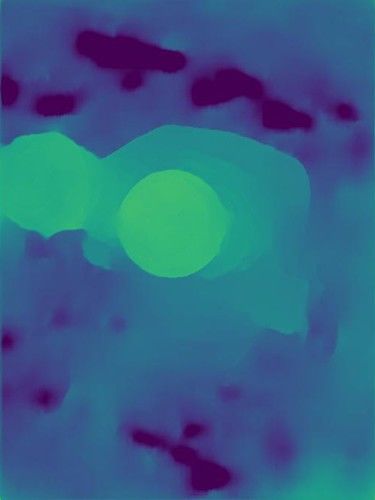} &
    \includegraphics[width=\resultswidthportrait]{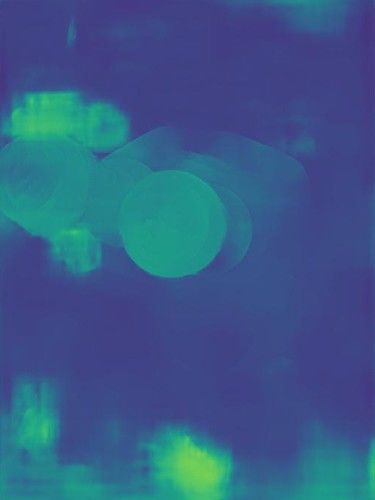} &
    \includegraphics[width=\resultswidthportrait]{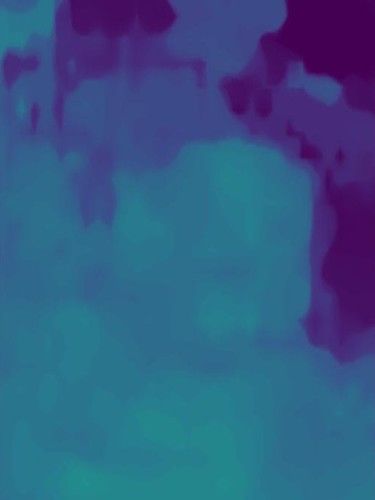} &
    \includegraphics[width=\resultswidthportrait]{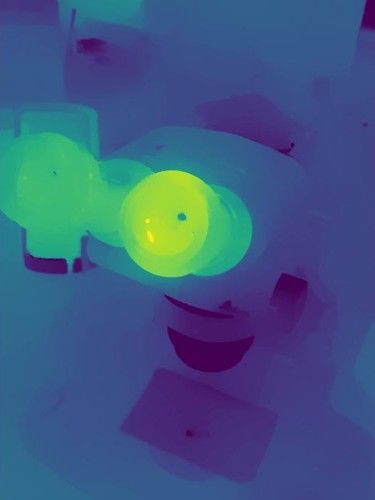} \\
    \fakecaption{(a) \,Input Image} & \fakecaption{(b)\,GT Depth} & \fakecaption{(c)\,DPNet (Affine)} & \fakecaption{(d)\, DPNet (Scale)} & \fakecaption{(e)\, DPNet (Affine)} & \fakecaption{(f)\,DORN \cite{DORN2018}} & \fakecaption{(g)\,Wadhwa \etal} \\
    \addlinespace[-1.2ex]
    \fakecaption{} & \fakecaption{} & \fakecaption{RGB + DP} & \fakecaption{RGB + DP} & \fakecaption{RGB} & \fakecaption{RGB} & \fakecaption{RGB + DP  \cite{wadhwa2018}} \\
    \end{tabular}
    \caption{Additional results similar to Table 5 in the main paper.}
    \label{fig:depth_showcase_portrait}
\end{figure*}

\begin{figure*}
    \centering
    \begin{tabular}{@{}c@{\,\,}c@{\,\,}c@{\,\,}c@{\,\,}c@{\,\,}c@{\,\,}c@{}}
    \includegraphics[width=\resultswidth]{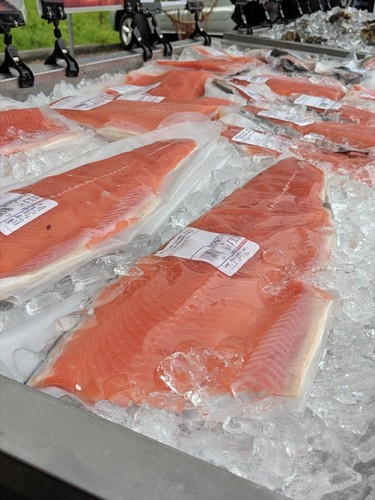} &
    \includegraphics[width=\resultswidth]{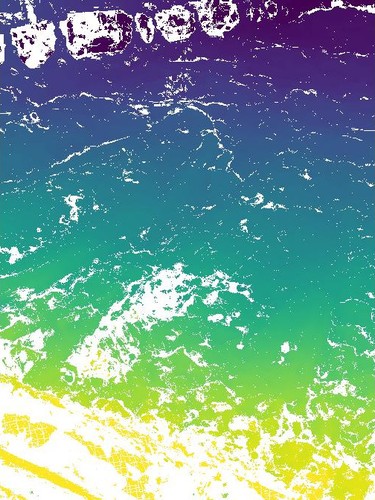} & 
    \includegraphics[width=\resultswidth]{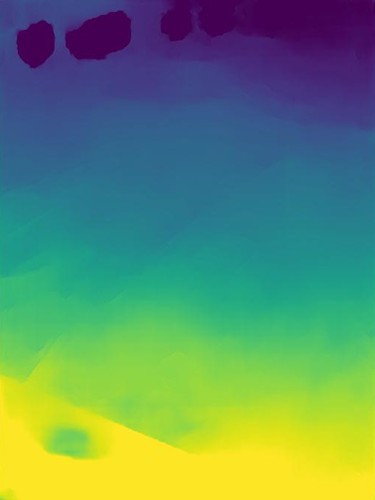} &
    \includegraphics[width=\resultswidth]{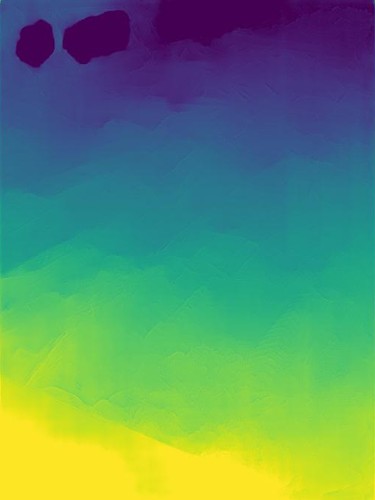} &
    \includegraphics[width=\resultswidth]{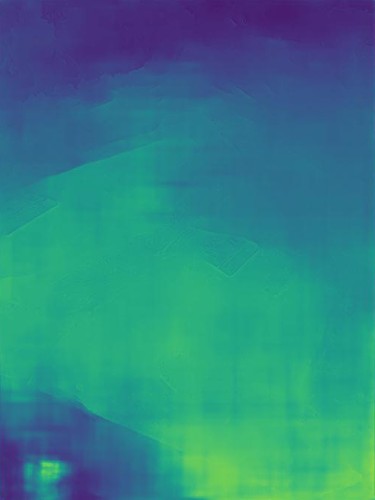} &
    \includegraphics[width=\resultswidth]{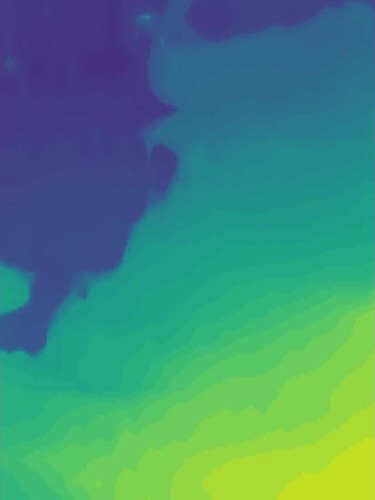} &
    \includegraphics[width=\resultswidth]{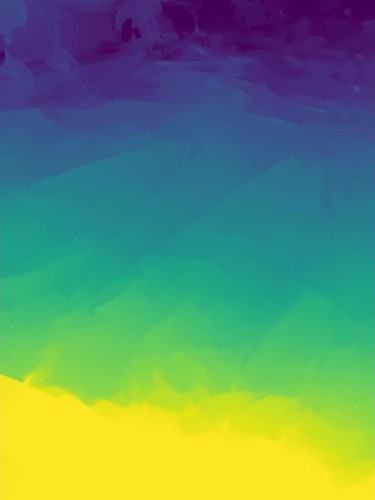} \\
    \includegraphics[width=\resultswidth]{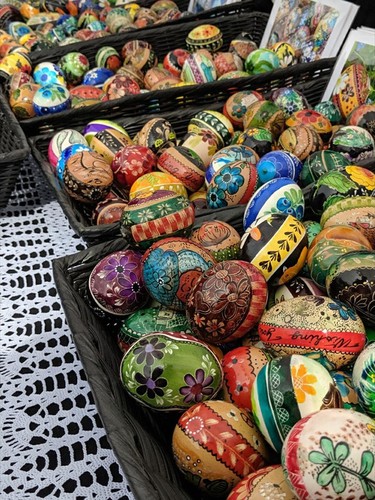} &
    \includegraphics[width=\resultswidth]{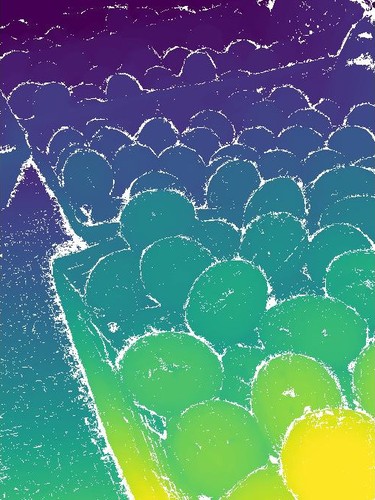} & 
    \includegraphics[width=\resultswidth]{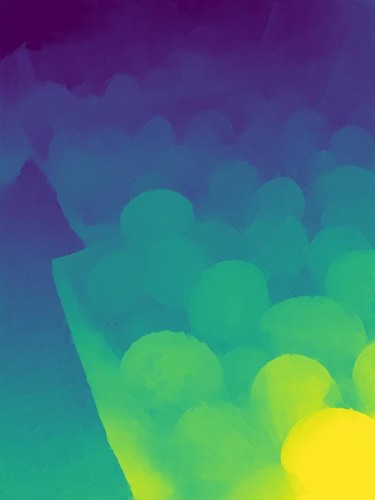} &
    \includegraphics[width=\resultswidth]{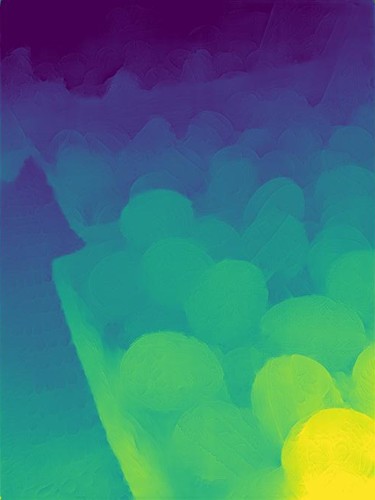} &
    \includegraphics[width=\resultswidth]{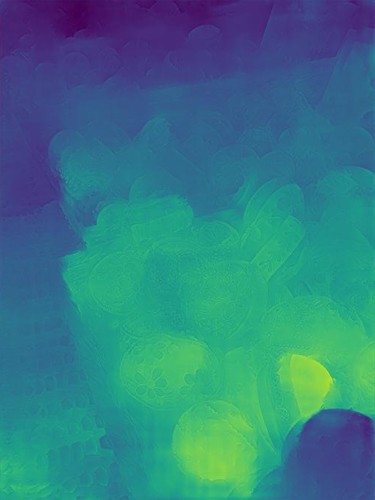} &
    \includegraphics[width=\resultswidth]{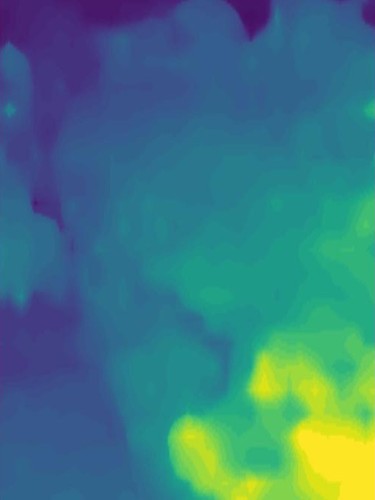} &
    \includegraphics[width=\resultswidth]{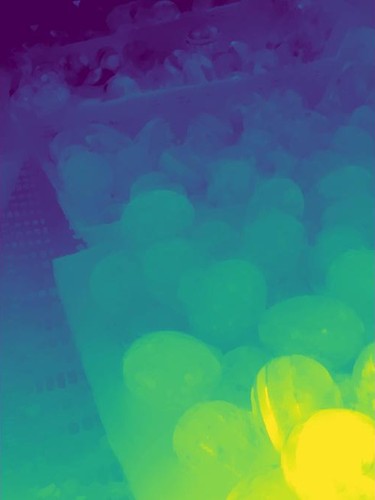} \\
    \includegraphics[width=\resultswidth]{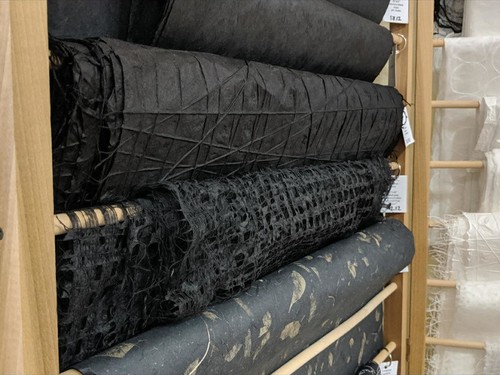} &
    \includegraphics[width=\resultswidth]{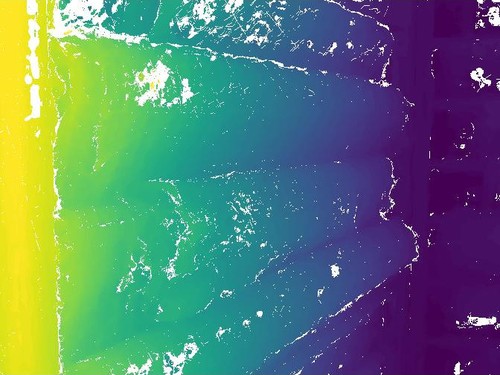} & 
    \includegraphics[width=\resultswidth]{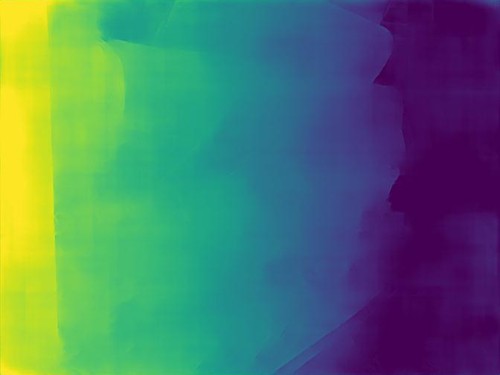} &
    \includegraphics[width=\resultswidth]{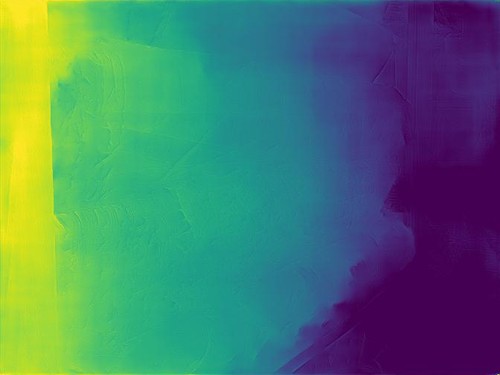} &
    \includegraphics[width=\resultswidth]{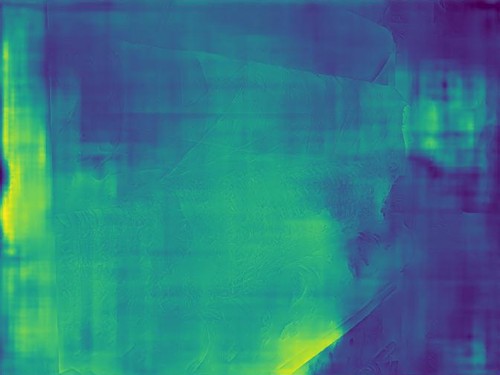} &
    \includegraphics[width=\resultswidth]{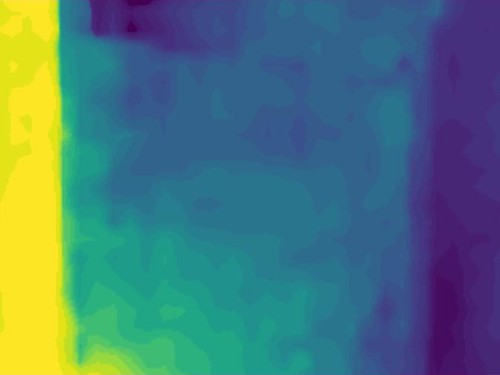} &
    \includegraphics[width=\resultswidth]{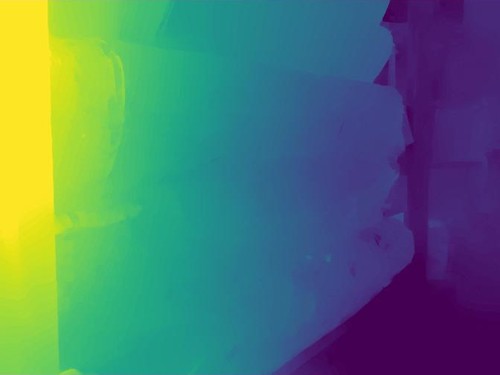} \\
    \includegraphics[width=\resultswidth]{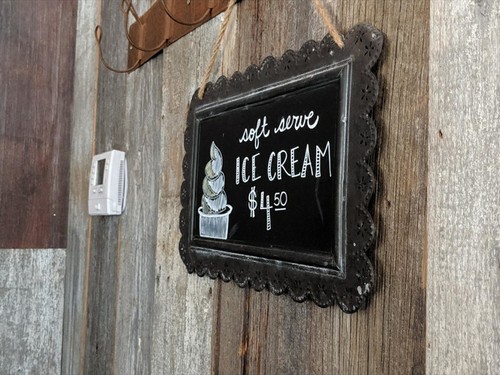} &
    \includegraphics[width=\resultswidth]{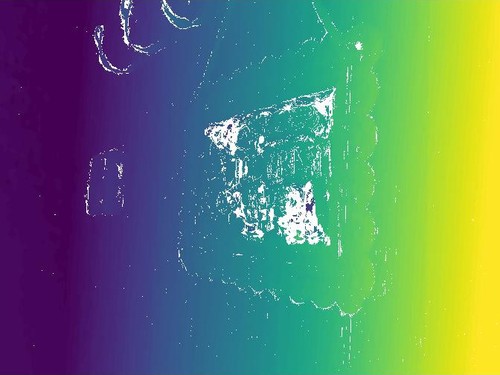} & 
    \includegraphics[width=\resultswidth]{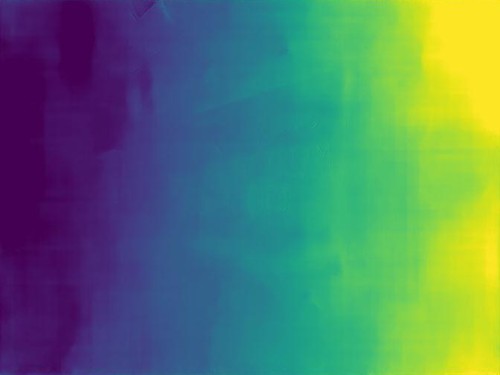} &
    \includegraphics[width=\resultswidth]{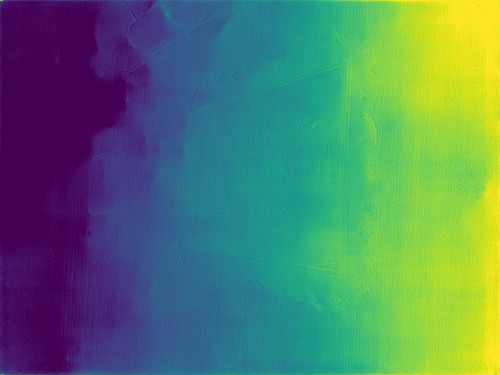} &
    \includegraphics[width=\resultswidth]{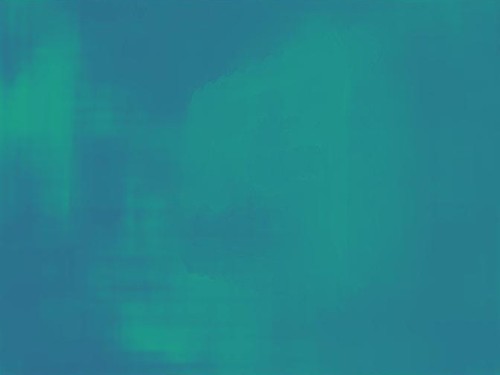} &
    \includegraphics[width=\resultswidth]{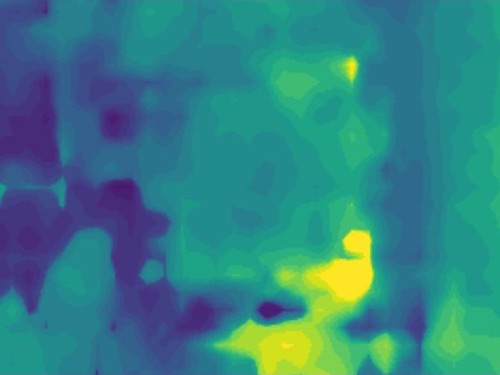} &
    \includegraphics[width=\resultswidth]{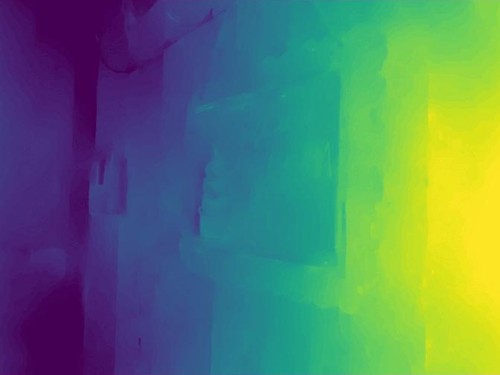} \\
    \includegraphics[width=\resultswidth]{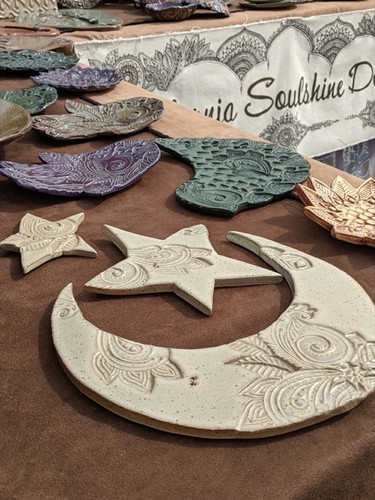} &
    \includegraphics[width=\resultswidth]{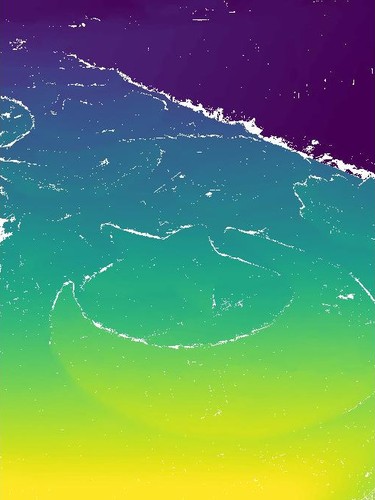} & 
    \includegraphics[width=\resultswidth]{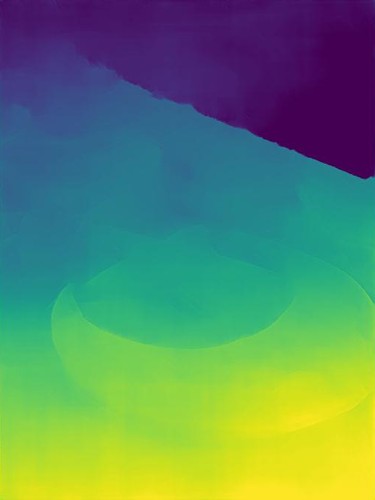} &
    \includegraphics[width=\resultswidth]{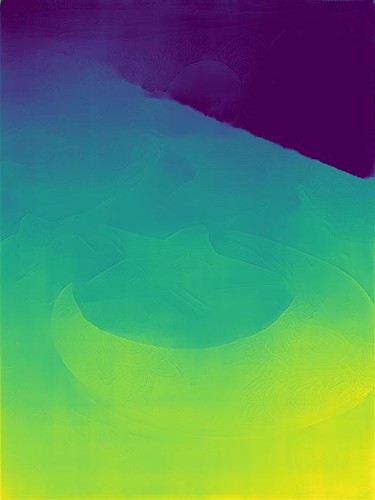} &
    \includegraphics[width=\resultswidth]{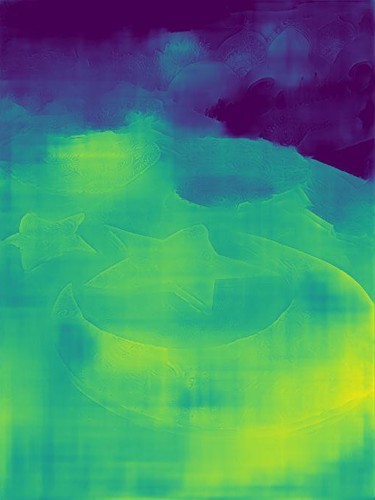} &
    \includegraphics[width=\resultswidth]{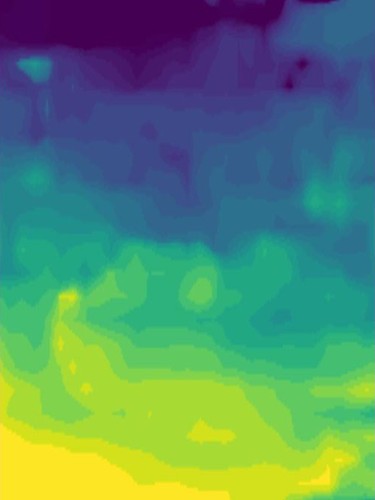} &
    \includegraphics[width=\resultswidth]{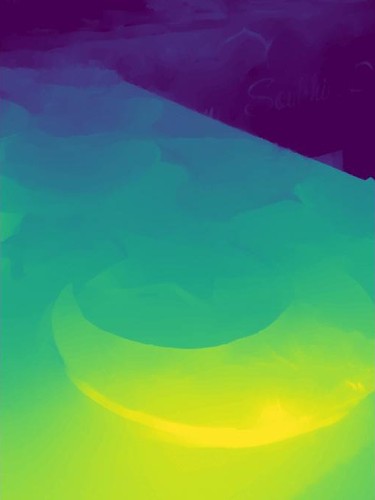} \\
    \fakecaption{(a) \,Input Image} & \fakecaption{(b)\,GT Depth} & \fakecaption{(c)\,DPNet (Affine)} & \fakecaption{(d)\, DPNet (Scale)} & \fakecaption{(e)\, DPNet (Affine)} & \fakecaption{(f)\,DORN \cite{DORN2018}} & \fakecaption{(g)\,Wadhwa \etal} \\
    \addlinespace[-1.2ex]
    \fakecaption{} & \fakecaption{} & \fakecaption{RGB + DP} & \fakecaption{RGB + DP} & \fakecaption{RGB} & \fakecaption{RGB} & \fakecaption{RGB + DP  \cite{wadhwa2018}} \\
    \end{tabular}
    \caption{Top 5 results of our method (DPNet with affine invariance) as determined by $1-|\rho_s|$ metric. These tend to be textured scenes with close focus distances.}
    \label{fig:best_5}
\end{figure*}

\begin{figure*}
    \centering
    \begin{tabular}{@{}c@{\,\,}c@{\,\,}c@{\,\,}c@{\,\,}c@{\,\,}c@{\,\,}c@{}}
    \includegraphics[width=\resultswidth]{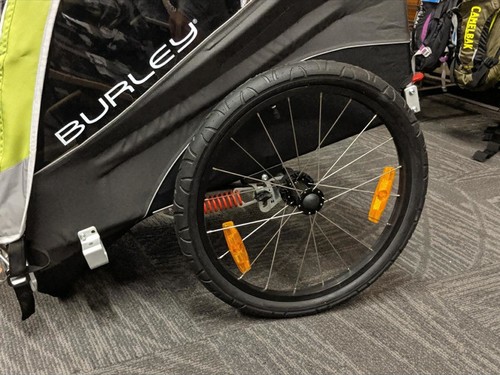} &
    \includegraphics[width=\resultswidth]{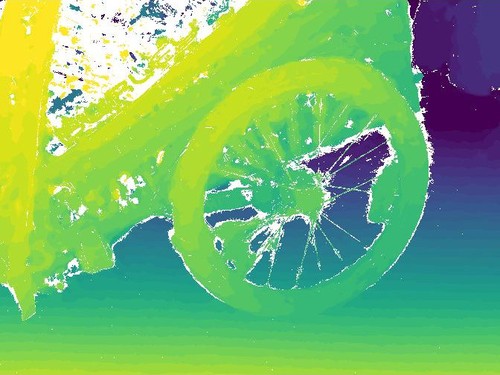} & 
    \includegraphics[width=\resultswidth]{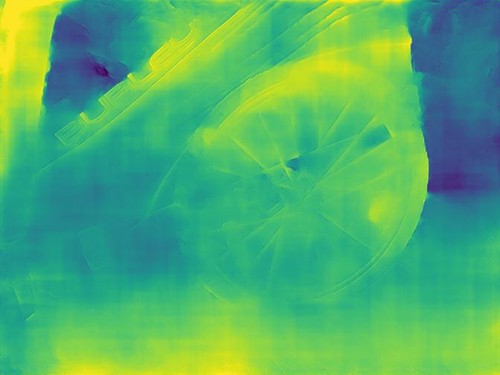} &
    \includegraphics[width=\resultswidth]{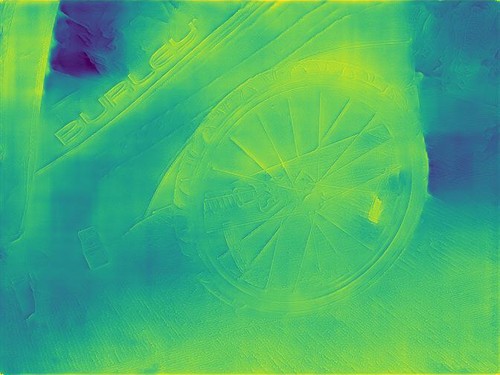} &
    \includegraphics[width=\resultswidth]{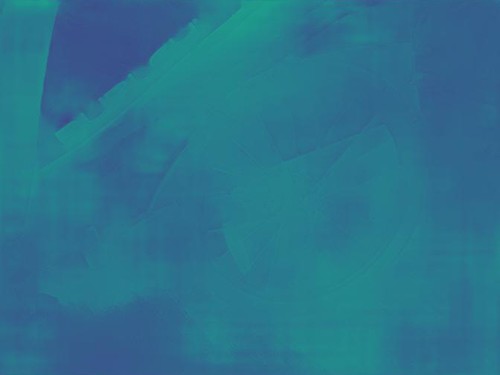} &
    \includegraphics[width=\resultswidth]{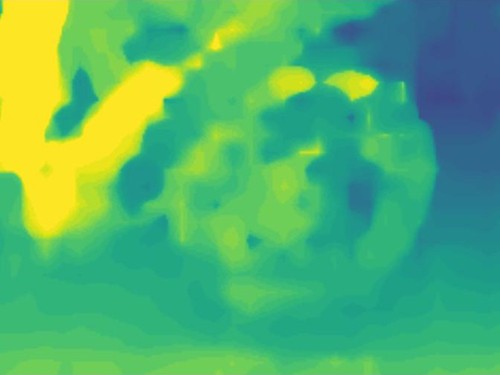} &
    \includegraphics[width=\resultswidth]{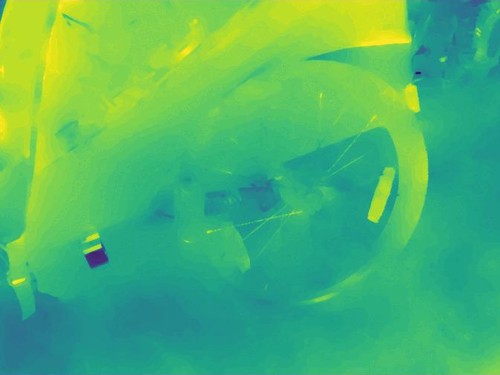} \\
    \includegraphics[width=\resultswidth]{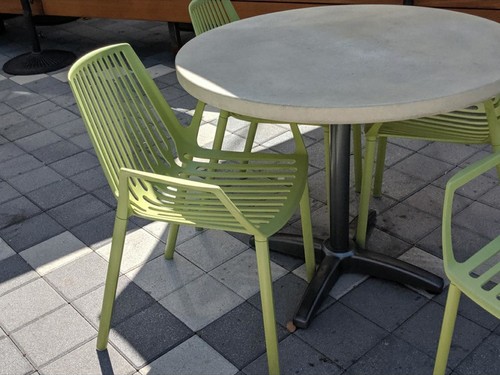} &
    \includegraphics[width=\resultswidth]{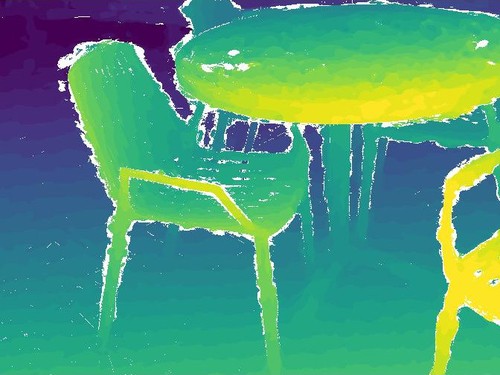} & 
    \includegraphics[width=\resultswidth]{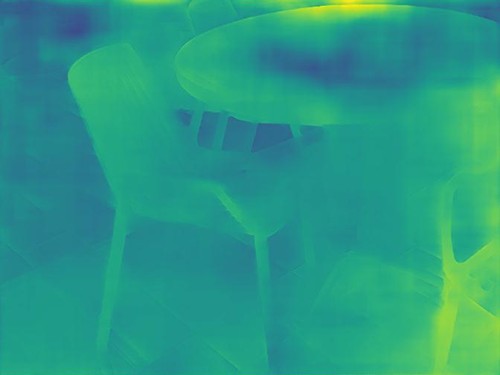} &
    \includegraphics[width=\resultswidth]{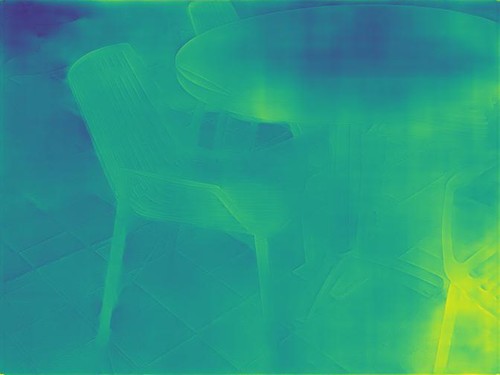} &
    \includegraphics[width=\resultswidth]{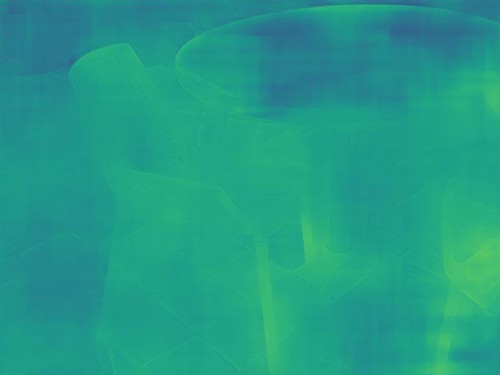} &
    \includegraphics[width=\resultswidth]{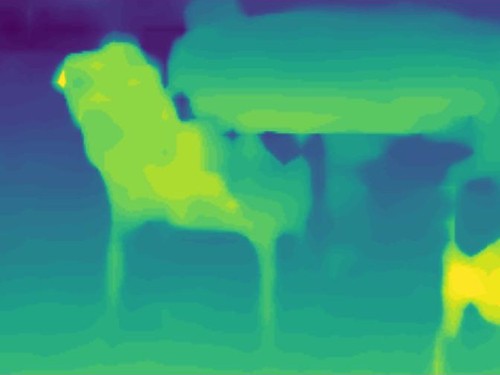} &
    \includegraphics[width=\resultswidth]{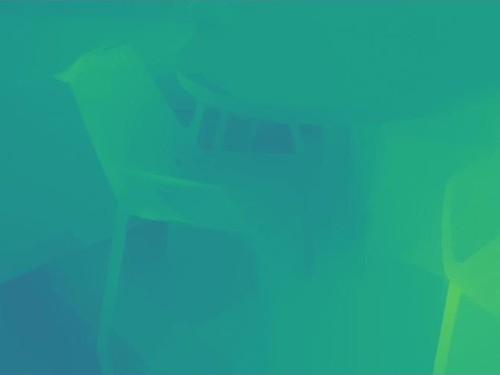} \\
    \includegraphics[width=\resultswidth]{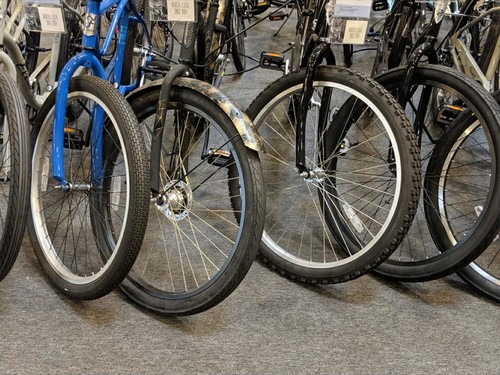} &
    \includegraphics[width=\resultswidth]{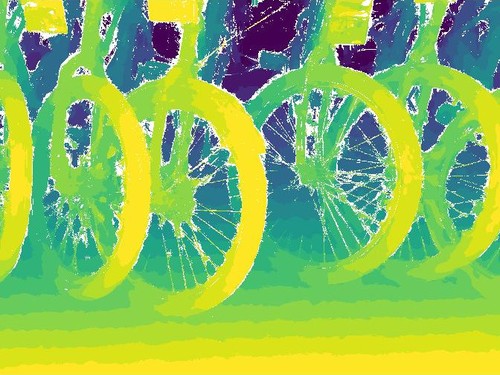} & 
    \includegraphics[width=\resultswidth]{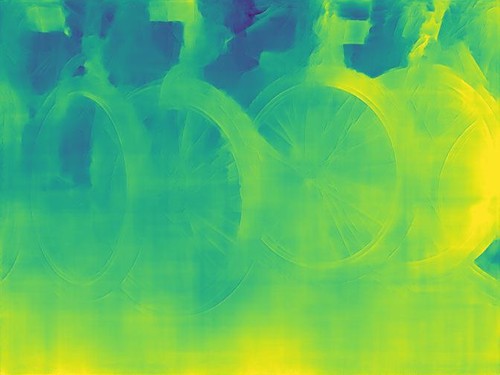} &
    \includegraphics[width=\resultswidth]{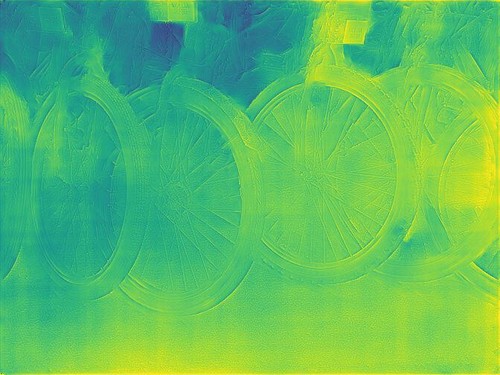} &
    \includegraphics[width=\resultswidth]{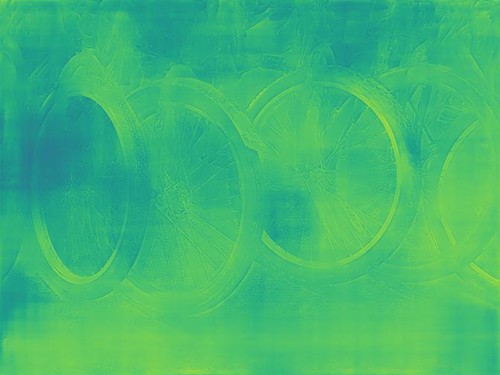} &
    \includegraphics[width=\resultswidth]{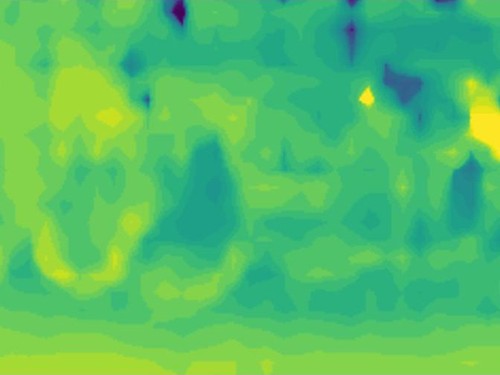} &
    \includegraphics[width=\resultswidth]{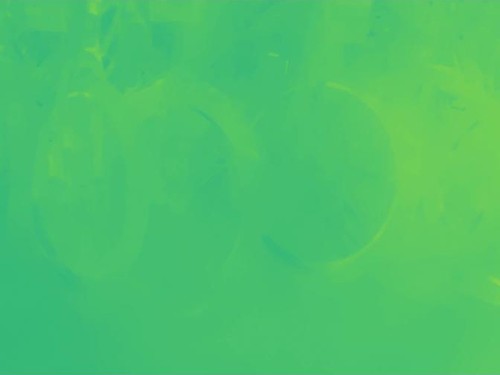} \\
    \includegraphics[width=\resultswidth]{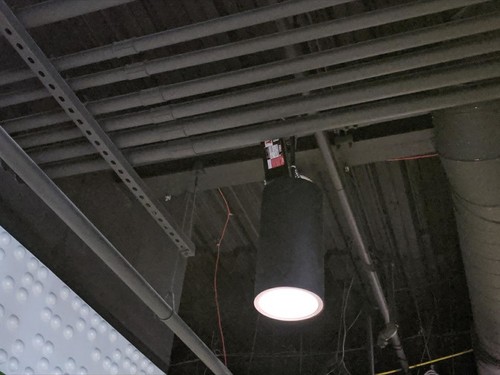} &
    \includegraphics[width=\resultswidth]{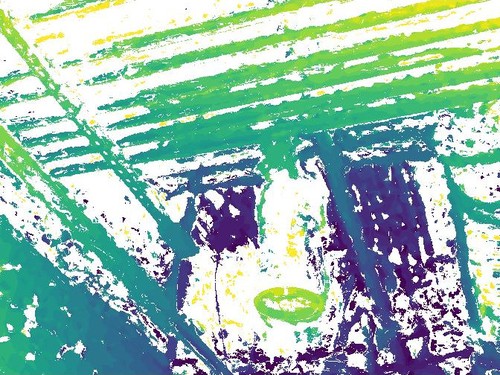} & 
    \includegraphics[width=\resultswidth]{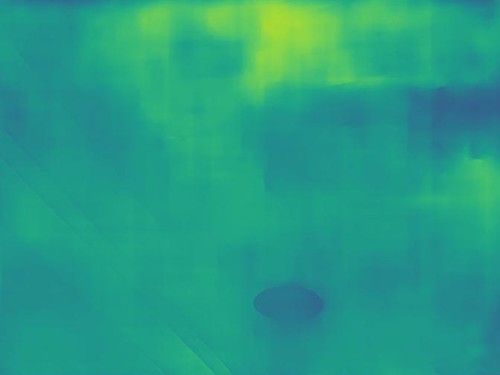} &
    \includegraphics[width=\resultswidth]{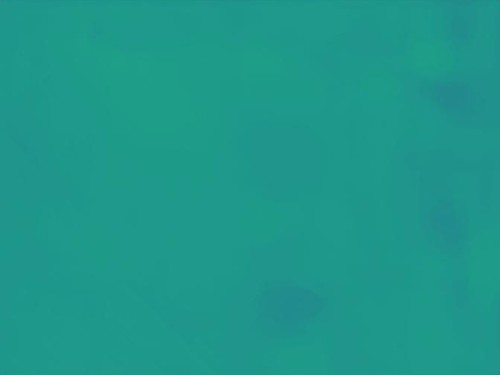} &
    \includegraphics[width=\resultswidth]{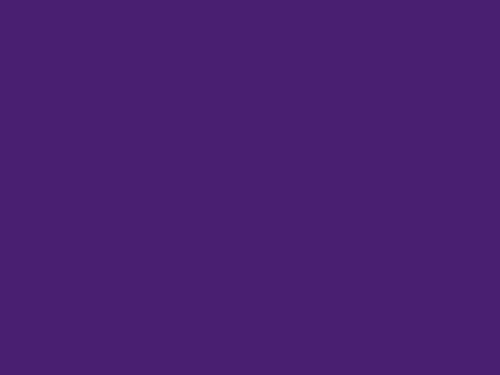} &
    \includegraphics[width=\resultswidth]{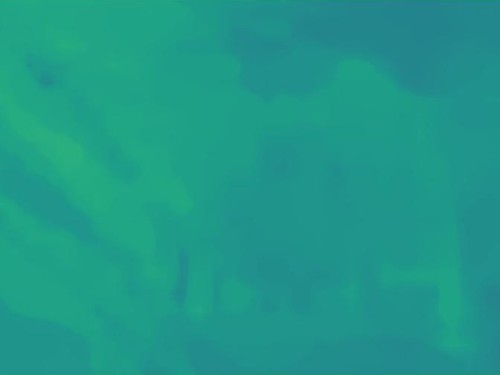} &
    \includegraphics[width=\resultswidth]{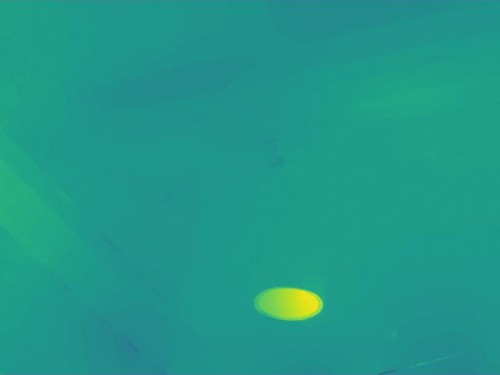} \\
    \includegraphics[width=\resultswidth]{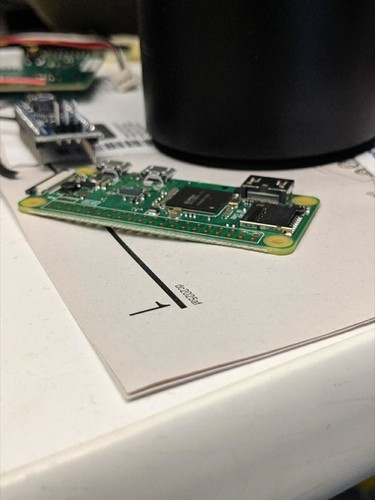} &
    \includegraphics[width=\resultswidth]{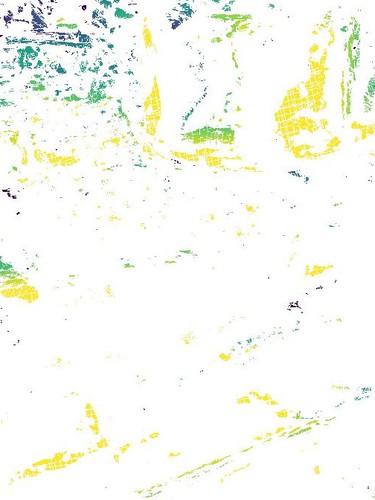} & 
    \includegraphics[width=\resultswidth]{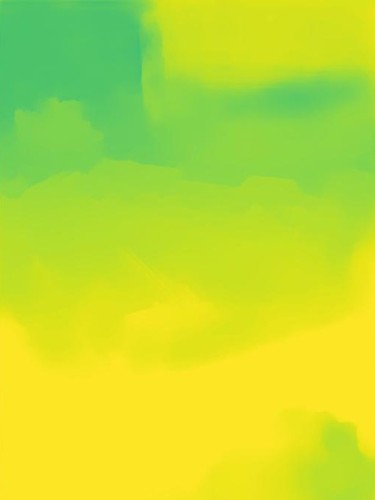} &
    \includegraphics[width=\resultswidth]{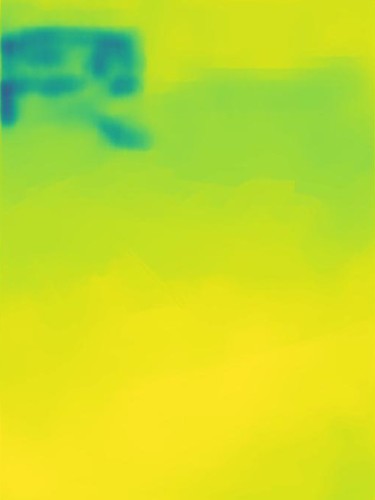} &
    \includegraphics[width=\resultswidth]{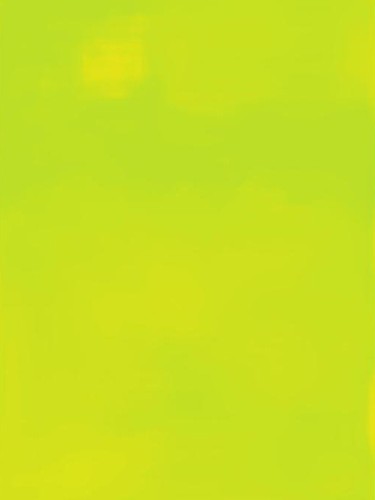} &
    \includegraphics[width=\resultswidth]{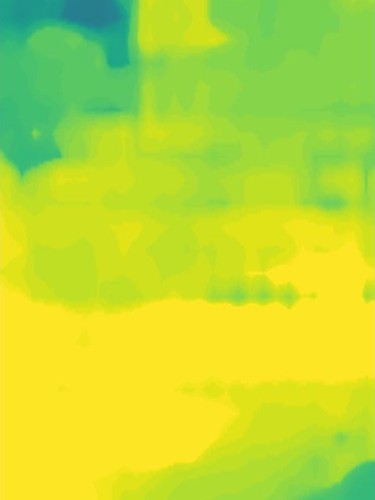} &
    \includegraphics[width=\resultswidth]{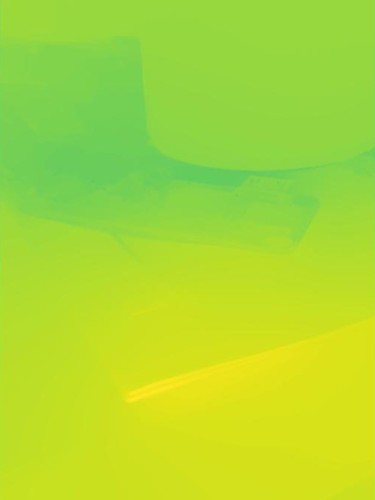} \\
    \fakecaption{(a) \,Input Image} & \fakecaption{(b)\,GT Depth} & \fakecaption{(c)\,DPNet (Affine)} & \fakecaption{(d)\, DPNet (Scale)} & \fakecaption{(e)\, DPNet (Affine)} & \fakecaption{(f)\,DORN \cite{DORN2018}} & \fakecaption{(g)\,Wadhwa \etal } \\
    \addlinespace[-1.2ex]
    \fakecaption{} & \fakecaption{} & \fakecaption{RGB + DP} & \fakecaption{RGB + DP} & \fakecaption{RGB} & \fakecaption{RGB} & \fakecaption{RGB + DP \cite{wadhwa2018}} \\
    \end{tabular}
    \caption{Worst 5 results of our method (DPNet with affine invariance) as determined by $1-|\rho_s|$ metric. These tend to be scenes with textureless surfaces or containing far off objects. For the fourth image in column (e), a particularly bad prediction results in an incorrect affine mapping that collapses all depths to a single value.}
    \label{fig:worst_5}
\end{figure*}

We show additional results in Fig.~\ref{fig:depth_showcase_landscape} and \ref{fig:depth_showcase_portrait}. In addition, we also show the best and the worst 5 results of our method as determined by $1-|\rho_s|$ metric in Fig.~\ref{fig:best_5} and \ref{fig:worst_5} respectively.

\subsection{Results on data from Wadhwa \etal \cite{wadhwa2018}}

\newcommand{\portraitmodewidth}{0.1935\textwidth}
\begin{figure*}
    \centering
    \begin{tabular}{@{}c@{\,\,}c@{\,\,}c@{\,\,}c@{\,\,}c@{}}
    \includegraphics[width=\portraitmodewidth]{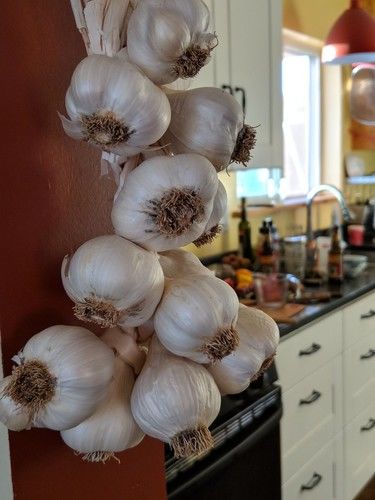} &
    \includegraphics[width=\portraitmodewidth]{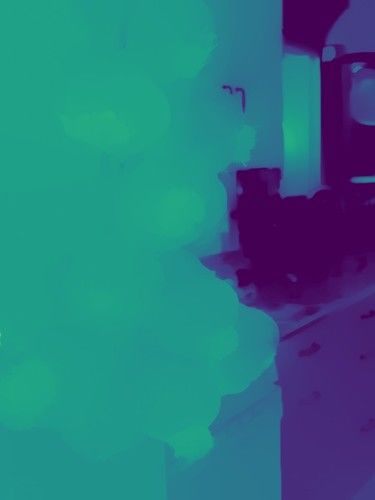} & 
    \includegraphics[width=\portraitmodewidth]{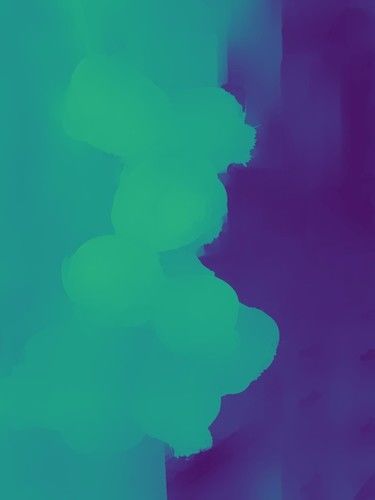} &
    \includegraphics[width=\portraitmodewidth]{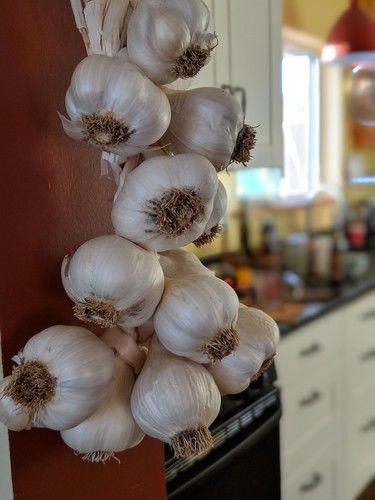} &
    \includegraphics[width=\portraitmodewidth]{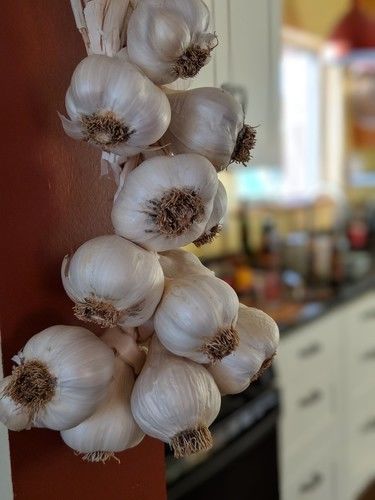}\\
    \includegraphics[width=\portraitmodewidth]{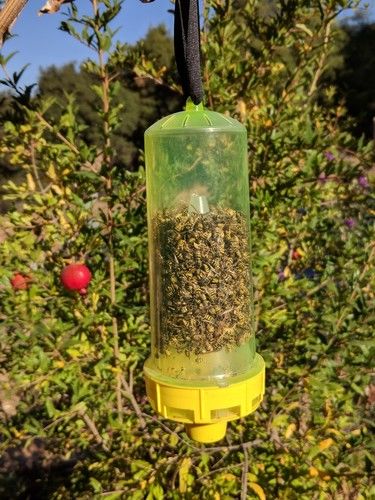} &
    \includegraphics[width=\portraitmodewidth]{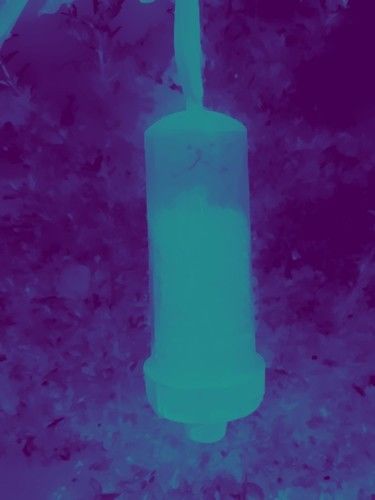} & 
    \includegraphics[width=\portraitmodewidth]{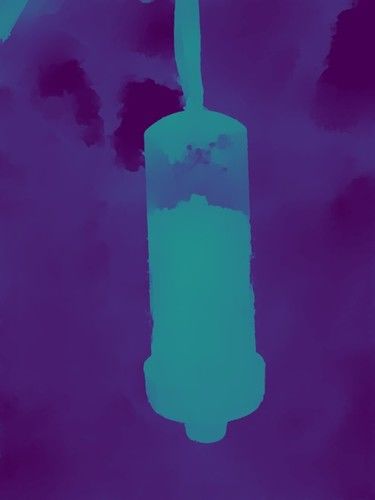} &
    \includegraphics[width=\portraitmodewidth]{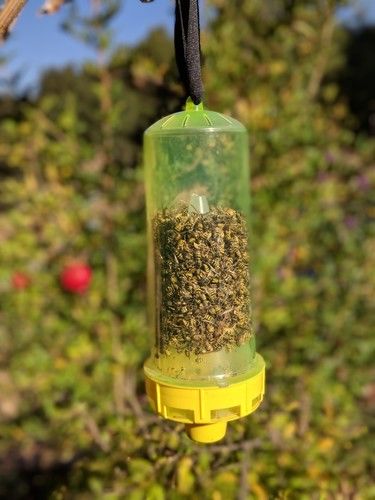} &
    \includegraphics[width=\portraitmodewidth]{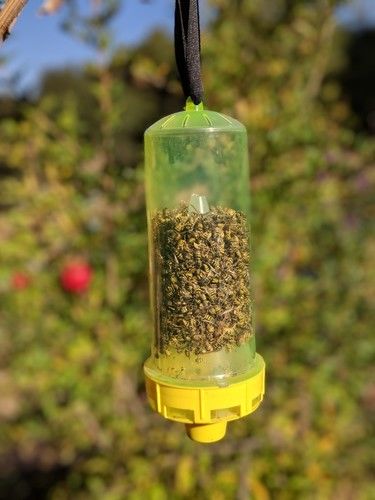}\\
    \includegraphics[width=\portraitmodewidth]{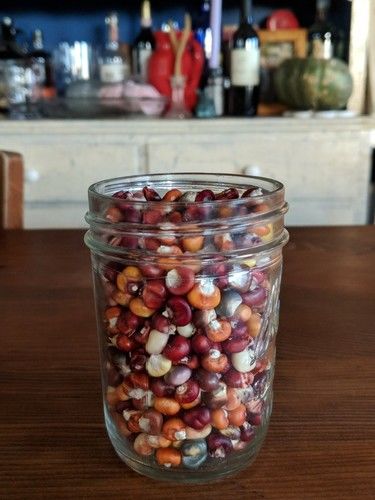} &
    \includegraphics[width=\portraitmodewidth]{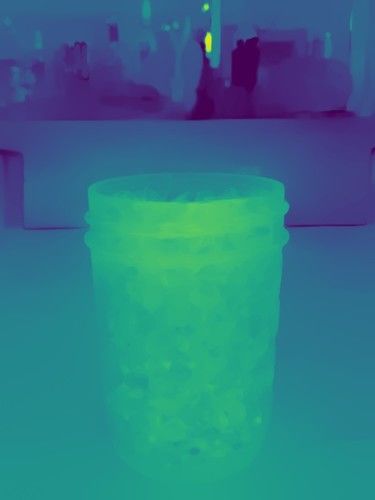} & 
    \includegraphics[width=\portraitmodewidth]{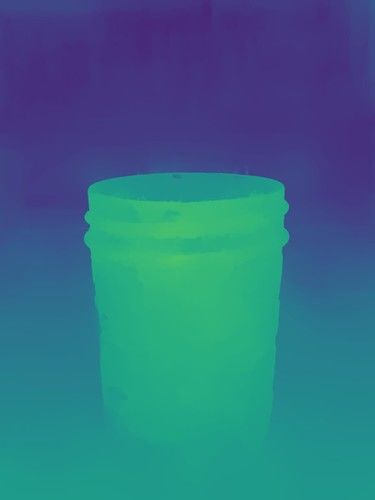} &
    \includegraphics[width=\portraitmodewidth]{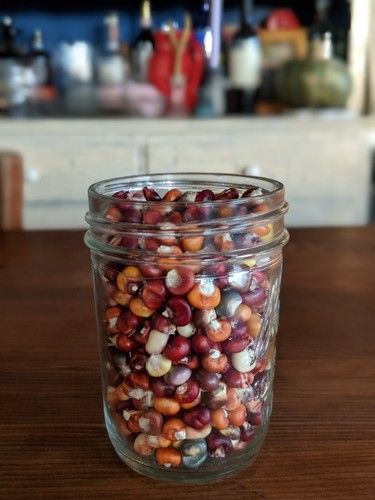} &
    \includegraphics[width=\portraitmodewidth]{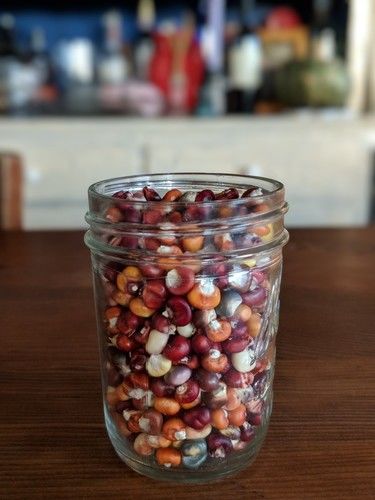}\\
    \includegraphics[width=\portraitmodewidth]{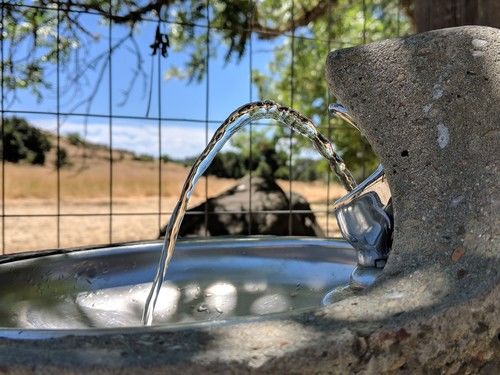} &
    \includegraphics[width=\portraitmodewidth]{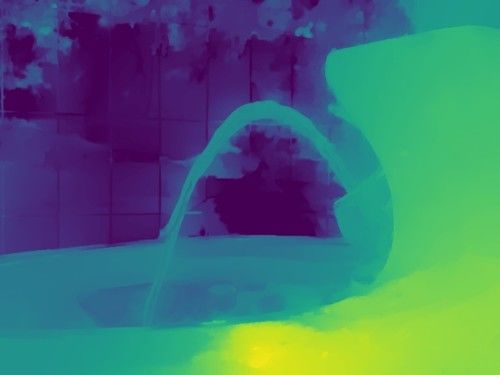} & 
    \includegraphics[width=\portraitmodewidth]{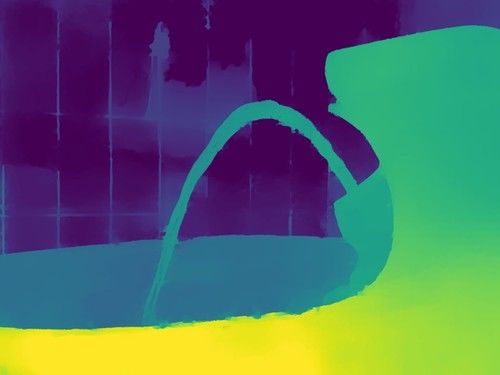} &
    \includegraphics[width=\portraitmodewidth]{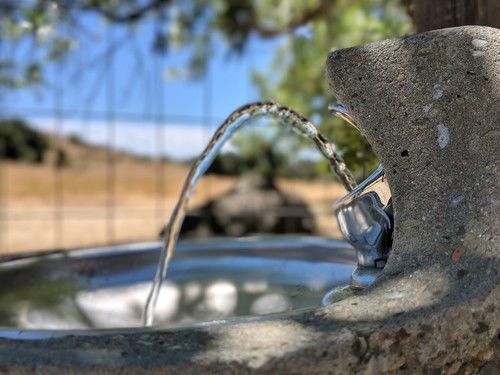} &
    \includegraphics[width=\portraitmodewidth]{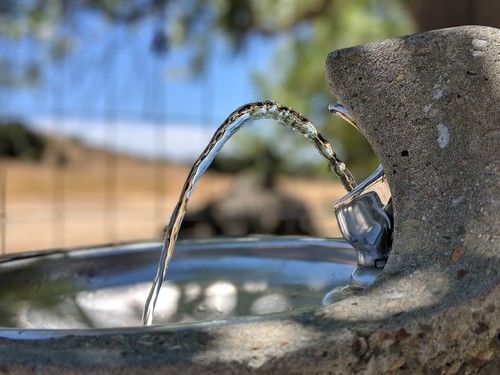}\\
    \fakecaption{(a) \,Input Image} & \fakecaption{(b)\, Depth from \cite{wadhwa2018}} & \fakecaption{(c)\,Our Depth} & \fakecaption{(d)\, Defocus result from \cite{wadhwa2018}} & \fakecaption{(e)\,Our defocus result}\\
    \end{tabular}
    \caption{Using our learned model to predict depth and render synthetic defocus results on data from Wadhwa \etal \cite{wadhwa2018}. While depth maps from \cite{wadhwa2018} contain errors, especially on saturated and optically blurred regions, our depth maps are clean and produce pleasing synthetic defocus results. E.g., notice the sharp highlights in the background in the first and the third image, uneven background blur in second image, and incorrect foreground blur in final image in results from \cite{wadhwa2018}.}
    \label{fig:portrait_mode_results}
\end{figure*}

We also run our model on the dual-pixel data provided by \cite{wadhwa2018} and show that affine invariant depth can be used to synthetically defocus an image (Fig.~\ref{fig:portrait_mode_results}). We use our DPNet model trained with only DP input, and the unknown affine mapping is determined by choosing the depth to focus at and the amount of synthetic defocus to render. Similar to \cite{wadhwa2018}, we ensure that depth maps are edge-aligned with the corresponding RGB image by applying a bilateral filter followed by joint bilateral upsampling \cite{kopf2007joint}. Rendering is done using the algorithm of \cite{wadhwa2018}. As seen in the figure, fewer errors in depth results in fewer errors in the synthetically defocused image.

\end{document}